# Convergence Analysis of Two-layer Neural Networks with ReLU Activation

Yuanzhi Li
Computer Science Department
Princeton University
yuanzhil@cs.princeton.edu

Yang Yuan
Computer Science Department
Cornell University
yangyuan@cs.cornell.edu

**Abstract**

In recent years, stochastic gradient descent (SGD) based techniques has become the standard tools for training neural networks. However, formal theoretical understanding of why SGD can train neural networks in practice is largely missing.

In this paper, we make progress on understanding this mystery by providing a convergence analysis for SGD on a rich subset of two-layer feedforward networks with ReLU activations. This subset is characterized by a special structure called "identity mapping". We prove that, if input follows from Gaussian distribution, with standard $O(1/\sqrt{d})$ initialization of the weights, SGD converges to the global minimum in polynomial number of steps. Unlike normal vanilla networks, the "identity mapping" makes our network asymmetric and thus the global minimum is unique. To complement our theory, we are also able to show experimentally that multi-layer networks with this mapping have better performance compared with normal vanilla networks.

Our convergence theorem differs from traditional non-convex optimization techniques. We show that SGD converges to optimal in "two phases": In phase I, the gradient points to the wrong direction, however, a potential function $g$ gradually decreases. Then in phase II, SGD enters a nice one point convex region and converges. We also show that the identity mapping is necessary for convergence, as it moves the initial point to a better place for optimization. Experiment verifies our claims.

## 1 Introduction

Deep learning is the mainstream technique for many machine learning tasks, including image recognition, machine translation, speech recognition, etc. [17]. Despite its success, the theoretical understanding on how it works remains poor. It is well known that neural networks have great expressive power [22, 7, 3, 8, 31]. That is, for every function there exists a set of weights on the neural network such that it approximates the function everywhere. However, it is unclear how to obtain the desired weights. In practice, the most commonly used method is stochastic gradient descent based methods (e.g., SGD, Momentum [40], Adagrad [10], Adam [25]), but to the best of our knowledge, there were no theoretical guarantees that such methods will find good weights.

In this paper, we give the first convergence analysis of SGD for two-layer feedforward network with ReLU activations. For this basic network, it is known that even in the simplified setting where the weights are initialized symmetrically and the ground truth forms orthonormal basis, gradient descent might get stuck at saddle points [41].

Inspired by the structure of residual network (ResNet) [21], we add an extra identity mapping for the hidden layer (see Figure 1). Surprisingly, we show that simply by adding this mapping, with the standard initialization scheme and small step size, SGD always converges to the ground truth. In other words, the optimization becomes significantly easier, after adding the identity mapping. See Figure 2, based on our analysis, the region near the identity matrix $\mathbf{I}$ contains only one global minimum without any saddle points or local minima, thus is easy for SGD to optimize. The role of the identity mapping here, is to move the initial point to this easier region (better initialization).

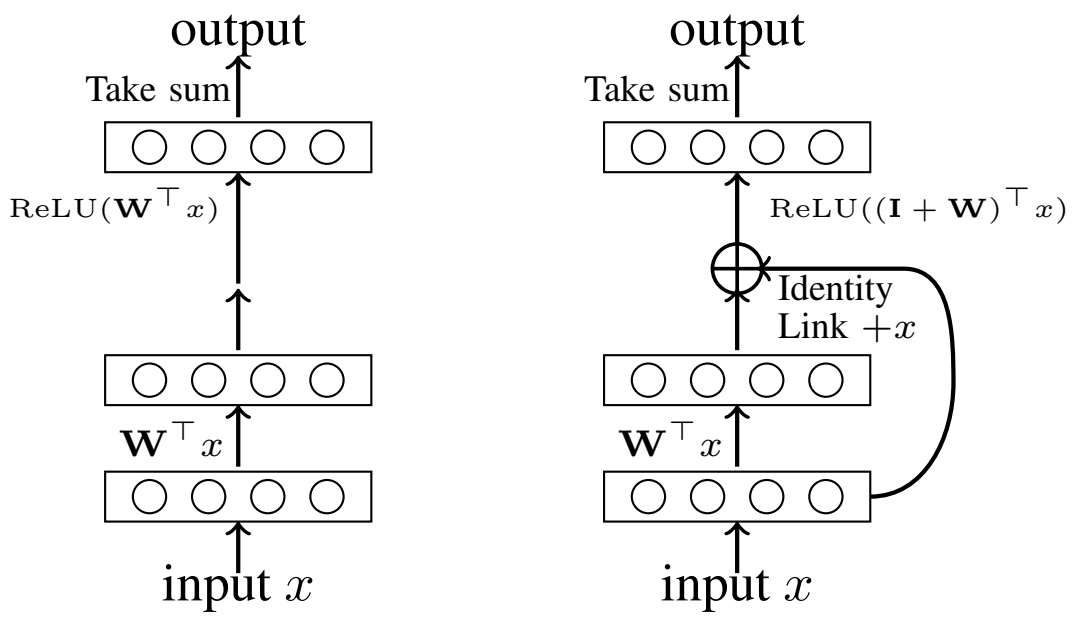

Figure 1: Vanilla network (left), with identity mapping (right)

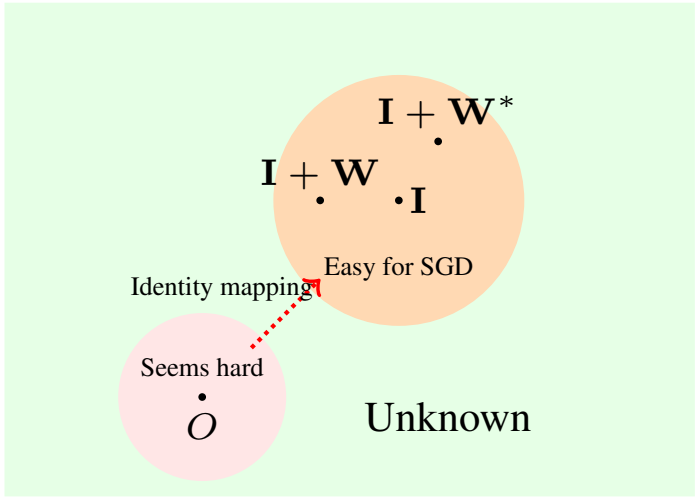

Figure 2: Illustration for our result.

Other than being feedforward and shallow, our network is different from ResNet in the sense that our identity mapping skips one layer instead of two. However, as we will show in Section 5.1, the skip-one-layer identity mapping already brings significant improvement to vanilla networks.

Formally, we consider the following function.

$$f(x, \mathbf{W}) = \|\mathrm{ReLU}((\mathbf{I}+\mathbf{W})^\top x)\|_1 \tag{1}$$

where $\mathrm{ReLU}(v) = \max(v, 0)$ is the ReLU activation function. $x \in \mathbb{R}^d$ is the input vector sampled from a Gaussian distribution, and $\mathbf{W} \in \mathbb{R}^{d\times d}$ is the weight matrix, where $d$ is the number of input units. Notice that $\mathbf{I}$ adds $e_i$ to column $i$ of $\mathbf{W}$, which makes $f$ *asymmetric* in the sense that by switching any two columns in $\mathbf{W}$, we get different functions.

Following the standard setting [34, 41], we assume that there exists a two-layer teacher network with weight $\mathbf{W}^*$. We train the student network using $\ell_2$ loss:

$$\mathsf{L}(\mathbf{W}) = \mathbb{E}_x[(f(x, \mathbf{W}) - f(x, \mathbf{W}^*))^2] \tag{2}$$

We will define a potential function $g$, and show that if $g$ is small, the gradient points to partially correct direction and we get closer to $\mathbf{W}^*$ after every SGD step. However, $g$ could be large and thus gradient might point to the reverse direction. Fortunately, we also show that if $g$ is large, by doing SGD, it will keep decreasing until it is small enough while maintaining the weight $\mathbf{W}$ in a nice region. We call the process of decreasing $g$ as Phase I, and the process of approaching $\mathbf{W}^*$ as Phase II. See Figure 3 and simulations in Section 5.3.

Our two phases framework is fundamentally different from any type of local convergence, as in Phase I, the gradient is pointing to the wrong direction to $\mathbf{W}^*$, so the path from $\mathbf{W}$ to $\mathbf{W}^*$ is non-convex, and SGD takes a long detour to arrive $\mathbf{W}^*$. This framework could be potentially useful for analyzing other non-convex problems.

To support our theory, we have done a few other experiments and got interesting observations. For example, as predicted by our theorem, we found that for multilayer feedforward network with identity mappings, zero initialization performs as good as random initialization. At the first glance, it contradicts the common belief "random initialization is necessary to break symmetry", but actually the identity mapping itself serves as the asymmetric component. See Section 5.4.

Another common belief is that neural network has lots of local minima and saddle points [9], so even if there exists a global minimum, we may not be able to arrive there. As a result, even when the teacher network is shallow, the student network usually needs to be deeper, otherwise it will underfit. However, both our theorem and our experiment show that if the shallow teacher network is in a pretty large region near identity (Figure 2), SGD always converges to the global minimum by initializing the weights $\mathbf{I}+\mathbf{W}$ in this region, with equally shallow student network. By contrast, wrong initialization gets stuck at local minimum and underfit. See Section 5.2.

## Related Work

**Expressivity**. Even two-layer network has great expressive power. For example, two-layer network with sigmoid activations could approximate any continuous function [22, 7, 3]. ReLU is the state-of-the-art activation function [30, 13], and has great expressive power as well [29, 32, 31, 4, 26].

**Learning**. Most previous results on learning neural network are negative [39, 28, 38], or positive but with algorithms other than SGD [23, 43, 37, 14, 15, 16], or with strong assumptions on the model [1, 2]. [35] proved that with high probability, there exists a continuous decreasing path from random initial point to the global minimum, but SGD may not follow this path. Recently, Zhong et al. showed that with initialization point found using tensor decomposition, gradient descent could find the ground truth for one hidden layer network [44].

**Linear network and independent activation**. Some previous works simplified the model by ignoring the activation functions and considering deep linear networks [36, 24] or deep linear residual networks [19], which can only learn linear functions. Some previous results are based on independent activation assumption that the activations of ReLU and the input are independent [5, 24].

**Saddle points**. It is observed that saddle point is not a big problem for neural networks [9, 18]. In general, if the objective is strict-saddle [11], SGD could escape all saddle points.

# 2 Preliminaries

Denote $x$ as the input vector in $\mathbb{R}^d$. For now, we first consider $x$ sampled from normal distribution $\mathcal{N}(0, \mathbf{I})$. Denote $\mathbf{W}^* = (w_1^*, \cdots, w_n^*) \in \mathbb{R}^{d\times d}$ as the weights for the teacher network, $\mathbf{W} = (w_1, \cdots, w_n) \in \mathbb{R}^{d\times d}$ as the weights for the student network, where $w_i^*, w_i \in \mathbb{R}^d$ are column vectors. $f(x, \mathbf{W}^*), f(x, \mathbf{W})$ are defined in (1), representing the teacher and student network.

We want to know whether a randomly initialized $\mathbf{W}$ will converge to $\mathbf{W}^*$, if we run SGD with $l_2$ loss defined in (2). Alternatively, we can write the loss $\mathsf{L}(\mathbf{W})$ as

$$\mathbb{E}_x[(\Sigma_i \mathrm{ReLU}(\langle e_i + w_i, x\rangle) - \Sigma_i \mathrm{ReLU}(\langle e_i + w_i^*, x\rangle))^2]$$

Taking derivative with respect to $w_j$, we get

$$\nabla \mathsf{L}(\mathbf{W})_j = 2\mathbb{E}_x\left[\left(\sum_i \mathrm{ReLU}(\langle e_i + w_i, x\rangle) - \sum_i \mathrm{ReLU}(\langle e_i + w_i^*, x\rangle)\right) x \mathbb{1}_{\langle e_j + w_j, x\rangle \geq 0}\right]$$

where $\mathbb{1}_e$ is the indicator function that equals 1 if the event $e$ is true, and 0 otherwise. Here $\nabla \mathsf{L}(\mathbf{W}) \in \mathbb{R}^{d\times d}$, and $\nabla \mathsf{L}(\mathbf{W})_j$ is its $j$-th column.

Denote $\theta_{i,j}$ as the angle between $e_i + w_i$ and $e_j + w_j$, $\theta_{i^*,j}$ as the angle between $e_i + w_i^*$ and $e_j + w_j$. Denote $\bar{v} = \frac{v}{\|v\|_2}$. Denote $\overline{\mathbf{I} + \mathbf{W}^*}$ and $\overline{\mathbf{I} + \mathbf{W}^*}$ as the column-normalized version of $\mathbf{I} + \mathbf{W}^*$ and $\mathbf{I} + \mathbf{W}$ such that every column has unit norm. Since the input is from a normal distribution, one can compute the expectation inside the gradient as follows.

**Lemma 2.1** (Eqn (13) from [41]). *If $x \sim \mathcal{N}(0, \mathbf{I})$, then* $-\nabla \mathsf{L}(\mathbf{W})_j = \sum_{i=1}^d \left(\frac{\pi}{2}(w_i^* - w_i) + \left(\frac{\pi}{2} - \theta_{i^*,j}\right)(e_i + w_i^*) - \left(\frac{\pi}{2} - \theta_{i,j}\right)(e_i + w_i) + \left(\|e_i + w_i^*\|_2 \sin\theta_{i^*,j} - \|e_i + w_i\|_2 \sin\theta_{i,j}\right)\overline{e_j + w_j}\right)$

**Remark.** Although the gradient of ReLU is not well defined at the point of zero, if we assume input $x$ is from the Gaussian distribution, the loss function becomes smooth, and the gradient is well defined everywhere.

Denote $u \in \mathbb{R}^d$ as the all one vector. Denote $\mathrm{Diag}(\mathbf{W})$ as the diagonal matrix of matrix $\mathbf{W}$, $\mathrm{Diag}(v)$ as a diagonal matrix whose main diagonal equals to the vector $v$. Denote $\text{Off-Diag}(\mathbf{W}) \triangleq \mathbf{W} - \mathrm{Diag}(\mathbf{W})$. Denote $[d]$ as the set $\{1, \cdots, d\}$. Throughout the paper, we abuse the notation of inner product between matrices $\mathbf{W}, \mathbf{W}^*, \nabla \mathsf{L}(\mathbf{W})$, such that $\langle \nabla \mathsf{L}(\mathbf{W}), \mathbf{W}\rangle$ means the summation of the entrywise products. $\|\mathbf{W}\|_2$ is the spectral norm of $\mathbf{W}$, and $\|\mathbf{W}\|_F$ is the Frobenius norm of $\mathbf{W}$. We define the potential function $g$ and variables $g_j, \mathbf{A}_j, \mathbf{A}$ below, which will be useful in the proof.

**Definition 2.2.** *We define the potential function* $g \triangleq \sum_{i=1}^d (\|e_i + w_i^*\|_2 - \|e_i + w_i\|_2)$, *and variable* $g_j \triangleq \sum_{i\neq j}(\|e_i + w_i^*\|_2 - \|e_i + w_i\|_2)$.

**Definition 2.3.** *Denote* $\mathbf{A}_j \triangleq \sum_{i\neq j}((e_i + w_i^*)\overline{e_i + w_i^*}^\top - (e_i + w_i)\overline{e_i + w_i}^\top)$, $\mathbf{A} \triangleq \sum_{i=1}^d ((e_i + w_i^*)\overline{e_i + w_i^*}^\top - (e_i + w_i)\overline{e_i + w_i}^\top) = (\mathbf{I} + \mathbf{W}^*)\overline{\mathbf{I} + \mathbf{W}^*}^\top - (\mathbf{I} + \mathbf{W})\overline{\mathbf{I} + \mathbf{W}}^\top$.

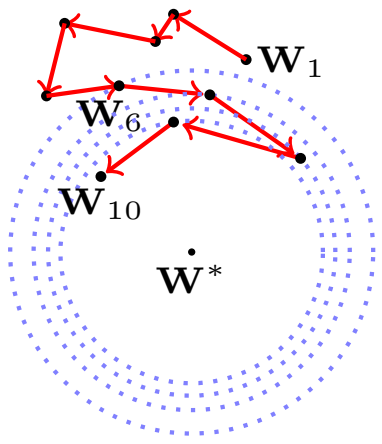

Figure 3: Phase I: $\mathbf{W}_1 \to \mathbf{W}_6$, $\mathbf{W}$ may go to the wrong direction but the potential is shrinking. Phase II: $\mathbf{W}_6 \to \mathbf{W}_{10}$, $\mathbf{W}$ gets closer to $\mathbf{W}^*$ in every step by one point convexity.

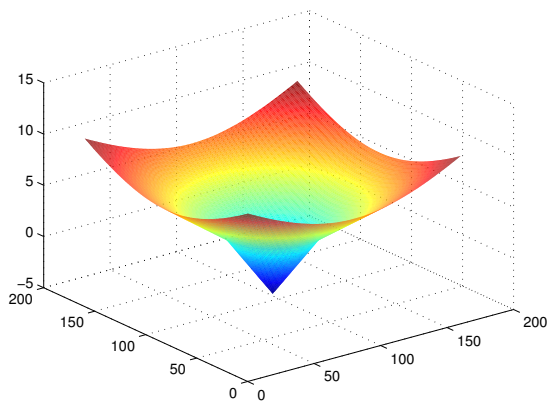

Figure 4: The function is one point strongly convex as every point's negative gradient points to the center, but not convex as any line between the center and the red region is below surface.

In this paper, we consider the standard SGD with mini batch method for training the neural network. Assume $\mathbf{W}_0$ is the initial point, and in step $t > 0$, we have the following updating rule:

$$\mathbf{W}_{t+1} = \mathbf{W}_t - \eta_t \mathbf{G}_t$$

where the stochastic gradient $\mathbf{G}_t = \nabla \mathsf{L}(\mathbf{W}_t) + \mathbf{E}_t$ with $\mathbb{E}[\mathbf{E}_t] = \mathbf{0}$ and $\|\mathbf{E}_t\|_F \leq \varepsilon$. Let $G_2 \triangleq 6d\gamma + \varepsilon, G_F \triangleq 6d^{1.5}\gamma + \varepsilon$, where $\gamma$ is the upper bound of $\|\mathbf{W}^*\|_2$ and $\|\mathbf{W}_0\|_2$ (defined later). As we will see in Lemma C.2, they are the upper bound of $\|\mathbf{G}_t\|_2$ and $\|\mathbf{G}_t\|_F$ respectively.

It's clear that $\mathsf{L}$ is not convex, In order to get convergence guarantees, we need a weaker condition called *one point convexity*.

**Definition 2.4** (One point strongly convexity). *A function $f(x)$ is called $\delta$-one point strongly convex in domain $\mathbf{D}$ with respect to point $x^*$, if $\forall x \in \mathbf{D}$, $\langle -\nabla f(x), x^* - x \rangle > \delta \|x^* - x\|_2^2$.*

By definition, if a function $f$ is strongly convex, it is also one point strongly convex in the entire space with respect to the global minimum. However, the reverse is not necessarily true, e.g., see Figure 4. If a function is one point strongly convex, then in every step a positive fraction of the negative gradient is pointing to the optimal point. As long as the step size is small enough, we will finally arrive the optimal point, possibly by a winding path. See Figure 3 for illustration, where starting from $\mathbf{W}_6$ (Phase II), we get closer to $\mathbf{W}^*$ in every step. Formally, we have the following lemma.

**Lemma 2.5.** *For function $f(\mathbf{W})$, consider the SGD update $\mathbf{W}_{t+1} = \mathbf{W}_t - \eta \mathbf{G}_t$, where $\mathbb{E}[\mathbf{G}_t] = \nabla f(\mathbf{W}_t)$, $\mathbb{E}[\|\mathbf{G}_t\|_F^2] \leq G^2$. Suppose for all $t$, $\mathbf{W}_t$ is always inside the $\delta$-one point strongly convex region with diameter $D$, i.e., $\|\mathbf{W}_t - \mathbf{W}^*\|_F \leq D$. Then for any $\alpha > 0$ and any $T$ such that $T^\alpha \log T \geq \frac{D^2\delta^2}{(1+\alpha)G^2}$, if $\eta = \frac{(1+\alpha)\log T}{\delta T}$, we have $\mathbb{E}\|\mathbf{W}_T - \mathbf{W}^*\|_F^2 \leq \frac{(1+\alpha)\log T G^2}{\delta^2 T}$.*

The proof can be found in Appendix J. Lemma 2.5 uses fixed step size, so it easily fits the standard practical scheme that shrinks $\eta$ by a factor of 10 after every a few epochs. For example, we may apply Lemma 2.5 every time $\eta$ gets changed. Notice that our lemma does not imply that $\mathbf{W}_T$ will converge to $\mathbf{W}^*$. Instead, it only says $\mathbf{W}_T$ will be sufficiently close to $\mathbf{W}^*$ with small step size $\eta$.

# 3 Main Theorem

**Theorem 3.1** (Main Theorem). *There exists* constants $\gamma > \gamma_0 > 0$ *such that If $x \sim \mathcal{N}(0, \mathbf{I})$, $\|\mathbf{W}_0\|_2, \|\mathbf{W}^*\|_2 \leq \gamma_0$, $d \geq 100$, $\varepsilon \leq \gamma^2$, then SGD for $\mathsf{L}(\mathbf{W})$ will find the ground truth $\mathbf{W}^*$ by two phases. In Phase I, by setting $\eta \leq \frac{\gamma^2}{G_2^2}$, the potential function will keep decreasing until it is smaller than $197\gamma^2$, which takes at most $\frac{1}{16\eta}$ steps. In Phase II, for any $\alpha > 0$ and any $T$ such that $T^\alpha \log T \geq \frac{36d}{100^4(1+\alpha)G_F^2}$, if we set $\eta = \frac{(1+\alpha)\log T}{\delta T}$, we have $\mathbb{E}\|\mathbf{W}_T - \mathbf{W}^*\|_F^2 \leq \frac{100^2(1+\alpha)\log T G_F^2}{9T}$.*

**Remarks.** Randomly initializing the weights with $O(1/\sqrt{d})$ is standard in deep learning, see [27, 12, 20]. It is also well known that if the entries are initialized with $O(1/\sqrt{d})$, the spectral norm of the random matrix is $O(1)$ [33]. So our result matches with the common practice. Moreover, as we will show in Section 5.5, networks with small average spectral norm already have good performance. Thus, our assumption $\|\mathbf{W}^*\|_2 = O(1)$ is reasonable. Notice that here we assume the *spectral norm* of $\mathbf{W}^*$ to be constant, which means the Frobenius norm $\|\mathbf{W}^*\|_F$ could be as big as $O(\sqrt{d})$.

The assumption that the input follows a Gaussian distribution is not necessarily true in practice (Although this is a common assumption appeared in the previous papers [5, 41, 42], and also considered plausible in [6]). We could easily generalize the analysis to rotation invariant distributions, and potentially more general distributions (see Section 6). Moreover, previous analyses either ignore the nonlinear activations and thus consider linear model [36, 24, 19], or directly [5, 24] or indirectly [41][1] assume that the activations are independent. By contrast, in our model the ReLU activations are highly correlated[2] as $\|\mathbf{W}\|_2, \|\mathbf{W}^*\|_2 = \Omega(1)$. As pointed out by [6], eliminating the unrealistic assumptions on activation independence is the central problem of analyzing the loss surface of neural network, which was not fully addressed by the previous analyses.

To prove the main theorem, we split the process and present the following two theorems, which will be proved in Appendix C and D.

**Theorem 3.2** (Phase I)**.** *There exists a constant* $\gamma > \gamma_0 > 0$ *such that If* $\|\mathbf{W}_0\|_2, \|\mathbf{W}^*\|_2 \leq \gamma_0$, $d \geq 100$, $\eta \leq \frac{\gamma^2}{G_2^2}$, $\varepsilon \leq \gamma^2$*, then* $g_t$ *will keep decreasing by a factor of* $1 - 0.5\eta d$ *for every step, until* $g_{t_1} \leq 197\gamma^2$ *for step* $t_1 \leq \frac{1}{16\eta}$*. After that, Phase II starts. That is, for every* $T > t_1$*, we have* $\|\mathbf{W}_T\|_2 \leq \frac{1}{100}$ *and* $g_T \leq 0.1$.

**Theorem 3.3** (Phase II)**.** *There exists a constant* $\gamma$ *such that if* $\|\mathbf{W}\|_2, \|\mathbf{W}^*\|_2 \leq \gamma$*, and* $g \leq 0.1$*, then* $\langle -\nabla \mathsf{L}(\mathbf{W}), \mathbf{W}^* - \mathbf{W}\rangle = \sum_{j=1}^{d} \langle -\nabla \mathsf{L}(\mathbf{W})_j, w_j^* - w_j\rangle > 0.03\|\mathbf{W}^* - \mathbf{W}\|_F^2$.

With these two theorems, we get the main theorem immediately.

*Proof for Theorem 3.1.* By Theorem 3.2, we know the statement for Phase I is true, and we will enter phase II in $\frac{1}{16\eta}$ steps. After entering Phase II, based on Theorem 3.3, we simply use Lemma 2.5 by setting $\delta = 0.03$, $D = \frac{\sqrt{d}}{50}, G = G_F$ to get the convergence guarantee. □

# 4 Overview of the Proofs

**General Picture**. In many convergence analyses for non-convex functions, one would like to show that $\mathsf{L}$ is one point strongly convex, and directly apply Lemma 2.5 to get the convergence result. However, this is not true for 2-layer neural network, as the gradient may point to the wrong direction, see Section 5.3.

So when is our $\mathsf{L}$ one point convex? Consider the following thought experiment: First, suppose $\|\mathbf{W}\|_2, \|\mathbf{W}^*\|_2 \to 0$, we know $\|w_i\|_2, \|w_i^*\|_2$ also go to 0. Thus, $e_i + w_i$ and $e_i + w_i^*$ are close to $e_i$. As a result, $\theta_{i,j}, \theta_{i^*,j} \approx \frac{\pi}{2}$, and $\theta_{i^*,i} \approx 0$. Based on Lemma 2.1, this gives us a naïve approximation of the negative gradient, i.e., $-\nabla \mathsf{L}(\mathbf{W})_j \approx \frac{\pi}{2}(w_j^* - w_j) + \frac{\pi}{2}\sum_{i=1}^{d}(w_i^* - w_i) + \overline{e_j + w_j}\sum_{i\neq j}(\|e_i + w_i^*\|_2 - \|e_i + w_i\|_2)$ .

While the first two terms $\frac{\pi}{2}(w_j^* - w_j)$ and $\frac{\pi}{2}\sum_{i=1}^{d}(w_i^* - w_i)$ have positive inner product with $\mathbf{W}^* - \mathbf{W}$, the last term $g_j = \overline{e_j + w_j}\sum_{i\neq j}(\|e_i + w_i^*\|_2 - \|e_i + w_i\|_2)$ can point to arbitrary direction. If the last term is small, it can be covered by the first two terms, and $\mathsf{L}$ becomes one point strongly convex. So we define a potential function closely related to the last term: $g = \sum_{i=1}^{d}(\|e_i + w_i^*\|_2 - \|e_i + w_i\|_2)$. We show that if $g$ is small enough, $\mathsf{L}$ is also one point strongly convex (Theorem 3.3).

However, from random initialization, $g$ can be as large as of $\Omega(\sqrt{d})$, which is too big to be covered. Fortunately, we show that if $g$ is big, it will gradually decrease simply by doing SGD on $\mathsf{L}$. More specifically, we introduce a *two phases* convergence analysis framework:

1. In Phase I, the potential function $g$ is decreasing to a small value.
2. In Phase II, $g$ remains small, so $\mathsf{L}$ is one point convex and thus $\mathbf{W}$ starts to converge to $\mathbf{W}^*$.

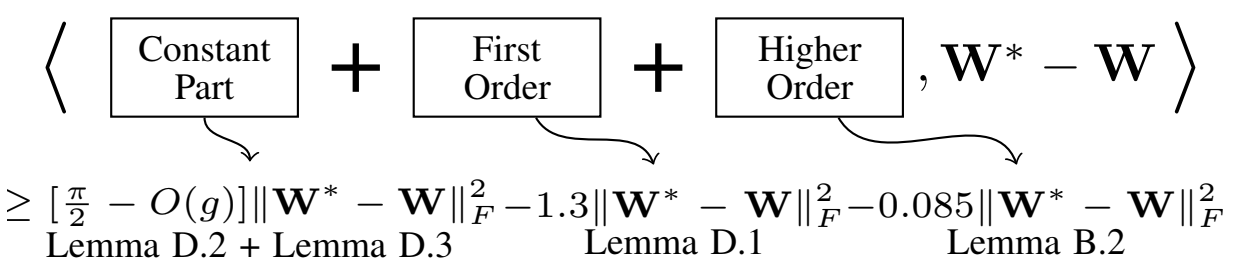

Figure 5: Lower bounds of inner product using Taylor expansion

We believe that this framework could be helpful for other non-convex problems.

**Technical difficulty: Phase I**. Our key technical challenge is to show that in Phase I, the potential function actually decreases to $O(1)$ after polynomial number of iterations. However, we cannot show this by merely looking at $g$ itself. Instead, we introduce an auxiliary variable $s = (\mathbf{W}^* - \mathbf{W})u$, where $u$ is the all one vector. By doing a careful calculation, we get their joint update rules (Lemma C.3 and Lemma C.4):

$$\begin{cases} s_{t+1} & \approx s_t - \frac{\pi\eta d}{2} s_t + \eta O(\sqrt{d} g_t + \sqrt{d}\gamma) \\ g_{t+1} & \approx g_t - \eta d g_t + \eta O(\gamma\sqrt{d}\|s_t\|_2 + d\gamma^2) \end{cases}$$

Solving this dynamics, we can show that $g_t$ will approach to (and stay around) $O(\gamma)$, thus we enter Phase II.

**Technical difficulty: Phase II**. Although the overall approximation in the thought experiment looks simple, the argument is based on an over simplified assumption that $\theta_{i^*,j}, \theta_{i,j} \approx \frac{\pi}{2}$ for $i \neq j$. However, when $\mathbf{W}^*$ has constant spectral norm, even when $\mathbf{W}$ is very close to $\mathbf{W}^*$, $\theta_{i,j^*}$ could be constantly far away from $\frac{\pi}{2}$, which prevents us from applying this approximation directly. To get a formal proof, we use the standard Taylor expansion and control the higher order terms. Specifically, we write $\theta_{i^*,j}$ as $\theta_{i^*,j} = \arccos\langle \overline{e_i + w_i^*}, \overline{e_j + w_j} \rangle$ and expand $\arccos$ at point 0, thus,

$$\theta_{i^*,j} = \frac{\pi}{2} - \langle \overline{e_i + w_i^*}, \overline{e_j + w_j} \rangle + O(\langle \overline{e_i + w_i^*}, \overline{e_j + w_j} \rangle^3)$$

However, even when $\mathbf{W} \approx \mathbf{W}^*$, the higher order term $O(\langle \overline{e_i + w_i^*}, \overline{e_j + w_j} \rangle^3)$ still can be as large as a constant, which is too big for us. Our trick here is to consider the "joint Taylor expansion":

$$\theta_{i^*,j} - \theta_{i,j} = \langle \overline{e_i + w_i} - \overline{e_i + w_i^*}, \overline{e_j + w_j} \rangle + O(|\langle \overline{e_i + w_i^*}, \overline{e_j + w_j} \rangle^3 - \langle \overline{e_i + w_i}, \overline{e_j + w_j} \rangle^3|)$$

As $\mathbf{W}$ approaches $\mathbf{W}^*$, $|\langle \overline{e_i + w_i^*}, \overline{e_j + w_j} \rangle^3 - \langle \overline{e_i + w_i}, \overline{e_j + w_j} \rangle^3|$ also tends to zero, therefore our approximation has bounded error.

In the thought experiment, we already know that the constant part in the Taylor expansion of $\nabla \mathsf{L}(\mathbf{W})$ is $\frac{\pi}{2} - O(g)$-one point convex. We show that after taking inner product with $\mathbf{W}^* - \mathbf{W}$, the first order terms are lower bounded by (roughly) $-1.3\|\mathbf{W}^* - \mathbf{W}\|_F^2$ and the higher order terms are lower bounded by $-0.085\|\mathbf{W}^* - \mathbf{W}\|_F^2$. Adding them together, we can see that $\mathsf{L}(\mathbf{W})$ is one point convex as long as $g$ is small. See Figure 5.

**Geometric Lemma**. In order to get through the whole analysis, we need tight bounds on a few common terms that appear everywhere. Instead of using naïve algebraic techniques, we come up with a nice geometric proof to get nearly optimal bounds. Due to space limit, we defer it to Appendix E.

# 5 Experiments

In this section, we present several simulation results to support our theory. Our code can be found in the supplementary materials.

## 5.1 Importance of identity mapping

In this experiment, we compare the standard ResNet [21] and *single skip model* where identity mapping skips only one layer. See Figure 6 for the single skip model. We also ran the vanilla network, where the identity mappings are completely removed.

[1] They assume input is Gaussian and the $\mathbf{W}^*$ is orthonormal, which means the activations are independent in teacher network.

[2] Let $\sigma_i$ be the output of i-th ReLU unit, then in our setting, $\sum_{i,j} \text{Cov}[\sigma_i, \sigma_j]$ can be as large as $\Omega(d)$, which is far from being independent.

Table 1: Test error of three 56-layer networks on Cifar-10

| | ResNet | Single skip | Vanilla |
|---|---|---|---|
| Test Err | 6.97% | 9.01% | 12.04% |

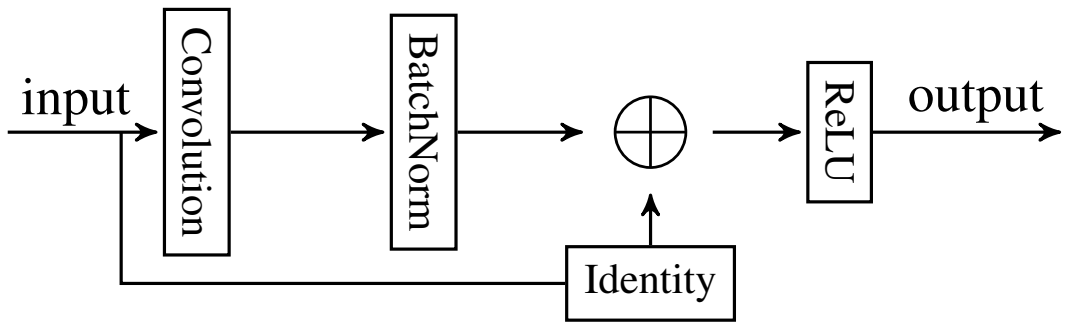

Figure 6: Illustration of one block in single skip model in Sec 5.1

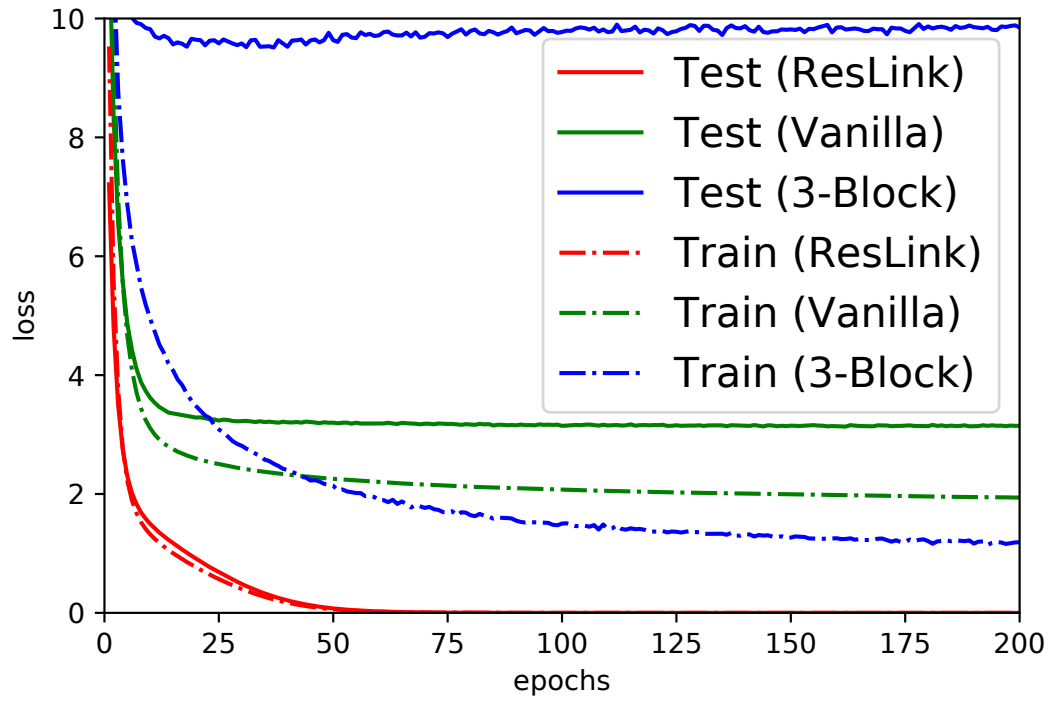

(a) Test Error, Train Error

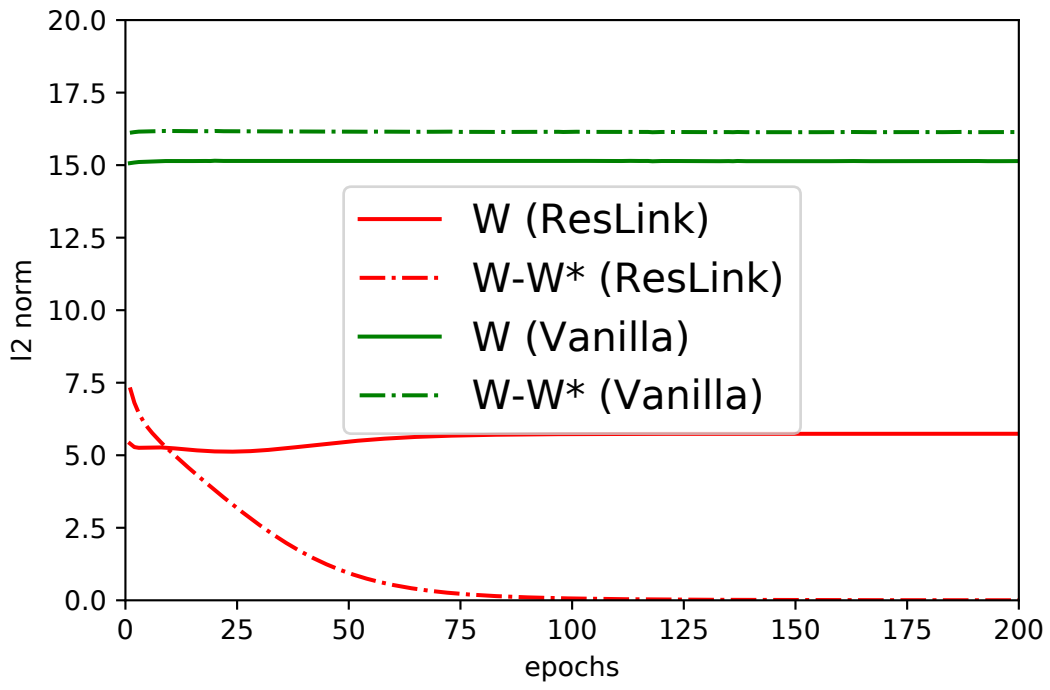

(b) $\|\mathbf{W}^* - \mathbf{W}\|_F, \|\mathbf{W}\|_F$

Figure 7: Verifying the global convergence

In this experiment, we choose Cifar-10 as the dataset, and all the networks have 56-layers. Other than the identity mappings, all other settings are identical and default. We run the experiments for 5 times and report the average test error. As we can see in Table 1, compared with vanilla network, by simply using a single skip identity mapping, one can already improve the test error by $3.03\%$, and is $2.04\%$ close to the ResNet. So single skip identity mapping brings significant improvement on test accuracy.

## 5.2 Global minimum convergence

In this experiment, we verify our main theorem that for two-layer teacher network and student network with identity mappings, as long as $\|\mathbf{W}_0\|_2, \|\mathbf{W}^*\|_2$ is small, SGD always converges to the global minimum $\mathbf{W}^*$, thus gives almost 0 training error and test error. We consider three student networks. The first one (ResLink) is defined using (2), the second one (Vanilla) is the same model without the identity mapping. The last one (3-Block) is a three block network with each block containing a linear layer (500 hidden nodes), a batch normalization and a ReLU layer. The teacher network always shares the same structure as the student network.

The input dimension is $100$. We generated a fixed $\mathbf{W}^*$ for all the trials with $\|\mathbf{W}^*\|_2 \approx 0.6, \|\mathbf{W}^*\|_F \approx 5.7$. We generated a training set of size $100,000$, and test set of size $10,000$, sampled from a Gaussian distribution. We use batch size 200, step size $0.001$. We run ResLink for 5 times with random initialization ($\|\mathbf{W}\|_2 \approx 0.6$ and $\|\mathbf{W}\|_F \approx 5$), and plot the curves by taking the average.

Figure 7(a) shows test error and training error of the three networks. Comparing Vanilla with 3-Block, we find that 3-Block is more expressive, so its training error is smaller compared with vanilla network; but it suffers from overfitting and has bigger test error. This is the standard overfitting vs underfitting tradeoff. Surprisingly, with only one hidden layer, ResLink has both zero test error and training error. If we look at Figure 7(b), we know the distance between $\mathbf{W}$ and $\mathbf{W}^*$ converges to 0, meaning ResLink indeed finds the global optimal in all 5 trials. By contrast, for vanilla network, which is essentially the same network with different initialization, $\|\mathbf{W} - \mathbf{W}^*\|_2$ does not converge to zero[3]. This is exactly what our theory predicted.

[3]To make comparison meaningful, we set $\mathbf{W} - \mathbf{I}$ to be the actual weight for Vanilla as its identity mapping is missing, which is why it has a much bigger initial norm.

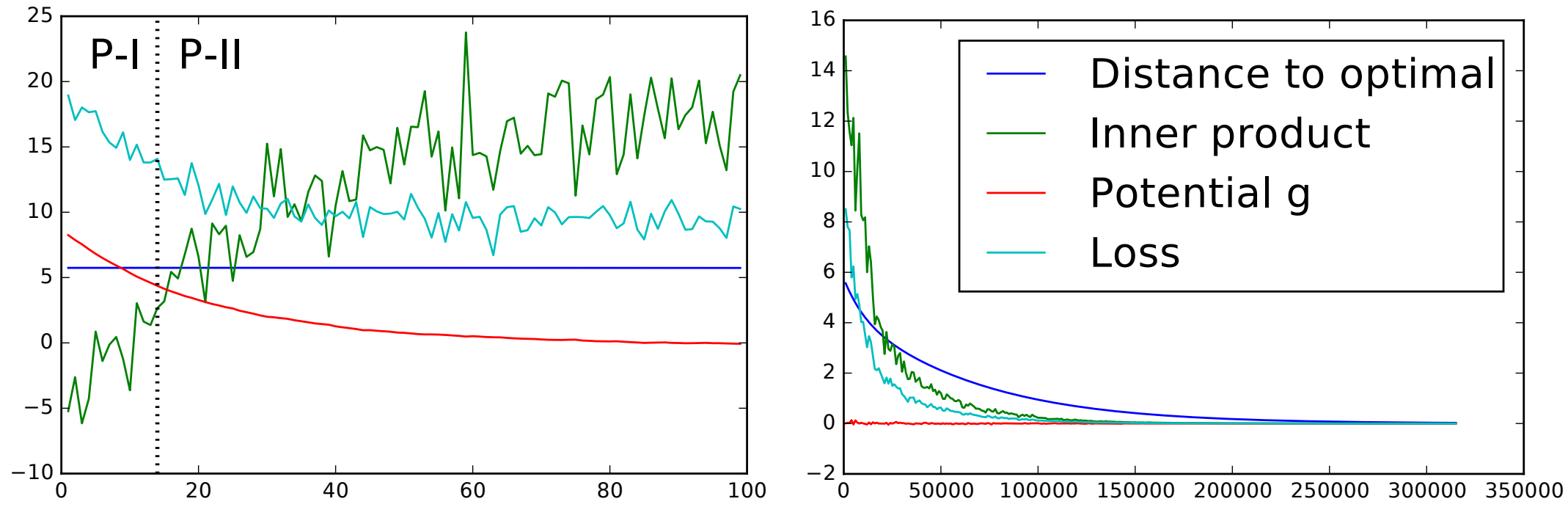

(a) First 100 iterations

(b) The entire process

Figure 8: Verifying the dynamics

## 5.3 Verify the dynamics

In this experiment, we verify our claims on the dynamics. Based on the analysis, we construct a $1500 \times 1500$ matrix $\mathbf{W}$ s.t. $\|\mathbf{W}\|_2 \approx 0.15$, $\|\mathbf{W}\|_F \approx 5$ , and set $\mathbf{W}^* = 0$. By plugging them into (2), one can see that even in this simple case that $\mathbf{W}^* = 0$, initially the gradient is pointing to the wrong direction, i.e., not one point convex. We then run SGD on $\mathbf{W}$ by using samples $x$ from Gaussian distribution, with batch size 300, step size 0.0001.

Figure 8(a) shows the first 100 iterations. We can see that initially the inner product defined in Definition 2.4 is negative, then after about 15 iterations, it turns positive, which means $\mathbf{W}$ is in the one point strongly convex region. At the same time, the potential $g$ keeps decreasing to a small value, while the distance to optimal (which also equals to $\|\mathbf{W}\|_F$ in this experiment) is not affected. They precisely match with our description of Phase I in Theorem 3.2.

After that, we enter Phase II and slowly approach to $\mathbf{W}^*$, see Figure 8(b). Notice that the potential $g$ is always very small, the inner product is always positive, and the distance to optimal is slowly decreasing. Again, they precisely match with our Theorem 3.3.

## 5.4 Zero initialization works

In this experiment, we used a simple 5-block neural network on MNIST, where every block contains a $784 * 784$ feedforward layer, an identity mapping, and a ReLU layer. Cross entropy criterion is used. We compare zero initialization with standard $O(1/\sqrt{d})$ random initialization. We found that for zero initialization, we can get $1.28\%$ test error, while for random initialization, we can get $1.27\%$ test error. Both results were obtained by taking average among 5 runs and use step size $0.1$, batch size $256$. If the identity mapping is removed, zero initialization no longer works.

## 5.5 Spectral norm of $\mathbf{W}^*$

We also applied the exact model $f$ defined in (1) to distinguish two classes in MNIST. For any input image $x$, We say it's in class A if $f(x, \mathbf{W}) < T_{A,B}$, and in class B otherwise. Here $T_{A,B}$ is the optimal threshold for the function $f(x, \mathbf{0})$ to distinguish $A$ and $B$. If $\mathbf{W} = \mathbf{0}$, we get $7\%$ training error for distinguish class 0 and class 1. However, it can be improved to $1\%$ with $\|\mathbf{W}\|_2 = 0.6$. We tried this experiment for all possible 45 pairs of classes in MNIST, and improve the average training error from $34\%$ (using $\mathbf{W} = \mathbf{0}$) to $14\%$ (using $\|\mathbf{W}\|_2 = 0.6$). Therefore our model with $\|\mathbf{W}\|_2 = \Omega(1)$ has reasonable expressive power, and is substantially different from just using the identity mapping alone.

## 6 Discussions

The assumption that the input is Gaussian can be relaxed in several ways. For example, when the distribution is $\mathcal{N}(0, \Sigma)$ where $\|\Sigma - \mathbf{I}\|_2$ is bounded by a small constant, the same result holds with slightly worse constants. Moreover, since the analysis relies Lemma 2.1, which is proved by converting the original input space into polar space, it is easy to generalize the calculation to rotation invariant distributions. Finally, for more general distributions, as long as we could explicitly compute the expectation, which is in the form of $O(\mathbf{W}^* - \mathbf{W})$ plus certain potential function, our analysis framework may also be applied.

There are many exciting open problems. For example, Our paper is the first one that gives solid SGD analysis for neural network with nonlinear activations, without unrealistic assumptions like independent activation assumption. It would be great if one could further extend it to multiple layers, which would be a major breakthrough of understanding optimization for deep learning. Moreover, our two phase framework could be applied to other non-convex problems as well.

## Acknowledgement

The authors want to thank Robert Kleinberg, Kilian Weinberger, Gao Huang, Adam Klivans and Surbhi Goel for helpful discussions, and the anonymous reviewers for their comments.

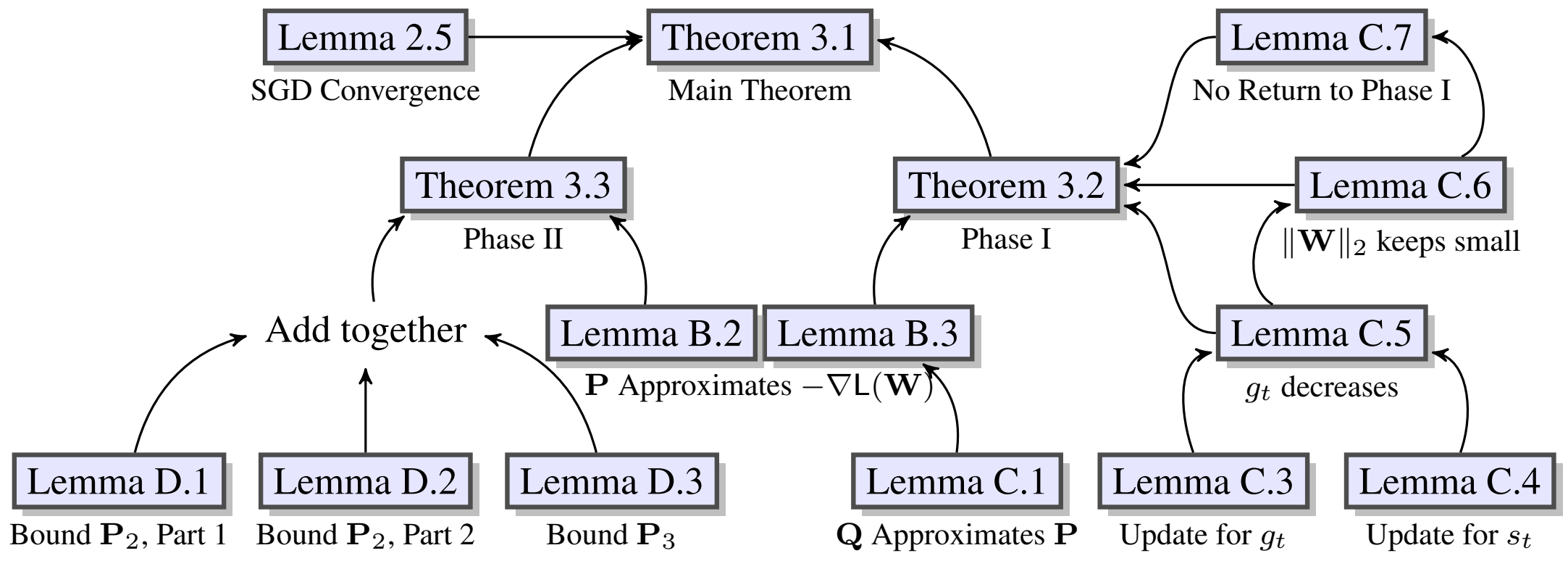

Figure 9: Flowchart of the proofs

# A Flowchart of the proofs

Although the proofs of our theorems are intricate, many lemmas have clear intuition behind the statement. Therefore, we add "*" to these lemmas, so that time constrained readers could feel confident to skip the proofs. We also plot a flowchart of the proofs in Figure 9 to help the readers spend time wisely.

Since the proofs are long and complicated, we choose to present them in a top-down way. That is, we present the main theorems (Theorem 3.1, Theorem 3.2, and Theorem 3.3) in the main paper, and then present the necessary lemmas in order to prove those main theorems in Section B, Section C and Section D. Finally, we present the proofs for those lemma in Section G, Section H and Section I, respectively.

# B Compute Approximation Matrix

The exact form of $-\nabla \mathsf{L}(\mathbf{W})_j$ in Lemma 2.1 contains variables like $\theta_{i^*,j}, \theta_{i,j}, \sin\theta_{i^*,j}, \sin\theta_{i,j}$, which are hard to deal with. In this section, we compute the approximation of these terms using Taylor series, and show that the approximation loss is minor. While the proofs are technically involved, the claims themselves are not surprising. Hence, we encourage the readers to skip the proofs (Appendix G) for the first reading.

Define the $j$-th column of the approximation matrix $\mathbf{P}$ as follows. See Definition 2.2 and Definition 2.3 for $g_j, \mathbf{A}_j$.

$$\begin{aligned}
\mathbf{P}_j &\triangleq \mathbf{P}_{1,j} + \mathbf{P}_{2,j} + \mathbf{P}_{3,j}, \quad \text{where} \\
\mathbf{P}_{1,j} &\triangleq \sum_{i=1}^{d} \frac{\pi}{2}(w_i^* - w_i), \\
\mathbf{P}_{2,j} &\triangleq g_j \overline{e_j + w_j} + \left(\mathbf{I} - \frac{1}{2}\overline{e_j + w_j} \cdot \overline{e_j + w_j}^\top\right) \mathbf{A}_j \overline{e_j + w_j}, \\
\mathbf{P}_{3,j} &\triangleq \left(\frac{\pi}{2} - \theta_{j^*,j}\right)(e_j + w_j^*) - \frac{\pi}{2}(e_j + w_j) + \|e_j + w_j^*\| \sin\theta_{j^*,j} \overline{e_j + w_j}.
\end{aligned}$$

Treat $\mathbf{P}_{1,j}, \mathbf{P}_{2,j}, \mathbf{P}_{3,j}$ as $j$-th column of matrix $\mathbf{P}_1, \mathbf{P}_2, \mathbf{P}_3$ respectively, we have $\mathbf{P} = \mathbf{P}_1 + \mathbf{P}_2 + \mathbf{P}_3$. Although $\mathbf{P}$ depends on $\mathbf{W}$, we abuse the notation and simply write $\mathbf{P}$.

**Claim B.1.** *$\mathbf{P}_j$ approximates $-\nabla \mathsf{L}(\mathbf{W})_j$ by setting $(\frac{\pi}{2} - \theta_{i,j}) \approx \langle \overline{e_i + w_i}, \overline{e_j + w_j} \rangle$, $(\frac{\pi}{2} - \theta_{i^*,j}) \approx \langle \overline{e_i + w_i^*}, \overline{e_j + w_j} \rangle$, $\sin\theta_{i,j} \approx 1 - \frac{1}{2}\langle \overline{e_i + w_i}, \overline{e_j + w_j} \rangle^2$ and $\sin\theta_{i^*,j} \approx 1 - \frac{1}{2}\langle \overline{e_i + w_i^*}, \overline{e_j + w_j} \rangle^2$.*

Below we show that the approximation loss is negligible in terms of one point convexity and spectral norm.

**Lemma* B.2.** *If $\|\mathbf{W}\|_2, \|\mathbf{W}^*\|_2 \le \gamma \le \frac{1}{100}$, $|\langle \mathbf{P} + \nabla \mathsf{L}(\mathbf{W}), \mathbf{W}^* - \mathbf{W} \rangle| < 0.085 \|\mathbf{W}^* - \mathbf{W}\|_F^2$.*

**Lemma* B.3.** *If $\|\mathbf{W}\|_2, \|\mathbf{W}^*\|_2 \le \gamma \le \frac{1}{100}$, $\|\mathbf{P} + \nabla \mathsf{L}(\mathbf{W})\|_2 \le 3.5\gamma^2$.*

# C Phase I: The Decreasing Potential Function

As we saw in Theorem 3.3, if $\|\mathbf{W}\|_2, \|\mathbf{W}^*\|_2$ is bounded by a constant $\gamma = \frac{1}{100}$, and the potential function $g \leq 0.1$, $\mathsf{L}(\mathbf{W})$ is 0.03-one point convex, which will give us convergence guarantee according to Lemma 2.5. However, $g$ could be larger than 0.1 initially, and as we run SGD, $\|\mathbf{W}\|_2$ might be larger than $\frac{1}{100}$ as well.

In this section, we address both problems by analyzing the dynamics of SGD, thus prove Theorem 3.2. The proofs can be found in Appendix H. Before proceeding to the interesting stuff, we need a simpler form of $\nabla \mathsf{L}(\mathbf{W})$ to work with, see below.

**Lemma C.1.** *If $\|\mathbf{W}\|_2, \|\mathbf{W}^*\|_2 \leq \gamma \leq \frac{1}{100}$, the negative gradient of $\mathsf{L}(\mathbf{W})$ is approximately*

$$\mathbf{Q}(\mathbf{W}) \triangleq \frac{\pi}{2}(\mathbf{W}^* - \mathbf{W})\left(\mathbf{I} + uu^\top\right) + (\mathbf{W}^* - \mathbf{W})^\top - 2\mathrm{Diag}(\mathbf{W}^* - \mathbf{W}) + g\overline{\mathbf{I} + \mathbf{W}}$$

*where $u$ is the all 1 vector. The approximation error is $\|\mathbf{Q}(\mathbf{W}) - [-\nabla \mathsf{L}(\mathbf{W})]\|_2 \leq 61\gamma^2$.*

We immediately get the bound of the gradient norm.

**Lemma* C.2.** *If $\|\mathbf{W}\|_2, \|\mathbf{W}^*\|_2 \leq \gamma \leq \frac{1}{100}$, $\|\nabla \mathsf{L}(\mathbf{W})\|_2 \leq 6d\gamma$.*

Now we are ready to analyze the dynamics. We use subscript $t$ under each variable to denote its value at the step $t$. For simplicity, let $\mathbf{Q}_t \triangleq \mathbf{Q}(\mathbf{W}_t)$. Define $s_t \triangleq (\mathbf{W}^* - \mathbf{W}_t)u$. We first compute the updating rule for $g_t$.

**Lemma C.3.** *If $\|\mathbf{W}_t\|_2, \|\mathbf{W}^*\|_2 \leq \gamma \leq \frac{1}{100}$, $d \geq 100$, $\eta \leq \frac{\gamma^2}{G_2^2}$, then $|g_{t+1}| \leq (1 - 0.95\eta d)|g_t| + 86\eta d\gamma^2 + 1.03\eta\sqrt{d}\varepsilon + 4.8\eta\|s_t\|_2\gamma\sqrt{d}$.*

The bound contains $\|s_t\|_2$ which could be large, so we also need to compute its updating rule:

**Lemma C.4.** *If $\|\mathbf{W}_t\|_2, \|\mathbf{W}^*\|_2 \leq \gamma \leq \frac{1}{100}$, then $\|s_{t+1}\|_2 \leq \left(1 - \eta\frac{(d+1)\pi}{2}\right)\|s_t\|_2 + \eta(6.61\gamma + 1.03|g_t| + \varepsilon)\sqrt{d}$.*

Combining the two lemmas, we are ready to show that $g_t$ will shrink, conditioned on that $\|\mathbf{W}_t\|_2$ is bounded by $\gamma$.

**Lemma C.5.** *If for every step $t > 0$, $\|\mathbf{W}_t\|_2, \|\mathbf{W}^*\|_2 \leq \gamma \leq \frac{1}{100}$,$d \geq 100$, $\eta \leq \frac{\gamma^2}{G_2^2}$, $\varepsilon \leq \gamma^2$, then $|g_t|$ will keep decreasing by a factor of $1 - 0.5\eta d$ for every step, until $|g_{t_1}| \leq 197\gamma^2$ for $t_1 \leq \frac{1}{16\eta}$.*

Fortunately, we also know that $\|\mathbf{W}_t\|_2$ is always bounded by $\gamma$ during the process described in Lemma C.5.

**Lemma C.6.** *There exists a constant $\gamma > \gamma_0 > 0$ such that if $\|\mathbf{W}_0\|_2, \|\mathbf{W}^*\|_2 \leq \gamma_0$, $d \geq 100$, $\eta \leq \frac{\gamma^2}{G_2^2}$, $\varepsilon \leq \gamma^2$, then in the process of Phase I (Lemma C.5), we always have $\|\mathbf{W}_T\|_2 \leq \gamma \leq \frac{1}{100}$ for any $T > 0$.*

Now, we are at the state where $|g_t|$ is small, and $\|\mathbf{W}_T\|_2 \leq \gamma$, which means we are in Phase II. The next lemma ensures that we will stay in Phase II forever.

**Lemma C.7.** *There exists a constant $\gamma_0 > \gamma > 0$ such that if $\|\mathbf{W}_0\|_2, \|\mathbf{W}^*\|_2 \leq \gamma_0$, $d \geq 100$, $\eta \leq \frac{\gamma^2}{G_2^2}$, $\varepsilon \leq \gamma^2$, then after $|g_{t_1}| \leq 197\gamma^2$, Phase I ends and Phase II starts. That is, for every $T > t_1$, $\|\mathbf{W}_T\|_2 \leq \gamma$ and $|g_T| \leq 0.1$.*

*Proof for Theorem 3.2.* We immediately get Theorem 3.2 by combining the above three lemmas. They show that $g_t$ will decrease to a small value in Phase I (Lemma C.5), $\|\mathbf{W}_t\|_2$ will keep small during this process (Lemma C.6), and they all keep small afterwards (Lemma C.7). □

# D Phase II: One Point Convexity

In this section, we prove Theorem 3.3. See detailed proofs in Appendix I. Using Lemma B.2, it suffices to bound

$$\langle \mathbf{P}, \mathbf{W}^* - \mathbf{W} \rangle = \sum_{j=1}^{d} \langle \mathbf{P}_{1,j} + \mathbf{P}_{2,j} + \mathbf{P}_{3,j}, w_j^* - w_j \rangle$$

Here the first term is easy to calculate.

$$\sum_{j=1}^{d} \langle \mathbf{P}_{1,j}, w_j^* - w_j \rangle = \frac{\pi}{2} \left\| \sum_{i=1}^{d} (w_i^* - w_i) \right\|_2^2 \geq 0 \tag{3}$$

For notational simplicity, denote

$$x_j \triangleq \left( \overline{e_j + w_j} \cdot \overline{e_j + w_j}^\top \right) (w_j^* - w_j),$$

$$\mathbf{X} \triangleq (x_1, \cdots, x_d) \tag{4}$$

$$z_j \triangleq \left( \mathbf{I} - \frac{1}{2} \overline{e_j + w_j} \cdot \overline{e_j + w_j}^\top \right) (w_j^* - w_j) \tag{5}$$

By Definition of $\mathbf{P}_{2,j}$ and (5), we have

$$\sum_{j=1}^{d} \langle \mathbf{P}_{2,j}, w_j^* - w_j \rangle = \sum_{j=1}^{d} \left\langle g_j \overline{e_j + w_j}, w_j^* - w_j \right\rangle + \sum_{j=1}^{d} z_j^\top \mathbf{A}_j \overline{e_j + w_j} \tag{6}$$

We bound the above two terms separately below.

**Lemma D.1.** *If* $\|\mathbf{W}\|_2, \|\mathbf{W}^*\|_2 \leq \gamma \leq \frac{1}{100}$, *then*

$$\sum_{j=1}^{d} z_j^\top \mathbf{A}_j \overline{e_j + w_j} \geq -(1.3 + 8\gamma) \|\mathbf{W}^* - \mathbf{W}\|_F^2 + \|\mathbf{W}^* - \mathbf{W}\|_F \|\mathbf{X}\|_F.$$

**Lemma D.2.** *If* $\|\mathbf{W}\|_2, \|\mathbf{W}^*\|_2 \leq \gamma \leq \frac{1}{100}$, *then*

$$\sum_{j=1}^{d} \langle g_j \overline{e_j + w_j}, w_j^* - w_j \rangle \geq -\|\mathbf{W}^* - \mathbf{W}\|_F \|\mathbf{X}\|_F - \frac{(1+\gamma) g \|\mathbf{W}^* - \mathbf{W}\|_F^2}{2(1 - 2\gamma)}$$

It remains to bound $\sum_{j=1}^{d} \langle \mathbf{P}_{3,j}, w_j^* - w_j \rangle$. We have the following lemma.

**Lemma D.3.** *If* $\|\mathbf{W}\|_2, \|\mathbf{W}^*\|_2 \leq \gamma \leq \frac{1}{100}$, $\sum_{j=1}^{d} \langle \mathbf{P}_{3,j}, w_j^* - w_j \rangle \geq \left( \frac{\pi}{2} - 0.021 \right) \|\mathbf{W}^* - \mathbf{W}\|_F^2$.

*Proof of Theorem 3.3.* By (3), (6), Lemma D.1, Lemma D.2 and Lemma D.3, we know

$$\langle \mathbf{P}, \mathbf{W}^* - \mathbf{W} \rangle \geq \left( \frac{\pi}{2} - 1.321 - 8\gamma - \frac{(1+\gamma) g}{2(1-2\gamma)} \right) \|\mathbf{W}^* - \mathbf{W}\|_F^2 > \left( 0.169 - \frac{(1+\gamma) g}{2(1-2\gamma)} \right) \|\mathbf{W}^* - \mathbf{W}\|_F^2$$

Using Lemma B.2, we get

$$\langle -\nabla \mathsf{L}(\mathbf{W}), \mathbf{W}^* - \mathbf{W} \rangle > \left( 0.084 - \frac{(1+\gamma) g}{2(1-2\gamma)} \right) \|\mathbf{W}^* - \mathbf{W}\|_F^2 > 0.03 \|\mathbf{W}^* - \mathbf{W}\|_F^2$$

The last inequality holds when $g \leq 0.1$. □

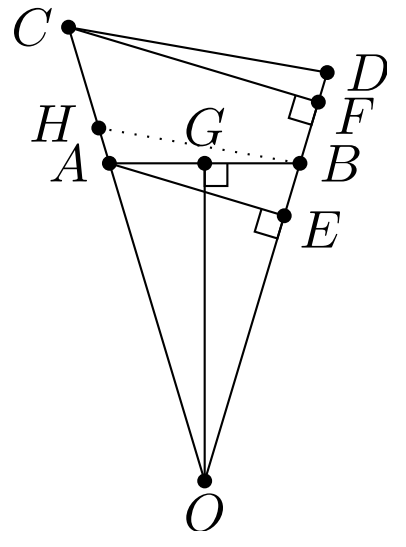

Figure 10: For Lemma E.1

# E A Geometric Lemma

In our proof, we need very tight bounds for a few terms. In order to get such bounds, we present a nice and intuitive geometric lemma as follows.

**Lemma E.1.** *If* $\|\mathbf{W}\|_2, \|\mathbf{W}^*\|_2 \le \gamma$, *then* $\forall i \in [d]$,

1. $\|\overline{e_i + w_i^*} - \overline{e_i + w_i}\|_2 \le \frac{\|(\mathbf{I} - \overline{e_i+w_i} \cdot \overline{e_i+w_i}^\top)(w_i^* - w_i)\|_2}{\sqrt{1-2\gamma}} \le \frac{\|w_i^* - w_i\|_2}{\sqrt{1-2\gamma}}$
2. $-\frac{\|w_i^* - w_i\|_2^2}{2(1-2\gamma)} \le \langle \overline{e_i + w_i^*} - \overline{e_i + w_i}, \overline{e_i + w_i} \rangle \le 0$
3. *if* $\gamma \le \frac{1}{100}$,$0 \le \theta_{i,i^*} \le 1.001\|w_i^* - w_i\|_2$.

*Proof.* See Figure 10. Denote $e_i + w_i^*$ as $\overrightarrow{OC}$, $e_i + w_i$ as $\overrightarrow{OD}$, $\overline{e_i + w_i^*}$ as $\overrightarrow{OA}$, $\overline{e_i + w_i}$ as $\overrightarrow{OB}$. Thus, $\|w_i^* - w_i\|_2 = \|\overrightarrow{DC}\|_2$.

**1.** Since $\overrightarrow{OD} \perp \overrightarrow{CF}$, we know $\|\overrightarrow{CD}\|_2 \ge \|\overrightarrow{CF}\|_2$. Since $\triangle CFO \sim \triangle AEO$, we know

$$\frac{\|\overrightarrow{CD}\|_2}{\|\overrightarrow{AE}\|_2} \ge \frac{\|\overrightarrow{CF}\|_2}{\|\overrightarrow{AE}\|_2} = \frac{\|\overrightarrow{OC}\|_2}{\|\overrightarrow{OA}\|_2} = \|e_i + w_i^*\|_2 \ge 1 - \gamma \tag{7}$$

The last inequality holds as $\|\mathbf{W}^*\|_2 \le \gamma$.

Notice that $\|\overrightarrow{OA}\|_2 = \|\overrightarrow{OB}\|_2 = 1$, we know $\triangle ABO$ is a isosceles triangle. Thus, $\|\overrightarrow{AG}\|_2 = \|\overrightarrow{GB}\|_2$. Notice that $\triangle ABE \sim \triangle BGO$, we have

$$\frac{\|\overrightarrow{AE}\|_2}{\|\overrightarrow{AB}\|_2} = \frac{\|\overrightarrow{OG}\|_2}{\|\overrightarrow{OB}\|_2} = \frac{\sqrt{1 - \|\overrightarrow{GB}\|_2^2}}{1} \tag{8}$$

WLOG, assume $\|\overrightarrow{OC}\|_2 \ge \|\overrightarrow{OD}\|_2$, as shown in the figure. We draw $\overrightarrow{HB} \parallel \overrightarrow{CD}$, and we know $\|\overrightarrow{OH}\|_2 \ge \|\overrightarrow{OB}\|_2 = \|\overrightarrow{OA}\|_2$. Since $\triangle CDO \sim \triangle HBO$, we have

$$\frac{\|\overrightarrow{CD}\|_2}{\|\overrightarrow{HB}\|_2} = \frac{\|\overrightarrow{OD}\|_2}{\|\overrightarrow{OB}\|_2} = \|\overrightarrow{OD}\|_2 \ge 1 - \gamma$$

So $\|\overrightarrow{CD}\|_2 \ge (1-\gamma)\|\overrightarrow{HB}\|_2$. On the other hand, $\angle BAO < \frac{\pi}{2}$, and $A$ is between $H$ and $O$, so $\angle BAH > \frac{\pi}{2}$, which means $\|\overrightarrow{HB}\|_2 \ge \|\overrightarrow{AB}\|_2 = 2\|\overrightarrow{GB}\|_2$. Thus, $\|\overrightarrow{GB}\|_2 \le \frac{\|\overrightarrow{HB}\|_2}{2} \le \frac{\|\overrightarrow{CD}\|_2}{2(1-\gamma)}$.

Substitute it into (8), we get

$$\frac{\|\overrightarrow{AE}\|_2}{\|\overrightarrow{AB}\|_2} \ge \sqrt{1 - \frac{\|\overrightarrow{CD}\|_2^2}{4(1-\gamma)^2}} \ge \sqrt{1 - \left(\frac{\gamma}{1-\gamma}\right)^2}$$

The last inequality holds since $\|\overrightarrow{CD}\|_2 = \|w_i^* - w_i\|_2 \le 2\gamma$.

Substitute this inequality into (7), we get

$$\begin{aligned} &\|\overline{e_i + w_i^*} - \overline{e_i + w_i}\|_2 = \|\overrightarrow{AB}\|_2 \\ \le & \frac{\|\overrightarrow{AE}\|_2}{\sqrt{1 - \left(\frac{\gamma}{1-\gamma}\right)^2}} \le \frac{\|\overrightarrow{CF}\|_2}{(1-\gamma)\sqrt{1 - \left(\frac{\gamma}{1-\gamma}\right)^2}} \quad (9) \\ \le & \frac{\|\overrightarrow{CD}\|_2}{(1-\gamma)\sqrt{1 - \left(\frac{\gamma}{1-\gamma}\right)^2}} = \frac{\|w_i^* - w_i\|_2}{\sqrt{1-2\gamma}} \quad (10) \end{aligned}$$

Notice that $\overline{e_i + w_i}^\top (w_i^* - w_i) = -\|\overrightarrow{DF}\|_2$, so $\overline{e_i + w_i} \cdot \overline{e_i + w_i}^\top (w_i^* - w_i) = \overrightarrow{DF}$. That means,

$$\|(\mathbf{I} - \overline{e_i + w_i} \cdot \overline{e_i + w_i}^\top)(w_i^* - w_i)\|_2 = \|\overrightarrow{DC} - \overrightarrow{DF}\|_2 = \|\overrightarrow{CF}\|_2$$

The lemma follows by (9) and (10).

**2.** By Figure 10, we know $|\langle \overline{e_i + w_i^*} - \overline{e_i + w_i}, \overline{e_i + w_i} \rangle| = \|\overrightarrow{BE}\|_2$. Since $\triangle ABE \sim \triangle GBO$, we have

$$\frac{\|\overrightarrow{BE}\|_2}{\|\overrightarrow{AB}\|_2} = \frac{\|\overrightarrow{GB}\|_2}{\|\overrightarrow{BO}\|_2} = \frac{\|\overrightarrow{AB}\|_2}{2}$$

Therefore, using (10) we get

$$|\langle \overline{e_i + w_i^*} - \overline{e_i + w_i}, \overline{e_i + w_i} \rangle| = \frac{\|\overrightarrow{AB}\|_2^2}{2} \le \frac{\|w_i^* - w_i\|_2^2}{2(1-2\gamma)}$$

Moreover, $\langle \overline{e_i + w_i^*} - \overline{e_i + w_i}, \overline{e_i + w_i} \rangle = \langle \overline{e_i + w_i^*}, \overline{e_i + w_i} \rangle - 1 \le 0$.

**3.** We know that

$$\begin{aligned} \theta_{i,i^*} &= 2 \arcsin \|\overrightarrow{AG}\|_2 = 2 \arcsin \frac{\|\overline{e_i + w_i^*} - \overline{e_i + w_i}\|_2}{2} \\ &\le \|\overline{e_i + w_i^*} - \overline{e_i + w_i}\|_2 + \frac{\|\overline{e_i + w_i^*} - \overline{e_i + w_i}\|_2^3}{8} \end{aligned}$$

The last inequality holds by Taylor's Series for arcsin, and the fact $\|\overline{e_i + w_i^*} - \overline{e_i + w_i}\|_2 = \|\overrightarrow{AB}\|_2 \le \|w_i^* - w_i\|_2 \le 2\gamma \le \frac{1}{50}$. Thus, we have $\theta_{i,i^*} \le 1.001\|w_i^* - w_i\|_2$. ☐

# F More Handy Lemmas

**Lemma* F.1.** *If* $\|\mathbf{W}\|_2, \|\mathbf{W}^*\|_2 \le \gamma$, *then*

- $\frac{(1-\gamma)^2}{(1+\gamma)^2}\mathbf{I} \preceq \overline{\mathbf{I}+\mathbf{W}}^\top \overline{\mathbf{I}+\mathbf{W}} \preceq \frac{(1+\gamma)^2}{(1-\gamma)^2}\mathbf{I}, \quad \frac{(1-\gamma)^2}{(1+\gamma)^2}\mathbf{I} \preceq \overline{\mathbf{I}+\mathbf{W}^*}^\top \overline{\mathbf{I}+\mathbf{W}^*} \preceq \frac{(1+\gamma)^2}{(1-\gamma)^2}\mathbf{I},$
- $(1-\gamma)^2\mathbf{I} \preceq (\mathbf{I}+\mathbf{W})^\top(\mathbf{I}+\mathbf{W}) \preceq (1+\gamma)^2\mathbf{I}, \quad (1-\gamma)^2\mathbf{I} \preceq (\mathbf{I}+\mathbf{W}^*)^\top(\mathbf{I}+\mathbf{W}^*) \preceq (1+\gamma)^2\mathbf{I}.$

*Therefore, the singular value of* $\overline{\mathbf{I}+\mathbf{W}}$ *is at most* $\frac{1+\gamma}{1-\gamma}$ *and at least* $\frac{1-\gamma}{1+\gamma}$. *The singular value of* $\mathbf{I}+\mathbf{W}$ *is at most* $1+\gamma$ *and at least* $1-\gamma$. *The same claims hold for* $\overline{\mathbf{I}+\mathbf{W}^*}$, $\mathbf{I}+\mathbf{W}^*$ *respectively.*

*Proof.* Since $\|\mathbf{W}\|_2 \le \gamma$, we have $1-\gamma \le \|\mathbf{I}+\mathbf{W}\|_2 \le 1+\gamma$, and $1-\gamma \le \|e_i+w_i\|_2 \le 1+\gamma$. Therefore, $\overline{\mathbf{I}+\mathbf{W}} = \mathbf{\Sigma}(\mathbf{I}+\mathbf{W})$ where $\mathbf{\Sigma}$ is a diagonal matrix whose entries are within $[\frac{1}{1+\gamma}, \frac{1}{1-\gamma}]$. Putting into $\overline{\mathbf{I}+\mathbf{W}}^\top \overline{\mathbf{I}+\mathbf{W}}$, we have

$$\overline{\mathbf{I}+\mathbf{W}}^\top \overline{\mathbf{I}+\mathbf{W}} = (\mathbf{I}+\mathbf{W})^\top \mathbf{\Sigma}^2 (\mathbf{I}+\mathbf{W}) \preceq \frac{1}{(1-\gamma)^2}(\mathbf{I}+\mathbf{W})^\top(\mathbf{I}+\mathbf{W}) \preceq \frac{(1+\gamma)^2}{(1-\gamma)^2}\mathbf{I}$$

Similarly we can show $\overline{\mathbf{I}+\mathbf{W}}^\top\overline{\mathbf{I}+\mathbf{W}} \succeq \frac{(1-\gamma)^2}{(1+\gamma)^2}\mathbf{I}$. Thus we know the singular value of $\overline{\mathbf{I}+\mathbf{W}}$ is at most $\frac{1+\gamma}{1-\gamma}$ and at least $\frac{1-\gamma}{1+\gamma}$. The same proof works for $\mathbf{I}+\mathbf{W}$, $\overline{\mathbf{I}+\mathbf{W}^*}$ and $\mathbf{I}+\mathbf{W}^*$. □

**Lemma* F.2.** *If* $\|\mathbf{W}\|_2, \|\mathbf{W}^*\|_2 \le \gamma \le \frac{1}{100}$, *we have*

$$|\langle\overline{e_i+w_i^*}, \overline{e_j+w_j}\rangle| \le 2.1\gamma, \quad |\langle\overline{e_i+w_i}, \overline{e_j+w_j}\rangle| \le 2.1\gamma$$

*Proof.* We know

$$|\langle\overline{e_i+w_i^*}, \overline{e_j+w_j}\rangle| = \frac{|\langle e_i+w_i^*, e_j+w_j\rangle|}{\|e_i+w_i^*\|_2\|e_j+w_j\|_2} \le \frac{|\langle e_i+w_i^*, e_j+w_j\rangle|}{(1-\gamma)^2} = \frac{|w_{i,j}^*|+|w_{i,j}|+|\langle w_i, w_j\rangle|}{(1-\gamma)^2} \le \frac{(2+\gamma)\gamma}{(1-\gamma)^2} \le 2.1\gamma$$

where the last inequality holds since $\gamma \le \frac{1}{100}$. The same analysis works for $\langle\overline{e_i+w_i}, \overline{e_j+w_j}\rangle$. □

**Lemma* F.3** (Triangle inequality between $e_i+w_i, e_i+w_i^*, w_i^*-w_i$)**.** $|\|e_i+w_i\|_2 - \|e_i+w_i^*\|_2| \le \|w_i^*-w_i\|_2$.

**Lemma* F.4.** *If* $\|\mathbf{W}\|_2, \|\mathbf{W}^*\|_2 \le \gamma$, $|g| \le 2d\gamma$.

*Proof.* By definition and Lemma F.3, we know $|g| = \sum_{i=1}^d(\|e_i+w_i^*\|_2 - \|e_i+w_i\|_2) \le \sum_{i=1}^d \|w_i^*-w_i\|_2 \le 2d\gamma$. □

**Lemma* F.5.** *If* $\|\mathbf{W}\|_2, \|\mathbf{W}^*\|_2 \le \gamma$, $|\langle\overline{e_i+w_i^*} - \overline{e_i+w_i}, \overline{e_j+w_j}\rangle| \le \frac{\|w_i^*-w_i\|_2}{\sqrt{1-2\gamma}}$.

*Proof.* By Cauchy Schwartz and Lemma E.1 term 1. □

**Lemma* F.6.** $|x^k-y^k| \le \frac{k}{2}|x-y|(|x|^{k-1}+|y|^{k-1})$.

*Proof.* $|x^k-y^k| = \left|(x-y)\sum_{t=1}^{k-1}\frac{x^ty^{k-t-1}+y^tx^{k-t-1}}{2}\right| \le \frac{k}{2}|x-y|(|x|^{k-1}+|y|^{k-1})$, where the last inequality holds since $|x^ty^{k-t-1}+y^tx^{k-t-1}| \le |x|^t|y|^{k-t-1}+|y|^t|x|^{k-t-1} \le |x|^{k-1}+|y|^{k-1}$, by rearrangement inequality. □

**Lemma* F.7.** *If* $\|\mathbf{W}\|_2, \|\mathbf{W}^*\|_2 \le \gamma \le \frac{1}{100}$, *for* $k \ge 3$, *we have*

$$\begin{aligned}
&\|\langle\overline{e_i+w_i^*}, \overline{e_j+w_j}\rangle^k(e_i+w_i^*) - \langle\overline{e_i+w_i}, \overline{e_j+w_j}\rangle^k(e_i+w_i)\|_2 \\
\le& 6(2.2\gamma)^{k-3}\left(\langle\overline{e_i+w_i^*}, \overline{e_j+w_j}\rangle^2 + \langle\overline{e_i+w_i}, \overline{e_j+w_j}\rangle^2\right)\|w_i^*-w_i\|_2
\end{aligned}$$

*Proof.*

$$\begin{aligned}
&\|\langle\overline{e_i+w_i^*}, \overline{e_j+w_j}\rangle^k(e_i+w_i^*) - \langle\overline{e_i+w_i}, \overline{e_j+w_j}\rangle^k(e_i+w_i)\|_2 \\
\le&\|w_i^*-w_i\|_2|\langle\overline{e_i+w_i^*}, \overline{e_j+w_j}\rangle^k| + \|(\langle\overline{e_i+w_i^*}, \overline{e_j+w_j}\rangle^k - \langle\overline{e_i+w_i}, \overline{e_j+w_j}\rangle^k)(e_i+w_i)\|_2 \\
\le&\|w_i^*-w_i\|_2|\langle\overline{e_i+w_i^*}, \overline{e_j+w_j}\rangle^k| + (1+\gamma)|\langle\overline{e_i+w_i^*}, \overline{e_j+w_j}\rangle^k - \langle\overline{e_i+w_i}, \overline{e_j+w_j}\rangle^k| \\
\overset{①}{\le}&\|w_i^*-w_i\|_2(2.1\gamma)^{k-2}\langle\overline{e_i+w_i^*}, \overline{e_j+w_j}\rangle^2 \\
&+\frac{(1+\gamma)k}{2}|\langle\overline{e_i+w_i^*} - \overline{e_i+w_i}, \overline{e_j+w_j}\rangle|(|\langle\overline{e_i+w_i^*}, \overline{e_j+w_j}\rangle|^{k-1} + |\langle\overline{e_i+w_i}, \overline{e_j+w_j}\rangle|^{k-1}) \\
\le&\langle\overline{e_i+w_i^*}, \overline{e_j+w_j}\rangle^2\left(\|w_i^*-w_i\|_2(2.1\gamma)^{k-2} + \frac{(1+\gamma)k(2.1\gamma)^{k-3}}{2}|\langle\overline{e_i+w_i^*} - \overline{e_i+w_i}, \overline{e_j+w_j}\rangle|\right) \\
&+\langle\overline{e_i+w_i}, \overline{e_j+w_j}\rangle^2\left(\frac{(1+\gamma)k(2.1\gamma)^{k-3}}{2}|\langle\overline{e_i+w_i^*} - \overline{e_i+w_i}, \overline{e_j+w_j}\rangle|\right) \\
\overset{②}{\le}&\|w_i^*-w_i\|_2\left[\left((2.1\gamma)^{k-2} + 0.52k(2.1\gamma)^{k-3}\right)\langle\overline{e_i+w_i^*}, \overline{e_j+w_j}\rangle^2 + 0.52k(2.1\gamma)^{k-3}\langle\overline{e_i+w_i}, \overline{e_j+w_j}\rangle^2\right] \\
\overset{③}{\le}&\|w_i^*-w_i\|_2\left[0.55k(2.1\gamma)^{k-3}\langle\overline{e_i+w_i^*}, \overline{e_j+w_j}\rangle^2 + 0.52k(2.1\gamma)^{k-3}\langle\overline{e_i+w_i}, \overline{e_j+w_j}\rangle^2\right] \\
\overset{④}{\le}&6(2.2\gamma)^{k-3}\left(\langle\overline{e_i+w_i^*}, \overline{e_j+w_j}\rangle^2 + \langle\overline{e_i+w_i}, \overline{e_j+w_j}\rangle^2\right)\|w_i^*-w_i\|_2
\end{aligned}$$

where ① uses Lemma F.2 and Lemma F.6, ② uses Lemma F.5, ③ holds as $\gamma \leq \frac{1}{100}$, and ④ holds since $0.55k(2.1)^{k-3} \leq 6(2.2)^{k-3}$ for $k \geq 3$. $\square$

**Lemma* F.8.** *If $\|\mathbf{W}\|_2, \|\mathbf{W}^*\|_2 \leq \gamma \leq \frac{1}{100}$, for $k \geq 2$,*

$$\begin{aligned}&\left|\|e_i + w_i\|_2 \langle \overline{e_i + w_i}, \overline{e_j + w_j}\rangle^{2k} - \|e_i + w_i^*\|_2 \langle \overline{e_i + w_i^*}, \overline{e_j + w_j}\rangle^{2k}\right| \\ \leq & 8(2.2\gamma)^{2k-3}\left(\langle \overline{e_i + w_i}, \overline{e_j + w_j}\rangle^2 + \langle \overline{e_i + w_i^*}, \overline{e_j + w_j}\rangle^2\right)\|w_i^* - w_i\|_2\end{aligned}$$

*Proof.*

$$\begin{aligned}&\left|\|e_i + w_i\|_2 \langle \overline{e_i + w_i}, \overline{e_j + w_j}\rangle^{2k} - \|e_i + w_i^*\|_2 \langle \overline{e_i + w_i^*}, \overline{e_j + w_j}\rangle^{2k}\right| \\ \leq & \|e_i + w_i\|_2 \left|\langle \overline{e_i + w_i}, \overline{e_j + w_j}\rangle^{2k} - \langle \overline{e_i + w_i^*}, \overline{e_j + w_j}\rangle^{2k}\right| + \left|\|e_i + w_i\|_2 - \|e_i + w_i^*\|_2\right| \langle \overline{e_i + w_i^*}, \overline{e_j + w_j}\rangle^{2k} \\ \overset{①}{\leq} & \|e_i + w_i\|_2 \left|\langle \overline{e_i + w_i}, \overline{e_j + w_j}\rangle^{2k} - \langle \overline{e_i + w_i^*}, \overline{e_j + w_j}\rangle^{2k}\right| + \|w_i^* - w_i\|_2 (2.1\gamma)^{2k-2} \langle \overline{e_i + w_i^*}, \overline{e_j + w_j}\rangle^2 \\ \overset{②}{\leq} & (1+\gamma)k |\langle \overline{e_i + w_i} - \overline{e_i + w_i^*}, \overline{e_j + w_j}\rangle| \left(|\langle \overline{e_i + w_i}, \overline{e_j + w_j}\rangle|^{2k-1} + |\langle \overline{e_i + w_i^*}, \overline{e_j + w_j}\rangle|^{2k-1}\right) \\ & + \|w_i^* - w_i\|_2 (2.1\gamma)^{2k-2} \langle \overline{e_i + w_i^*}, \overline{e_j + w_j}\rangle^2 \\ \overset{③}{\leq} & \left[\frac{(1+\gamma)k(2.1\gamma)^{2k-3}}{\sqrt{1-2\gamma}} \langle \overline{e_i + w_i}, \overline{e_j + w_j}\rangle^2 + \left(\frac{(1+\gamma)k(2.1\gamma)^{2k-3}}{\sqrt{1-2\gamma}} + (2.1\gamma)^{2k-2}\right) \langle \overline{e_i + w_i^*}, \overline{e_j + w_j}\rangle^2\right] \|w_i^* - w_i\|_2 \\ \overset{④}{\leq} & 1.05k(2.1\gamma)^{2k-3}\left(\langle \overline{e_i + w_i}, \overline{e_j + w_j}\rangle^2 + \langle \overline{e_i + w_i^*}, \overline{e_j + w_j}\rangle^2\right)\|w_i^* - w_i\|_2 \\ \overset{⑤}{\leq} & 8(2.2\gamma)^{2k-3}\left(\langle \overline{e_i + w_i}, \overline{e_j + w_j}\rangle^2 + \langle \overline{e_i + w_i^*}, \overline{e_j + w_j}\rangle^2\right)\|w_i^* - w_i\|_2\end{aligned}$$

where ① uses Lemma F.2 and Lemma F.3, ② uses Lemma F.6, ③ uses Lemma F.5, ④ holds as $\gamma \leq \frac{1}{100}$, and ⑥ holds as $1.05k(2.1)^{2k-3} \leq 8(2.2)^{2k-3}$ for $k \geq 2$. $\square$

**Lemma* F.9.** *If $\|\mathbf{W}\|_2, \|\mathbf{W}^*\|_2 \leq \gamma$, for fixed $j \in [d]$,*

$$\sum_{i \neq j} \langle \overline{e_i + w_i}, \overline{e_j + w_j}\rangle^2 \leq \frac{4\gamma}{(1-\gamma)^2}, \qquad \sum_{i \neq j} \langle \overline{e_i + w_i^*}, \overline{e_j + w_j}\rangle^2 \leq \frac{4\gamma(1+\gamma)}{1-2\gamma}.$$

*Similarly, for fixed $i \in [d]$,*

$$\sum_{j \neq i} \langle \overline{e_i + w_i}, \overline{e_j + w_j}\rangle^2 \leq \frac{4\gamma}{(1-\gamma)^2}, \qquad \sum_{j \neq i} \langle \overline{e_i + w_i^*}, \overline{e_j + w_j}\rangle^2 \leq \frac{4\gamma(1+\gamma)}{1-2\gamma}.$$

*Proof.* By matrix multiplication,

$$\sum_{i=1}^{d} \langle \overline{e_i + w_i^*}, \overline{e_j + w_j}\rangle^2 = \sum_{i=1}^{d} \overline{e_j + w_j}^\top \overline{e_i + w_i^*} \cdot \overline{e_i + w_i^*}^\top \overline{e_j + w_j} = \overline{e_j + w_j}^\top \overline{\mathbf{I} + \mathbf{W}^*} \cdot \overline{\mathbf{I} + \mathbf{W}^*}^\top \overline{e_j + w_j}$$

By Lemma F.1, we know $\overline{\mathbf{I} + \mathbf{W}^*} \cdot \overline{\mathbf{I} + \mathbf{W}^*}^\top \preceq \frac{(1+\gamma)^2}{(1-\gamma)^2}\mathbf{I}$. That means, $\sum_{i=1}^{d} \langle \overline{e_i + w_i^*}, \overline{e_j + w_j}\rangle^2 \leq \frac{(1+\gamma)^2}{(1-\gamma)^2}$. On the other hand, by Lemma E.1 term 2, $\langle \overline{e_j + w_j^*}, \overline{e_j + w_j}\rangle^2 = (1 - \langle \overline{e_j + w_j^*} - \overline{e_j + w_j}, \overline{e_j + w_j}\rangle)^2 \geq 1 - \frac{\|w_i^* - w_i\|_2^2}{1-2\gamma}$.

Therefore, we know

$$\sum_{i \neq j} \langle \overline{e_i + w_i^*}, \overline{e_j + w_j}\rangle^2 \leq \frac{(1+\gamma)^2}{(1-\gamma)^2} - 1 + \frac{\|w_i^* - w_i\|_2^2}{1-2\gamma} = \frac{4\gamma}{(1-\gamma)^2} + \frac{\|w_i^* - w_i\|_2^2}{1-2\gamma} \leq \frac{4\gamma(1+\gamma)}{1-2\gamma}$$

Using the same analysis, we get $\sum_{i \neq j} \langle \overline{e_i + w_i}, \overline{e_j + w_j}\rangle^2 \leq \frac{(1+\gamma)^2}{(1-\gamma)^2} - 1 = \frac{4\gamma}{(1-\gamma)^2}$. The analysis for fixed $i$ is similar. $\square$

**Lemma* F.10.** *For any matrix* $\mathbf{A}$, *we have* $\|\mathrm{Diag}(\mathbf{A})\|_2 \le \|\mathbf{A}\|_2$ *and* $\|\text{Off-Diag}(\mathbf{A})\|_2 \le 2\|\mathbf{A}\|_2$.

*Proof.* By definition, we know $\|\mathrm{Diag}(\mathbf{A})\|_2 = \max_{i\in[d]} e_i^\top \mathbf{A} e_i \le \max_{v\in\mathbb{R}^d} v^\top \mathbf{A} v = \|\mathbf{A}\|_2$, and $\|\text{Off-Diag}(\mathbf{A})\|_2 \le \|\mathbf{A}\|_2 + \|\mathrm{Diag}(\mathbf{A})\|_2 \le 2\|\mathbf{A}\|_2$. □

**Lemma* F.11.** *If* $\|\mathbf{W}\|_2, \|\mathbf{W}^*\|_2 \le \gamma$, $\|\mathbf{A}\|_2 \le \frac{2\gamma(\gamma^2+3)}{1-\gamma^2}$.

*Proof.* By Lemma F.1, we have

$$\|\mathbf{A}\|_2 = \|(\mathbf{I}+\mathbf{W}^*)\overline{\mathbf{I}+\mathbf{W}^*}^\top - (\mathbf{I}+\mathbf{W})\overline{\mathbf{I}+\mathbf{W}}^\top\|_2 \le \frac{(1+\gamma)^2}{1-\gamma} - \frac{(1-\gamma)^2}{1+\gamma} = \frac{2\gamma(\gamma^2+3)}{1-\gamma^2}. \qquad \square$$

**Lemma* F.12.** *If* $\|\mathbf{W}\|_2, \|\mathbf{W}^*\|_2 \le \gamma \le \frac{1}{100}$, $|\overline{e_j+w_j}^\top \mathbf{A} \overline{e_j+w_j} - e_j^\top \mathbf{A} e_j| \le 5\gamma^2$.

*Proof.*

$$|\overline{e_j+w_j}^\top \mathbf{A}\overline{e_j+w_j} - e_j^\top \mathbf{A} e_j| \le |\overline{e_j+w_j}^\top \mathbf{A}(\overline{e_j+w_j} - e_j)| + |(\overline{e_j+w_j} - e_j)^\top \mathbf{A} e_j| \overset{①}{\le} \frac{4\gamma^2(\gamma^2+3)}{1-\gamma^2} \overset{②}{<} 5\gamma^2$$

where ① uses Cauchy Schwartz, Lemma F.11 and $\|\overline{e_j+w_j} - e_j\|_2 \le \gamma$, and ② holds as $\gamma \le \frac{1}{100}$. □

**Lemma* F.13.** *For any* $i \in [n]$, $||[\|e_i + w_i^*\|_2 - \|e_i + w_i\|_2] - [w_{i,i}^* - w_{i,i}]| \le 6.07\gamma^2$.

*Proof.*

$$\begin{aligned}
\|e_i + w_i\|_2 - \|e_i + w_i^*\|_2 &= \langle e_i + w_i, \overline{e_i+w_i}\rangle - \langle e_i + w_i^*, \overline{e_i + w_i^*}\rangle \\
&= \langle e_i + w_i, \overline{e_i+w_i} - \overline{e_i+w_i^*}\rangle + \langle w_i - w_i^*, \overline{e_i+w_i^*}\rangle \\
&= \langle w_i - w_i^*, e_i\rangle + \langle e_i + w_i, \overline{e_i+w_i} - \overline{e_i+w_i^*}\rangle + \langle w_i - w_i^*, \overline{e_i+w_i^*} - e_i\rangle \\
&= w_{i,i} - w_{i,i}^* + \langle e_i + w_i, \overline{e_i+w_i} - \overline{e_i+w_i^*}\rangle + \langle w_i - w_i^*, \overline{e_i+w_i^*} - e_i\rangle
\end{aligned}$$

As a result,

$$\begin{aligned}
&|[\|e_i + w_i\|_2 - \|e_i + w_i^*\|_2] - [w_{i,i} - w_{i,i}^*]| \le |\langle e_i + w_i, \overline{e_i+w_i} - \overline{e_i+w_i^*}\rangle| + |\langle w_i - w_i^*, \overline{e_i+w_i^*} - e_i\rangle| \\
&\overset{①}{\le} \frac{(1+\gamma)2\gamma^2}{1-2\gamma} + 4\gamma^2 \le 6.07\gamma^2
\end{aligned}$$

where ① uses Lemma E.1 term 2 and $\|\overline{e_i+w_i^*} - e_i\|_2 \le 2\gamma$, and Cauchy Schwartz. So the claim follows. □

**Corollary F.14.** $|g - \mathrm{Tr}(\mathbf{W}^* - \mathbf{W})| \le 6.07d\gamma^2$.

**Lemma* F.15.** $\overline{\mathbf{I}+\mathbf{W}}$ *is close to* $\mathbf{I}$ *on its diagonals, and close to* $\mathbf{W}$ *on its off-diagonals. More specifically, if* $\|\mathbf{W}\|_2, \|\mathbf{W}^*\|_2 \le \gamma \le \frac{1}{100}$,

$$\begin{aligned}
&\|\mathrm{Diag}(\overline{\mathbf{I}+\mathbf{W}}) - \mathbf{I}\|_2 \le \frac{\gamma^2}{2(1-\gamma)^2}, && \|\mathrm{Diag}(\overline{\mathbf{I}+\mathbf{W}^*}) - \mathbf{I}\|_2 \le \frac{\gamma^2}{2(1-\gamma)^2} \\
&\|\text{Off-Diag}(\overline{\mathbf{I}+\mathbf{W}} - \mathbf{W})\|_2 \le \frac{4\gamma^2}{1-\gamma}, && \|\text{Off-Diag}(\overline{\mathbf{I}+\mathbf{W}^*} - \mathbf{W}^*)\|_2 \le \frac{4\gamma^2}{1-\gamma} \\
&\|\overline{\mathbf{I}+\mathbf{W}} - \mathbf{I}\|_2 \le 2.05\gamma, && \|\overline{\mathbf{I}+\mathbf{W}^*} - \mathbf{I}\|_2 \le 2.05\gamma
\end{aligned}$$

*Proof.* For the diagonal terms,

$$\begin{aligned}
\|\mathrm{Diag}(\overline{\mathbf{I}+\mathbf{W}}) - \mathbf{I}\|_2 &= \max_j |\overline{\mathbf{I}+\mathbf{W}}_{j,j} - 1| = \max_j \left|\frac{1 + w_{j,j} - \|e_j + w_j\|_2}{\|e_j + w_j\|_2}\right| \\
&\le \max_j \left|\frac{(1+w_{j,j})^2 - \|e_j + w_j\|_2^2}{\|e_j + w_j\|_2}\right| \left|\frac{1}{1 + w_{j,j} + \|e_j + w_j\|_2}\right| \le \max_j \frac{\sum_{i\ne j} w_{j,i}^2}{2(1-\gamma)^2} \le \frac{\gamma^2}{2(1-\gamma)^2}
\end{aligned}$$

For the off-diagonal terms, we know $\overline{\mathbf{I}+\mathbf{W}} = (\mathbf{I}+\mathbf{W})\mathbf{\Sigma}$ for some diagonal matrix $\mathbf{\Sigma}$, so

$$\|\text{Off-Diag}(\overline{\mathbf{I}+\mathbf{W}} - \mathbf{W})\|_2 = \|\text{Off-Diag}((\mathbf{I}+\mathbf{W})\mathbf{\Sigma} - \mathbf{W})\|_2 = \|\text{Off-Diag}((\mathbf{\Sigma}-\mathbf{I})\mathbf{W})\|_2 \overset{①}{\leq} 2\|(\mathbf{\Sigma}-\mathbf{I})\mathbf{W}\|_2 \leq \frac{4\gamma^2}{1-\gamma}$$

where ① uses Lemma F.10. For the difference between $\overline{\mathbf{I}+\mathbf{W}}$ and $\mathbf{I}$, we split $\overline{\mathbf{I}+\mathbf{W}}$ into diagonal and off-diagonal parts:

$$\begin{aligned}
&\|\overline{\mathbf{I}+\mathbf{W}} - \mathbf{I}\|_2 = \|\text{Diag}(\overline{\mathbf{I}+\mathbf{W}}) + \text{Off-Diag}(\overline{\mathbf{I}+\mathbf{W}}) - \mathbf{I}\|_2 \\
=&\|\text{Off-Diag}(\mathbf{W})\|_2 + \frac{\gamma^2}{2(1-\gamma)^2} + \frac{4\gamma^2}{1-\gamma} \overset{①}{\leq} 2\|\mathbf{W}\|_2 + \frac{\gamma^2(9-8\gamma)}{2(1-\gamma)^2} \leq 2.05\gamma
\end{aligned}$$

where ① uses Lemma F.10. □

**Lemma* F.16.** *If* $\|\mathbf{W}\|_2, \|\mathbf{W}^*\|_2 \leq \gamma \leq \frac{1}{100}$,

$$\|\mathbf{A} - [\mathbf{W}^* - \mathbf{W} + (\mathbf{W}^* - \mathbf{W})^\top - \text{Diag}(\mathbf{W}^* - \mathbf{W})]\|_2 \leq 9.2\gamma^2$$

*Proof.* By definition,

$$\begin{aligned}
&\left\|\left[(\mathbf{I}+\mathbf{W}^*)\overline{\mathbf{I}+\mathbf{W}^*}^\top - (\mathbf{I}+\mathbf{W})\overline{\mathbf{I}+\mathbf{W}}^\top\right] - \left[(\mathbf{W}^*-\mathbf{W}) + (\overline{\mathbf{I}+\mathbf{W}^*}^\top - \overline{\mathbf{I}+\mathbf{W}}^\top)\right]\right\|_2 \\
=&\|\mathbf{W}^*(\overline{\mathbf{I}+\mathbf{W}^*}^\top - \mathbf{I}) - \mathbf{W}(\overline{\mathbf{I}+\mathbf{W}}^\top - \mathbf{I})\|_2 \leq \|\mathbf{W}^*(\overline{\mathbf{I}+\mathbf{W}^*}^\top - \mathbf{I})\|_2 + \|\mathbf{W}(\overline{\mathbf{I}+\mathbf{W}})^\top - \mathbf{I})\|_2 \\
\leq&2.05\gamma^2 + 2.05\gamma^2 = 4.1\gamma^2
\end{aligned}$$

where the last inequality uses Lemma F.15. Below we further approximate $\overline{\mathbf{I}+\mathbf{W}^*}^\top - \overline{\mathbf{I}+\mathbf{W}}^\top$.

$$\begin{aligned}
&\left\|\left[\overline{\mathbf{I}+\mathbf{W}^*}^\top - \overline{\mathbf{I}+\mathbf{W}}^\top\right] - \left[(\mathbf{W}^*-\mathbf{W})^\top - \text{Diag}(\mathbf{W}^*-\mathbf{W})\right]\right\|_2 \\
=&\left\|\text{Diag}(\overline{\mathbf{I}+\mathbf{W}^*}^\top - \overline{\mathbf{I}+\mathbf{W}}^\top) + \text{Off-Diag}(\overline{\mathbf{I}+\mathbf{W}^*}^\top - \overline{\mathbf{I}+\mathbf{W}}^\top) - \left[(\mathbf{W}^*-\mathbf{W})^\top - \text{Diag}(\mathbf{W}^*-\mathbf{W})\right]\right\|_2 \\
\overset{①}{\leq}&\|\text{Off-Diag}(\overline{\mathbf{I}+\mathbf{W}^*}^\top - \overline{\mathbf{I}+\mathbf{W}}^\top) - \text{Off-Diag}(\mathbf{W}^*-\mathbf{W})^\top\|_2 + \frac{\gamma^2}{(1-\gamma)^2} \\
\overset{②}{\leq}&\frac{4\gamma^2}{1-\gamma} + \frac{\gamma^2}{(1-\gamma)^2} \leq 5.1\gamma^2
\end{aligned}$$

where ① uses Lemma F.15, ② uses Lemma F.15 Combining everything,

$$\|\mathbf{A} - [\mathbf{W}^* - \mathbf{W} + (\mathbf{W}^* - \mathbf{W})^\top - \text{Diag}(\mathbf{W}^* - \mathbf{W})]\|_2 \leq 9.2\gamma^2$$ □

Using Lemma F.10, we immediately have the following corollary.

**Corollary F.17.** $\|\text{Diag}(\mathbf{A}) - \text{Diag}(\mathbf{W}^* - \mathbf{W})\|_2 \leq 9.2\gamma^2$.

**Lemma* F.18.** *For* $\eta \leq \frac{1}{\pi d}$,

$$\left\|\mathbf{I} - \eta\left(\frac{\pi}{2}uu^\top + \left(\frac{\pi}{2}+1\right)\mathbf{I}\right)\right\|_2 \leq \left(1 - \eta\left(\frac{\pi}{2}+1\right)\right)$$

*Proof.* Consider another basis $(e'_1, \cdots, e'_d)$ where $e'_1 = \frac{u}{\|u\|_2}$. For every unit vector $v = (v_1, \cdots, v_d)$ in this new space, we know

$$v^T\left(\mathbf{I} - \eta\left(\frac{\pi}{2}uu^\top + \left(\frac{\pi}{2}+1\right)\mathbf{I}\right)\right)v = \|v\|_2^2 - \eta\left(\frac{\pi}{2}+1\right)\|v\|_2^2 - \frac{\pi\eta d}{2}v_1^2$$

Hence we get

$$0 \leq v^T\left(\mathbf{I} - \eta\left(\frac{\pi}{2}uu^\top + \left(\frac{\pi}{2}+1\right)\mathbf{I}\right)\right)v \leq \left(1 - \eta\left(\frac{\pi}{2}+1\right)\right)\|v\|_2^2$$

By definition of matrix norm, the lemma follows. □

# G Proofs for Section B

## G.1 Proof for Claim B.1

Comparing with Lemma 2.1, we know that for fixed $j$, $\mathbf{P}_{1,j}$ is already contained in $-\nabla \mathsf{L}(\mathbf{W})_j$ as the first term, while $\mathbf{P}_{3,j}$ is simply the summand when $i = j$, ignoring the first term. Below we show how to obtain $\mathbf{P}_{2,j}$ from $i \neq j$ cases. We will bound the approximation error in Lemma B.2 and Lemma B.3.

$$
\begin{aligned}
&\sum_{i \neq j} \left( \left( \frac{\pi}{2} - \theta_{i^*,j} \right) (e_i + w_i^*) - \left( \frac{\pi}{2} - \theta_{i,j} \right) (e_i + w_i) + (\|e_i + w_i^*\| \sin \theta_{i^*,j} - \|e_i + w_i\| \sin \theta_{i,j}) \, \overline{e_j + w_j} \right) \\
\approx &\sum_{i \neq j} \left( \langle \overline{e_i + w_i^*}, \overline{e_j + w_j} \rangle (e_i + w_i^*) - \langle \overline{e_i + w_i}, \overline{e_j + w_j} \rangle (e_i + w_i) \right) \\
&+ \sum_{i \neq j} \left( \|e_i + w_i^*\| \left( 1 - \frac{1}{2} \langle \overline{e_i + w_i^*}, \overline{e_j + w_j} \rangle^2 \right) - \|e_i + w_i\| \left( 1 - \frac{1}{2} \langle \overline{e_i + w_i}, \overline{e_j + w_j} \rangle^2 \right) \right) \overline{e_j + w_j} \\
= &\sum_{i \neq j} ((e_i + w_i^*) \overline{e_i + w_i^*}^\top - (e_i + w_i) \overline{e_i + w_i}^\top) \overline{e_j + w_j} \\
&+ \sum_{i \neq j} \left( \|e_i + w_i^*\| - \|e_i + w_i\| - \frac{1}{2} \overline{e_j + w_j}^\top \overline{e_i + w_i^*} \|e_i + w_i^*\| \overline{e_i + w_i^*}^\top \overline{e_j + w_j} \right. \\
&\left. + \frac{1}{2} \overline{e_j + w_j}^\top \overline{e_i + w_i} \|e_i + w_i\| \overline{e_i + w_i}^\top \overline{e_j + w_j} \right) \overline{e_j + w_j} \\
= &\mathbf{A}_j \overline{e_j + w_j} + \left( \sum_{i \neq j} (\|e_i + w_i^*\| - \|e_i + w_i\|) - \sum_{i \neq j} \frac{1}{2} \overline{e_j + w_j}^\top (e_i + w_i^*) \overline{e_i + w_i^*}^\top \overline{e_j + w_j} \right. \\
&\left. + \sum_{i \neq j} \frac{1}{2} \overline{e_j + w_j}^\top (e_i + w_i) \overline{e_i + w_i}^\top \overline{e_j + w_j} \right) \overline{e_j + w_j} \\
= &\mathbf{A}_j \overline{e_j + w_j} + \left( g_j - \frac{1}{2} \overline{e_j + w_j}^\top \mathbf{A}_j \overline{e_j + w_j} \right) \overline{e_j + w_j} = \mathbf{P}_{2,j}.
\end{aligned}
$$

## G.2 Proof for Lemma B.2

In order to prove this lemma, we bound the approximation loss of $\theta_{i,j}, \theta_{i^*,j}$ in Lemma G.1, and the approximation loss of $\sin \theta_{i,j}, \sin \theta_{i^*,j}$ in Lemma G.2.

**Lemma* G.1** (Approximation loss related to $\theta_{i,j}, \theta_{i^*,j}$). *If* $\|\mathbf{W}\|_2, \|\mathbf{W}^*\|_2 \leq \gamma \leq \frac{1}{100}$,

$$
\begin{aligned}
&\sum_{j=1}^{d} \sum_{i \neq j} \left| \left\langle (\frac{\pi}{2} - \theta_{i^*,j} - \langle \overline{e_i + w_i^*}, \overline{e_j + w_j} \rangle)(e_i + w_i^*) - (\frac{\pi}{2} - \theta_{i,j} - \langle \overline{e_i + w_i}, \overline{e_j + w_j} \rangle)(e_i + w_i), w_j^* - w_j \right\rangle \right| \\
\leq &0.083 \|\mathbf{W}^* - \mathbf{W}\|_F^2
\end{aligned}
$$

*Proof.* By definition, $\frac{\pi}{2} - \theta_{i^*,j} = \arcsin \langle \overline{e_i + w_i^*}, \overline{e_j + w_j} \rangle$, and $\frac{\pi}{2} - \theta_{i,j} = \arcsin \langle \overline{e_i + w_i}, \overline{e_j + w_j} \rangle$.

The Taylor series of $\arcsin x$ at $x = 0$ is $\sum_{k=0}^{\infty} \frac{(2k)!}{4^k (k!)^2 (2k+1)} x^{2k+1}$, where for $k \geq 1$,

$$
\frac{(2k)!}{4^k (k!)^2 (2k+1)} \leq \frac{1}{6} \tag{11}
$$

Thus,

$$\sum_{j=1}^{d}\sum_{i\neq j}\left|\left\langle(\frac{\pi}{2}-\theta_{i^*,j}-\langle\overline{e_i+w_i^*},\overline{e_j+w_j}\rangle)(e_i+w_i^*)-(\frac{\pi}{2}-\theta_{i,j}-\langle\overline{e_i+w_i},\overline{e_j+w_j}\rangle)(e_i+w_i),w_j^*-w_j\right\rangle\right|$$

$$\overset{①}{\leq}\sum_{j=1}^{d}\sum_{i\neq j}\sum_{k=1}^{\infty}\frac{1}{6}\left|\left\langle\langle\overline{e_i+w_i^*},\overline{e_j+w_j}\rangle^{2k+1}(e_i+w_i^*)-\langle\overline{e_i+w_i},\overline{e_j+w_j}\rangle^{2k+1}(e_i+w_i),w_j^*-w_j\right\rangle\right|$$

$$\overset{②}{\leq}\sum_{j=1}^{d}\sum_{i\neq j}\sum_{k=1}^{\infty}\frac{1}{6}\left\|\langle\overline{e_i+w_i^*},\overline{e_j+w_j}\rangle^{2k+1}(e_i+w_i^*)-\langle\overline{e_i+w_i},\overline{e_j+w_j}\rangle^{2k+1}(e_i+w_i)\right\|_2\|w_j^*-w_j\|_2$$

$$\overset{③}{\leq}\sum_{j=1}^{d}\sum_{i\neq j}\sum_{k=1}^{\infty}(2.2\gamma)^{2k-2}\left(\langle\overline{e_i+w_i^*},\overline{e_j+w_j}\rangle^2+\langle\overline{e_i+w_i},\overline{e_j+w_j}\rangle^2\right)\|w_i^*-w_i\|_2\|w_j^*-w_j\|_2$$

$$\overset{④}{\leq}\sum_{j=1}^{d}\sum_{i\neq j}1.01\left(\langle\overline{e_i+w_i^*},\overline{e_j+w_j}\rangle^2+\langle\overline{e_i+w_i},\overline{e_j+w_j}\rangle^2\right)\|w_i^*-w_i\|_2\|w_j^*-w_j\|_2$$

$$\overset{⑤}{\leq}1.01\left(\sum_{j=1}^{d}\sum_{i\neq j}\left(\langle\overline{e_i+w_i^*},\overline{e_j+w_j}\rangle^2+\langle\overline{e_i+w_i},\overline{e_j+w_j}\rangle^2\right)\|w_i^*-w_i\|_2^2\right)^{\frac{1}{2}}$$

$$\left(\sum_{j=1}^{d}\sum_{i\neq j}\left(\langle\overline{e_i+w_i^*},\overline{e_j+w_j}\rangle^2+\langle\overline{e_i+w_i},\overline{e_j+w_j}\rangle^2\right)\|w_j^*-w_j\|_2^2\right)^{\frac{1}{2}}$$

$$\leq1.01\left[\sum_{i=1}^{d}\|w_i^*-w_i\|_2^2\left(\sum_{i\neq j}\left(\langle\overline{e_i+w_i^*},\overline{e_j+w_j}\rangle^2+\langle\overline{e_i+w_i},\overline{e_j+w_j}\rangle^2\right)\right)\right]^{\frac{1}{2}}$$

$$\left[\sum_{j=1}^{d}\|w_j^*-w_j\|_2^2\left(\sum_{i\neq j}\left(\langle\overline{e_i+w_i^*},\overline{e_j+w_j}\rangle^2+\langle\overline{e_i+w_i},\overline{e_j+w_j}\rangle^2\right)\right)\right]^{\frac{1}{2}}$$

$$\overset{⑥}{\leq}1.01\left(\frac{4\gamma}{(1-\gamma)^2}+\frac{4\gamma(1+\gamma)}{1-2\gamma}\right)\|\mathbf{W}^*-\mathbf{W}\|_F^2\overset{⑦}{\leq}0.083\|\mathbf{W}^*-\mathbf{W}\|_F^2$$

where ① is by Taylor series, ② uses Cauchy Schwartz, ③ uses Lemma F.7, ④ holds as $\gamma\leq\frac{1}{100}$, ⑤ uses Cauchy Schwartz, ⑥ uses Lemma F.9, ⑦ holds as $\gamma\leq\frac{1}{100}$.

□

**Lemma* G.2** (Approximation loss related to $\sin\theta_{i,j},\sin\theta_{i^*,j}$)**.** *If* $\|\mathbf{W}\|_2,\|\mathbf{W}^*\|_2\leq\gamma\leq\frac{1}{100}$,

$$\sum_{j=1}^{d}\sum_{i\neq j}\Bigg|\Bigg(\|e_i+w_i^*\|_2\left(\sin\theta_{i^*,j}-1+\frac{1}{2}\langle\overline{e_i+w_i^*},\overline{e_j+w_j}\rangle^2\right)-$$
$$\|e_i+w_i\|_2\left(\sin\theta_{i,j}-1+\frac{1}{2}\langle\overline{e_i+w_i},\overline{e_j+w_j}\rangle^2\right)\Bigg)\langle\overline{e_j+w_j},w_j^*-w_j\rangle\Bigg|\leq0.002\|\mathbf{W}^*-\mathbf{W}\|_F^2$$

*Proof.* By definition, we know $\theta_{i^*,j}=\arccos\langle\overline{e_i+w_i^*},\overline{e_j+w_j}\rangle$, and $\theta_{i,j}=\arccos\langle\overline{e_i+w_i},\overline{e_j+w_j}\rangle$. The Taylor series of $\sin(\arccos x)$ at $x=0$ is $1-\frac{x^2}{2}-\frac{x^4}{8}-\frac{x^6}{16}-\frac{5x^8}{128}-\cdots=\sum_{k=0}^{\infty}c_kx^{2k}$, where $c_k\leq\frac{1}{8}$ for $k\geq2$.

Thus,

$$\sum_{j=1}^{d}\sum_{i\neq j}\Bigg|\Bigg(\|e_i+w_i^*\|_2\left(\sin\theta_{i^*,j}-1+\frac{1}{2}\langle\overline{e_i+w_i^*},\overline{e_j+w_j}\rangle^2\right)-$$

$$\|e_i+w_i\|_2\left(\sin\theta_{i,j}-1+\frac{1}{2}\langle\overline{e_i+w_i},\overline{e_j+w_j}\rangle^2\right)\Bigg)\langle\overline{e_j+w_j},w_j^*-w_j\rangle\Bigg|$$

$$\overset{①}{\leq}\sum_{j=1}^{d}\sum_{i\neq j}\left|\sum_{k=2}^{\infty}\frac{1}{8}\left(\|e_i+w_i\|_2\langle\overline{e_i+w_i},\overline{e_j+w_j}\rangle^{2k}-\|e_i+w_i^*\|_2\langle\overline{e_i+w_i^*},\overline{e_j+w_j}\rangle^{2k}\right)\right|\|w_j^*-w_j\|_2$$

$$\overset{②}{\leq}\sum_{j=1}^{d}\sum_{i\neq j}\sum_{k=2}^{\infty}(2.2\gamma)^{2k-3}\left(\langle\overline{e_i+w_i},\overline{e_j+w_j}\rangle^2+\langle\overline{e_i+w_i^*},\overline{e_j+w_j}\rangle^2\right)\|w_i^*-w_i\|_2\|w_j^*-w_j\|_2$$

$$\overset{③}{\leq}2.3\gamma\left(\sum_{j=1}^{d}\sum_{i\neq j}\left(\langle\overline{e_i+w_i},\overline{e_j+w_j}\rangle^2+\langle\overline{e_i+w_i^*},\overline{e_j+w_j}\rangle^2\right)\|w_i^*-w_i\|_2^2\right)^{\frac{1}{2}}$$

$$\left(\sum_{j=1}^{d}\sum_{i\neq j}\left(\langle\overline{e_i+w_i},\overline{e_j+w_j}\rangle^2+\langle\overline{e_i+w_i^*},\overline{e_j+w_j}\rangle^2\right)\|w_j^*-w_j\|_2^2\right)^{\frac{1}{2}}$$

$$\leq 2.3\gamma\left[\sum_{i=1}^{d}\|w_i^*-w_i\|_2^2\left(\sum_{j\neq i}\left(\langle\overline{e_i+w_i},\overline{e_j+w_j}\rangle^2+\langle\overline{e_i+w_i^*},\overline{e_j+w_j}\rangle^2\right)\right)\right]^{\frac{1}{2}}$$

$$\left[\sum_{j=1}^{d}\|w_j^*-w_j\|_2^2\left(\sum_{i\neq j}\left(\langle\overline{e_i+w_i},\overline{e_j+w_j}\rangle^2+\langle\overline{e_i+w_i^*},\overline{e_j+w_j}\rangle^2\right)\right)\right]^{\frac{1}{2}}$$

$$\overset{④}{\leq}2.3\gamma\left(\frac{4\gamma}{(1-\gamma)^2}+\frac{4\gamma(1+\gamma)}{1-2\gamma}\right)\|\mathbf{W}^*-\mathbf{W}\|_F^2\overset{⑤}{<}0.002\|\mathbf{W}^*-\mathbf{W}\|_F^2$$

where ① is by Taylor series, ② uses Lemma F.8 and Cauchy Schwartz, ③ uses Cauchy Schwartz and $\gamma\leq\frac{1}{100}$, ④ uses Lemma F.9, and ⑤ holds as $\gamma\leq\frac{1}{100}$. □

*Proof for Lemma B.2.* Combining the results from Lemma G.1 and Lemma G.2, the lemma follows. □

## G.3 Proof for Lemma B.3

Denote $\mathbf{\Delta}\triangleq\mathbf{P}+\nabla\mathsf{L}(\mathbf{W})$. This lemma is harder to prove than the previous one since we need to bound the spectral norm of a matrix $\mathbf{\Delta}$. First of all, we need to represent $\mathbf{\Delta}$. Again, the difference has two parts: approximation for $\theta_{i,j},\theta_{i^*,j}$, and $\sin\theta_{i,j},\sin\theta_{i^*,j}$. Denote the two parts as $\mathbf{\Delta}_1,\mathbf{\Delta}_2$, where $\mathbf{\Delta}=\mathbf{\Delta}_1+\mathbf{\Delta}_2$. From the proof of Lemma G.1, we know the $j$-th column of the first part is

$$\mathbf{\Delta}_{1,j}\triangleq\sum_{i\neq j}\sum_{k=1}^{\infty}\frac{(2k)!}{4^k(k!)^2(2k+1)}\left(\langle\overline{e_i+w_i^*},\overline{e_j+w_j}\rangle^{2k+1}(e_i+w_i^*)-\langle\overline{e_i+w_i},\overline{e_j+w_j}\rangle^{2k+1}(e_i+w_i)\right)$$

And the $j$-th column of the second part is

$$\mathbf{\Delta}_{2,j}\triangleq\sum_{i\neq j}\sum_{k=2}^{\infty}c_k\left(\|e_i+w_i\|_2\langle\overline{e_i+w_i},\overline{e_j+w_j}\rangle^{2k}-\|e_i+w_i^*\|_2\langle\overline{e_i+w_i^*},\overline{e_j+w_j}\rangle^{2k}\right)\overline{e_j+w_j}$$

Below we bound $\|\mathbf{\Delta}_1\|_2$ in Lemma G.3, and bounds $\|\mathbf{\Delta}_2\|_2$ in Lemma G.4.

**Lemma* G.3.** *If* $\|\mathbf{W}\|_2,\|\mathbf{W}^*\|_2\leq\gamma\leq\frac{1}{100}$, $\|\mathbf{\Delta}_1\|_2\leq 3.4\gamma^2$.

*Proof.* Define $\mathbf{U}, \mathbf{V}$ such that for $i = j, \mathbf{U}_{i,j} = \mathbf{V}_{i,j} = 0$, and for $i \neq j$,

$$\mathbf{U}_{i,j} = \sum_{k=1}^{\infty} \frac{(2k)!}{4^k (k!)^2 (2k+1)} \langle \overline{e_i + w_i^*}, \overline{e_j + w_j} \rangle^{2k+1}, \mathbf{V}_{i,j} = \sum_{k=1}^{\infty} \frac{(2k)!}{4^k (k!)^2 (2k+1)} \langle \overline{e_i + w_i}, \overline{e_j + w_j} \rangle^{2k+1}$$

By matrix multiplication,

$$\mathbf{\Delta}_1 = \sum_{i=1}^{d} [(\mathbf{I} + \mathbf{W}^*)_{*,i} \mathbf{U}_{i,*} - (\mathbf{I} + \mathbf{W})_{*,i} \mathbf{V}_{i,*}] = (\mathbf{I} + \mathbf{W}^*)\mathbf{U} - (\mathbf{I} + \mathbf{W})\mathbf{V} \tag{12}$$

So it suffices to bound $\|\mathbf{U}\|_2, \|\mathbf{V}\|_2$. For $i \neq j$,

$$|\mathbf{U}_{i,j}| = \left| \sum_{k=1}^{\infty} \frac{(2k)!}{4^k (k!)^2 (2k+1)} \langle \overline{e_i + w_i^*}, \overline{e_j + w_j} \rangle^{2k+1} \right| \overset{①}{\leq} \sum_{k=1}^{\infty} \frac{(2.1\gamma)^{2k-1}}{6} \langle \overline{e_i + w_i^*}, \overline{e_j + w_j} \rangle^2 \leq 0.4\gamma \langle \overline{e_i + w_i^*}, \overline{e_j + w_j} \rangle^2$$

where ① uses Lemma F.2 and (11). Now, we know

$$\|\mathbf{U}\|_1 \overset{①}{=} \max_j \sum_{i=1}^{d} |\mathbf{U}_{i,j}| \leq \max_j \sum_{i \neq j} 0.4\gamma \langle \overline{e_i + w_i^*}, \overline{e_j + w_j} \rangle^2 \overset{②}{\leq} \frac{1.6(1+\gamma)\gamma^2}{1 - 2\gamma} \leq 1.65\gamma^2$$

where ① is by definition, ② uses Lemma F.9. Similarly,

$$\|\mathbf{U}\|_\infty = \max_i \sum_{j=1}^{d} |\mathbf{U}_{i,j}| \leq \max_i \sum_{j \neq i} 0.4\gamma \langle \overline{e_i + w_i^*}, \overline{e_j + w_j} \rangle^2 \leq 1.65\gamma^2$$

By Hölder's inequality, we have

$$\|\mathbf{U}\|_2 \leq \sqrt{\|\mathbf{U}\|_1 \|\mathbf{U}\|_\infty} \leq 1.65\gamma^2$$

Now we do the same analysis for $\mathbf{V}$.

$$\begin{aligned} |\mathbf{V}_{i,j}| &= \left| \sum_{k=1}^{\infty} \frac{(2k)!}{4^k (k!)^2 (2k+1)} \langle \overline{e_i + w_i}, \overline{e_j + w_j} \rangle^{2k+1} \right| \\ &\leq \sum_{k=1}^{\infty} \frac{(2.1\gamma)^{2k-1}}{6} \langle \overline{e_i + w_i}, \overline{e_j + w_j} \rangle^2 \leq 0.4\gamma \langle \overline{e_i + w_i}, \overline{e_j + w_j} \rangle^2 \end{aligned}$$

Hence, $\|\mathbf{V}\|_1 = \max_j \sum_{i=1}^{d} |\mathbf{V}_{i,j}| \leq \max_j \sum_{i \neq j} 0.4\gamma \langle \overline{e_i + w_i}, \overline{e_j + w_j} \rangle^2 \leq 1.65\gamma^2$. Similarly, $\|\mathbf{V}\|_\infty \leq 1.65\gamma^2$, and by Hölder's inequality, $\|\mathbf{V}\|_2 \leq \sqrt{\|\mathbf{V}\|_1 \|\mathbf{V}\|_\infty} \leq 1.65\gamma^2$. Using (12), we get

$$\|\mathbf{\Delta}_1\|_2 \leq \|\mathbf{I} + \mathbf{W}^*\|_2 \|\mathbf{U}\|_2 + \|\mathbf{I} + \mathbf{W}\|_2 \|\mathbf{V}\|_2 \leq 2(1+\gamma)1.65\gamma^2 < 3.4\gamma^2$$ □

**Lemma* G.4.** *If* $\|\mathbf{W}\|_2, \|\mathbf{W}^*\|_2 \leq \gamma \leq \frac{1}{100}$, $\|\mathbf{\Delta}_2\|_2 \leq 6\gamma^3$.

*Proof.* By definition, we can write

$$\mathbf{\Delta}_2 = \overline{\mathbf{I} + \mathbf{W}} \text{Diag} \left\{ \sum_{i \neq j} \sum_{k=2}^{\infty} c_k \left( \|e_i + w_i\|_2 \langle \overline{e_i + w_i}, \overline{e_j + w_j} \rangle^{2k} - \|e_i + w_i^*\|_2 \langle \overline{e_i + w_i^*}, \overline{e_j + w_j} \rangle^{2k} \right) \right\}_{j=1}^{d}$$

So it suffices to bound the norm of the diagonal matrix, which is the maximum of the diagonal entries. For any $j \in [d]$, we have

$$\begin{aligned}
&\left|\sum_{i\neq j}\sum_{k=2}^{\infty} c_k \left(\|e_i+w_i\|_2\langle\overline{e_i+w_i},\overline{e_j+w_j}\rangle^{2k} - \|e_i+w_i^*\|_2\langle\overline{e_i+w_i^*},\overline{e_j+w_j}\rangle^{2k}\right)\right| \\
\leq&\sum_{i\neq j}\sum_{k=2}^{\infty}\frac{1}{8}\left(\|e_i+w_i\|_2\langle\overline{e_i+w_i},\overline{e_j+w_j}\rangle^{2k}| + |\|e_i+w_i^*\|_2\langle\overline{e_i+w_i^*},\overline{e_j+w_j}\rangle^{2k}\right) \\
\overset{①}{\leq}&\sum_{i\neq j}\sum_{k=2}^{\infty}\frac{1}{4}(1+\gamma)(2.1\gamma)^{2k-2}\left(\langle\overline{e_i+w_i},\overline{e_j+w_j}\rangle^2 + \langle\overline{e_i+w_i^*},\overline{e_j+w_j}\rangle^2\right) \\
\overset{②}{\leq}&0.6\gamma^2\sum_{i\neq j}\left(\langle\overline{e_i+w_i},\overline{e_j+w_j}\rangle^2 + \langle\overline{e_i+w_i^*},\overline{e_j+w_j}\rangle^2\right) \\
\overset{③}{\leq}&0.6\gamma^2\left(\frac{4\gamma}{(1-\gamma)^2}+\frac{4\gamma(1+\gamma)}{1-2\gamma}\right) < 5\gamma^3
\end{aligned}$$

where ① uses Lemma F.2, ② uses $\gamma \leq \frac{1}{100}$, ③ uses Lemma F.9. So we get $\|\mathbf{\Delta}_2\|_2 \leq \frac{1+\gamma}{1-\gamma}5\gamma^3 \leq 6\gamma^3$. □

*Proof for Lemma B.3.* Combining the results from Lemma G.3 and Lemma G.4, the lemma follows. □

# H Proofs for Section C

## H.1 Proof for Lemma C.1

In Lemma B.3, we use $\mathbf{P}(\mathbf{W})$ to approximate $-\nabla\mathsf{L}(\mathbf{W})$ in terms of spectral norm, with approximation loss $3.5\gamma^2$. Below we will get $\mathbf{Q}(\mathbf{W})$ from $\mathbf{P}(\mathbf{W})$ by removing a few more lower order terms.

By definition 2.3, we have

$$\begin{aligned}
\mathbf{P}_{2,j} =& g\overline{e_j+w_j} - (\|e_j+w_j^*\|_2 - \|e_j+w_j\|_2)\overline{e_j+w_j} + \left(\mathbf{I}-\frac{1}{2}\overline{e_j+w_j}\cdot\overline{e_j+w_j}^\top\right)\mathbf{A}\overline{e_j+w_j} \\
&+\left(\mathbf{I}-\frac{1}{2}\overline{e_j+w_j}\cdot\overline{e_j+w_j}^\top\right)(e_j+w_j) - \left(\mathbf{I}-\frac{1}{2}\overline{e_j+w_j}\cdot\overline{e_j+w_j}^\top\right)(e_j+w_j^*)\overline{e_j+w_j^*}^\top\overline{e_j+w_j} \\
=& g\overline{e_j+w_j} - (\|e_j+w_j^*\|_2 - \|e_j+w_j\|_2)\overline{e_j+w_j} + \left(\mathbf{I}-\frac{1}{2}\overline{e_j+w_j}\cdot\overline{e_j+w_j}^\top\right)\mathbf{A}\overline{e_j+w_j} \\
&+\frac{1}{2}(e_j+w_j) - (e_j+w_j^*)\overline{e_j+w_j^*}^\top\overline{e_j+w_j} + \frac{1}{2}\overline{e_j+w_j}\|e_j+w_j^*\|_2(\overline{e_j+w_j^*}^\top\overline{e_j+w_j})^2 \\
=& g\overline{e_j+w_j} + \left(\mathbf{I}-\frac{1}{2}\overline{e_j+w_j}\cdot\overline{e_j+w_j}^\top\right)\mathbf{A}\overline{e_j+w_j} + \frac{3}{2}(e_j+w_j) - \overline{e_j+w_j^*}^\top\overline{e_j+w_j}(e_j+w_j^*) \\
&+\left(\frac{1}{2}\|e_j+w_j^*\|_2(\overline{e_j+w_j^*}^\top\overline{e_j+w_j})^2 - \|e_j+w_j^*\|_2\right)\overline{e_j+w_j} \\
=& g\overline{e_j+w_j} + \left(\mathbf{I}-\frac{1}{2}\overline{e_j+w_j}\cdot\overline{e_j+w_j}^\top\right)\mathbf{A}\overline{e_j+w_j} - w_j^* + w_j + (1-\overline{e_j+w_j^*}^\top\overline{e_j+w_j})(e_j+w_j^*) \\
&+\left(\frac{1}{2}\|e_j+w_j\|_2 + \frac{1}{2}\|e_j+w_j^*\|_2(\overline{e_j+w_j^*}^\top\overline{e_j+w_j})^2 - \|e_j+w_j^*\|_2\right)\overline{e_j+w_j}
\end{aligned}$$

Combining every column together, we get

$$\mathbf{P}_2 = g\overline{\mathbf{I}+\mathbf{W}} + \mathbf{A}\overline{\mathbf{I}+\mathbf{W}} - \frac{1}{2}\overline{\mathbf{I}+\mathbf{W}}\mathrm{Diag}(\{\overline{e_j+w_j}^\top\mathbf{A}\overline{e_j+w_j}\}_{j=1}^d) - (\mathbf{W}^*-\mathbf{W}) + \overline{\mathbf{I}+\mathbf{W}^*}\mathbf{\Sigma}_1 + \overline{\mathbf{I}+\mathbf{W}}\mathbf{\Sigma}_2$$

where

$$\boldsymbol{\Sigma}_1 = \mathrm{Diag}(\{(\|e_j + w_j^*\|_2 - \|e_j + w_j^*\|_2 \overline{e_j + w_j^*}^\top \overline{e_j + w_j})\}_{j=1}^d)$$
$$\boldsymbol{\Sigma}_2 = \mathrm{Diag}(\{\frac{1}{2}\|e_j + w_j\|_2 + \frac{1}{2}\|e_j + w_j^*\|_2 (\overline{e_j + w_j^*}^\top \overline{e_j + w_j})^2 - \|e_j + w_j^*\|_2\}_{j=1}^d)$$

Using Lemma F.12, we replace $\overline{e_j + w_j}^\top \mathbf{A} \overline{e_j + w_j}$ with $e_j^\top \mathbf{A} e_j$. By Lemma F.1,

$$\left\| \mathbf{P}_2 - \left[ g\overline{\mathbf{I} + \mathbf{W}} + \mathbf{A}\overline{\mathbf{I} + \mathbf{W}} - \frac{1}{2}\overline{\mathbf{I} + \mathbf{W}}\mathrm{Diag}(\mathbf{A}) - (\mathbf{W}^* - \mathbf{W}) + \overline{\mathbf{I} + \mathbf{W}^*}\boldsymbol{\Sigma}_1 + \overline{\mathbf{I} + \mathbf{W}}\boldsymbol{\Sigma}_2 \right] \right\|_2 \leq \frac{5(1+\gamma)}{2(1-\gamma)} < 2.6\gamma^2$$

We then focus on the middle two summands in the sum.

$$\mathbf{A}\overline{\mathbf{I} + \mathbf{W}} - \frac{1}{2}\overline{\mathbf{I} + \mathbf{W}}\mathrm{Diag}(\mathbf{A}) = (\mathbf{A} - \frac{1}{2}\mathrm{Diag}(\mathbf{A})) + \mathbf{A}(\overline{\mathbf{I} + \mathbf{W}} - \mathbf{I}) - \frac{1}{2}(\overline{\mathbf{I} + \mathbf{W}} - \mathbf{I})\mathrm{Diag}(\mathbf{A})$$

By Lemma F.10, $\|\mathrm{Diag}(\mathbf{A})\|_2 \leq \|\mathbf{A}\|_2$, so

$$\left\| \left[ \mathbf{A}\overline{\mathbf{I} + \mathbf{W}} - \frac{1}{2}\overline{\mathbf{I} + \mathbf{W}}\mathrm{Diag}(\mathbf{A}) \right] - \left[ \mathbf{A} - \frac{1}{2}\mathrm{Diag}(\mathbf{A}) \right] \right\|_2 = \left\| \mathbf{A}(\overline{\mathbf{I} + \mathbf{W}} - \mathbf{I}) - \frac{1}{2}(\overline{\mathbf{I} + \mathbf{W}} - \mathbf{I})\mathrm{Diag}(\mathbf{A}) \right\|_2$$
$$\leq \|\mathbf{A}\|_2 \|\overline{\mathbf{I} + \mathbf{W}} - \mathbf{I}\|_2 + \frac{1}{2}\|\overline{\mathbf{I} + \mathbf{W}} - \mathbf{I}\|_2 \|\mathrm{Diag}(\mathbf{A})\|_2 \overset{①}{\leq} \frac{3\gamma(\gamma^2+3)}{1-\gamma^2} 2.05\gamma < 18.5\gamma^2$$

where ① uses Lemma F.11 and Lemma F.15.

Moreover, by Lemma E.1 term 2, we know $\|\boldsymbol{\Sigma}_1\|_2 \leq \max_{i \in [d]} (1+\gamma) \frac{\|w_i^* - w_i\|_2^2}{2(1-2\gamma)} \leq 2.07\gamma^2$, and in $\boldsymbol{\Sigma}_2$,

$$\left| \frac{1}{2}\|e_j + w_j^*\|_2 (\overline{e_j + w_j^*}^\top \overline{e_j + w_j})^2 - \frac{1}{2}\|e_j + w_j^*\|_2 \right| \leq \frac{1}{2}(1+\gamma) \left| \overline{e_j + w_j^*}^\top \overline{e_j + w_j} - 1 \right| \left| \overline{e_j + w_j^*}^\top \overline{e_j + w_j} + 1 \right| \leq 2.07\gamma^2$$

so the following terms approximates $\mathbf{P}_2$ with approximation loss $(2.6 + 18.5 + 2.07 + 2.07)\gamma^2 < 25.3\gamma^2$.

$$\overline{\mathbf{I} + \mathbf{W}}(g\mathbf{I} - \boldsymbol{\Sigma}_3) + \mathbf{A} - \frac{1}{2}\mathrm{Diag}(\mathbf{A}) - (\mathbf{W}^* - \mathbf{W})$$

where $\boldsymbol{\Sigma}_3 = \mathrm{Diag}(\{\frac{1}{2}\|e_j + w_j^*\|_2 - \frac{1}{2}\|e_j + w_j\|_2\}_{j=1}^d)$.

By Lemma F.16 and Corollary F.17, we know $\|\mathbf{A} - [\mathbf{W}^* - \mathbf{W} + (\mathbf{W}^* - \mathbf{W})^\top - \mathrm{Diag}(\mathbf{W}^* - \mathbf{W})]\|_2 \leq 9.2\gamma^2$ and $\|\mathrm{Diag}(\mathbf{A}) - \mathrm{Diag}(\mathbf{W}^* - \mathbf{W})\|_2 \leq 9.2\gamma^2$. Therefore, with approximation loss of $18.4\gamma^2$, we get

$$\left\| \left[ \mathbf{A} - \frac{1}{2}\mathrm{Diag}(\mathbf{A}) \right] - \left[ \mathbf{W}^* - \mathbf{W} + (\mathbf{W}^* - \mathbf{W})^\top - \frac{3}{2}\mathrm{Diag}(\mathbf{W}^* - \mathbf{W}) \right] \right\|_2 \leq 18.4\gamma^2$$

We then approximate $\boldsymbol{\Sigma}_3$:

$$\|(\overline{\mathbf{I} + \mathbf{W}})\boldsymbol{\Sigma}_3 - (\overline{\mathbf{I} + \mathbf{W}})\frac{1}{2}\mathrm{Diag}(\mathbf{W}^* - \mathbf{W})\|_2 \leq \frac{1+\gamma}{1-\gamma} \left( \frac{1}{2} \max_j |\|e_j + w_j^*\|_2 - \|e_j + w_j\|_2 - w_{j,j}^* + w_{j,j}| \right) < 3.1\gamma^2$$

where the last inequality is by Lemma F.13. Moreover,

$$\|\overline{\mathbf{I} + \mathbf{W}} \left( \frac{1}{2}\mathrm{Diag}(\mathbf{W}^* - \mathbf{W}) \right) - \frac{1}{2}\mathrm{Diag}(\mathbf{W}^* - \mathbf{W})\|_2$$
$$\leq \|\overline{\mathbf{I} + \mathbf{W}} - \mathbf{I}\|_2 \left\| \frac{1}{2}\mathrm{Diag}(\mathbf{W}^* - \mathbf{W}) \right\|_2 < 2.05\gamma \left( \frac{1}{2} \max_i |w_{i,i}^* - w_{i,i}| \right) < 2.05\gamma^2$$

Putting everything together, with approximation loss of $(25.3 + 18.4 + 3.1 + 2.05)\gamma^2 = 49\gamma^2$ to $\mathbf{P}_2$, we get

$$(\mathbf{W}^* - \mathbf{W})^\top - 2\mathrm{Diag}(\mathbf{W}^* - \mathbf{W}) + g\overline{\mathbf{I} + \mathbf{W}}$$

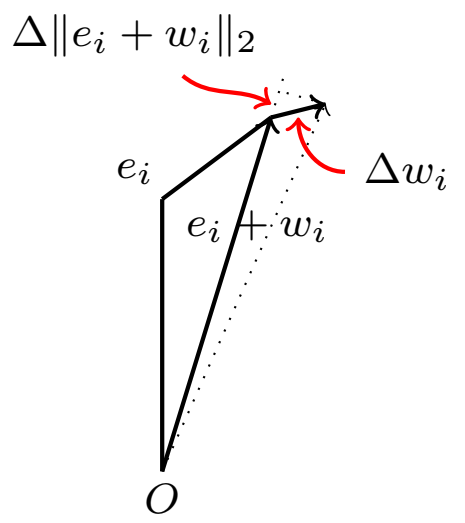

Figure 11: $\Delta g$ is approximately (the summation of) the projection of $\Delta w_i$ onto $\overline{e_i + w_i}$

For $\mathbf{P}_3$, using the same idea in the proof of Lemma D.3, we have

$$\mathbf{P}_3 = \frac{\pi}{2}(\mathbf{W}^* - \mathbf{W}) + \left(\overline{\mathbf{I}+\mathbf{W}} - \overline{\mathbf{I}+\mathbf{W}^*}\right)\mathbf{\Sigma}_4 + \overline{\mathbf{I}+\mathbf{W}}\mathbf{\Sigma}_5$$

where $\mathbf{\Sigma}_4 = \text{Diag}(\{\theta_{j,j^*}\|e_j + w_j^*\|_2\}_{j=1}^d), \mathbf{\Sigma}_5 = \text{Diag}(\{\|e_j + w_j^*\|_2 \sin\theta_{j,j^*} - \theta_{j,j^*}\|e_j + w_j^*\|_2\}_{j=1}^d)$. By Taylor's Theorem, we know $\|\mathbf{\Sigma}_5\|_2 \le \|\text{Diag}(\{\|e_j + w_j^*\|_2\theta_{j,j^*}^3/3\}_{j=1}^d)\|_2$.

Notice that $\theta_{j,j^*} \le 2.002\gamma$ by Lemma E.1 term 3, and $\|\overline{\mathbf{I}+\mathbf{W}} - \overline{\mathbf{I}+\mathbf{W}^*}\|_2 \le \frac{1+\gamma}{1-\gamma} - \frac{1-\gamma}{1+\gamma} \le 4.001\gamma$.

Consequently,

$$\begin{aligned}\left\|\mathbf{P}_3 - \frac{\pi}{2}(\mathbf{W}^* - \mathbf{W})\right\|_2 &\le \|\left(\overline{\mathbf{I}+\mathbf{W}} - \overline{\mathbf{I}+\mathbf{W}^*}\right)\mathbf{\Sigma}_4\|_2 + \|\overline{\mathbf{I}+\mathbf{W}}\mathbf{\Sigma}_5\|_2 \\ &< 4.001 * 2.002(1+\gamma)\gamma^2 + \frac{(1+\gamma)^2}{3(1-\gamma)}(2.002\gamma)^3 < 8.1\gamma^2 + 2.8\gamma^3 < 8.2\gamma^2\end{aligned}$$

we only need to keep the term $\frac{\pi}{2}(\mathbf{W}^* - \mathbf{W})$ with approximation loss $8.2\gamma^2$ to $\mathbf{P}_3$.

Now, combining the approximations to $\mathbf{P}_2$ and $\mathbf{P}_3$, and Lemma B.3, we have the following matrix with $(49 + 8.2 + 3.5)\gamma^2 < 61\gamma^2$ approximation loss to $-\nabla \mathsf{L}(\mathbf{W})$:

$$\frac{\pi}{2}(\mathbf{W}^* - \mathbf{W})\left(\mathbf{I} + uu^\top\right) + (\mathbf{W}^* - \mathbf{W})^\top - 2\text{Diag}(\mathbf{W}^* - \mathbf{W}) + g\overline{\mathbf{I}+\mathbf{W}}$$

where $u$ is the all 1 vector.

## H.2 Proof for Lemma C.2

By Lemma F.4, we know $|g| \le 2d\gamma$. Using Lemma C.1,

$$\begin{aligned}\|\nabla \mathsf{L}(\mathbf{W})\|_2 &\le 61\gamma^2 + \left\|\frac{\pi}{2}(\mathbf{W}^* - \mathbf{W})\left(\mathbf{I} + uu^\top\right) + (\mathbf{W}^* - \mathbf{W})^\top - 2\text{Diag}(\mathbf{W}^* - \mathbf{W}) + g\overline{\mathbf{I}+\mathbf{W}}\right\|_2 \\ &\le 61\gamma^2 + (d+1)\pi\gamma + 2\gamma + 4\gamma + |g|\frac{1+\gamma}{1-\gamma} < 61\gamma^2 + (d+3)\pi\gamma + 2.05d\gamma < 6d\gamma.\end{aligned}$$

## H.3 Proof for Lemma C.3

In this proof, we use $w_j$ to represent the $j$-th column of $\mathbf{W}_t$, and denote $\triangle w_j$ as the $j$-th column of $\mathbf{G}_t$.

### H.3.1 $\Delta g_t \approx \eta\langle \mathsf{L}(\mathbf{W}_t), \overline{\mathbf{I}+\mathbf{W}_t}\rangle$

For the intuition of this section, see Figure 11. The changes in potential function $g$ is essentially the changes in $\|e_i + w_i\|_2$ (summing over $i$), which is approximately $\Delta w_i$ projected onto $\overline{e_i + w_i}$. If we write it in matrix form, we get $\Delta g_t \approx \eta\langle \mathsf{L}(\mathbf{W}_t), \overline{\mathbf{I}+\mathbf{W}_t}\rangle$.

By definition we know $\|\mathbf{G}_t\|_2 = \|\nabla \mathsf{L}(\mathbf{W}_t) + \mathbf{E}_t\|_2 \overset{①}{\le} \|\nabla \mathsf{L}(\mathbf{W}_t)\|_2 + \|\mathbf{E}_t\|_2 \overset{②}{\le} 6d\gamma + \varepsilon = G_2$, where ① uses triangle inequality, ② uses Lemma C.2. We have

$$\eta\|\triangle w_j\|_2 \le \eta\|\mathbf{G}_t\|_2 \le \frac{\gamma^2}{G_2} \le \frac{\gamma}{6d}, \qquad \eta^2\|\triangle w_j\|_2 \le \eta\|\mathbf{G}_t\|_2^2 \le \gamma^2 \tag{13}$$

By Definition 2.2, we know

$$\triangle g_t \triangleq g_{t+1} - g_t = \sum_{j=1}^{d}\left(\frac{\langle e_j + w_j, e_j + w_j\rangle}{\|e_j + w_j\|_2} - \frac{\langle e_j + w_j - \eta\triangle w_j, e_j + w_j - \eta\triangle w_j\rangle}{\|e_j + w_j - \eta\triangle w_j\|_2}\right)$$
$$= \sum_{j=1}^{d}\left(\frac{\langle e_j + w_j, e_j + w_j\rangle\|e_j + w_j - \eta\triangle w_j\|_2 - \langle e_j + w_j - \eta\triangle w_j, e_j + w_j - \eta\triangle w_j\rangle\|e_j + w_j\|_2}{\|e_j + w_j\|_2\|e_j + w_j - \eta\triangle w_j\|_2}\right)$$
$$= \sum_{j=1}^{d}\left(\frac{\|e_j + w_j\|_2(\|e_j + w_j - \eta\triangle w_j\|_2 - \|e_j + w_j\|_2) + 2\eta\langle\triangle w_j, e_j + w_j\rangle - \eta^2\|\triangle w_j\|_2^2}{\|e_j + w_j - \eta\triangle w_j\|_2}\right)$$

If we project $\eta\triangle w_j$ onto the $\overline{e_j + w_j}$ direction, we get

$$\|e_j + w_j - \eta\triangle w_j\|_2 = \sqrt{(\|e_j + w_j\|_2 - \langle\overline{e_j + w_j}, \eta\triangle w_j\rangle)^2 + (\|\eta\triangle_j\|_2^2 - \langle\overline{e_j + w_j}, \eta\triangle w_j\rangle^2)^2}$$
$$\le\sqrt{(\|e_j + w_j\|_2 - \langle\overline{e_j + w_j}, \eta\triangle w_j\rangle)^2 + \|\eta\triangle w_j\|_2^2} \overset{①}{\le} \|e_j + w_j\|_2 - \langle\overline{e_j + w_j}, \eta\triangle w_j\rangle + \|\eta\triangle w_j\|_2^2$$

Using (13), we have $\|e_j + w_j\|_2 - \langle\overline{e_j + w_j}, \eta\triangle w_j\rangle \ge \frac{1}{2}$. By taking square on both sides, we know ① holds. It is trivial to show that $\|e_j + w_j - \eta\triangle w_j\|_2 \ge \|e_j + w_j\|_2 - \langle\overline{e_j + w_j}, \eta\triangle w_j\rangle$, so we know

$$-\langle\overline{e_j + w_j}, \eta\triangle w_j\rangle \le \|e_j + w_j - \eta\triangle w_j\|_2 - \|e_j + w_j\|_2 \le -\langle\overline{e_j + w_j}, \eta\triangle w_j\rangle + \|\eta\triangle w_j\|_2^2 \tag{14}$$

Thus, with approximation loss $\sum_{j=1}^{d}\frac{\|e_j+w_j\|_2\|\eta\triangle w_j\|_2^2}{\|e_j+w_j-\eta\triangle w_j\|_2}$, we have :

$$\triangle g_t \approx \sum_{j=1}^{d}\left(\frac{-\|e_j + w_j\|_2\langle\overline{e_j + w_j}, \eta\triangle w_j\rangle + 2\eta\langle\triangle w_j, e_j + w_j\rangle - \eta^2\|\triangle w_j\|_2^2}{\|e_j + w_j - \eta\triangle w_j\|_2}\right)$$
$$= \sum_{j=1}^{d}\frac{\eta\langle\triangle w_j, e_j + w_j\rangle - \eta^2\|\triangle w_j\|_2^2}{\|e_j + w_j - \eta\triangle w_j\|_2}$$
$$= \sum_{j=1}^{d}\frac{-\eta^2\|\triangle w_j\|_2^2}{\|e_j + w_j - \eta\triangle w_j\|_2} + \sum_{j=1}^{d}\frac{(\|e_j + w_j\|_2 - \|e_j + w_j - \eta\triangle w_j\|_2)\eta\langle\triangle w_j, \overline{e_j + w_j}\rangle}{\|e_j + w_j - \eta\triangle w_j\|_2} + \eta\langle\mathbf{G}_t, \overline{\mathbf{I} + \mathbf{W}_t}\rangle$$

Thus we get the following approximation for $\triangle g_t$.

$$|\triangle g_t - \eta\langle\mathbf{G}_t, \overline{\mathbf{I} + \mathbf{W}_t}\rangle|$$
$$\le \sum_{j=1}^{d}\left|\frac{-\eta^2\|\triangle w_j\|_2^2}{\|e_j + w_j - \eta\triangle w_j\|_2} + \frac{(\|e_j + w_j\|_2 - \|e_j + w_j - \eta\triangle w_j\|_2)\eta\langle\triangle w_j, \overline{e_j + w_j}\rangle}{\|e_j + w_j - \eta\triangle w_j\|_2} + \frac{\|e_j + w_j\|_2\|\eta\triangle w_j\|_2^2}{\|e_j + w_j - \eta\triangle w_j\|_2}\right|$$
$$\overset{①}{\le} \sum_{j=1}^{d}\left[\left|\frac{\eta\langle\triangle w_j, \overline{e_j + w_j}\rangle(\eta\langle\triangle w_j, \overline{e_j + w_j}\rangle + \|\eta\triangle w_j\|_2^2)}{\|e_j + w_j - \eta\triangle w_j\|_2}\right| + 0.02\eta^2\|\triangle w_j\|_2^2\right]$$
$$\overset{②}{\le} \sum_{j=1}^{d}\left[\frac{\eta^2\|\triangle w_j\|_2^2 + \eta^3\|\triangle w_j\|_2^3}{\|e_j + w_j - \eta\triangle w_j\|_2} + 0.02\eta\gamma^2\right] \overset{③}{\le} 1.04\eta d\gamma^2$$

where ① uses (14) again, and ② ③ uses (13), $\gamma \le \frac{1}{100}$ and $\|e_j + w_j - \eta\triangle w_j\|_2 \ge 0.98$.

Thus $|\triangle g_t - \eta\langle\nabla \mathsf{L}(\mathbf{W}_t), \overline{\mathbf{I} + \mathbf{W}_t}\rangle| \le 1.04\eta d\gamma^2 + |\eta\langle\mathbf{E}_t, \overline{\mathbf{I} + \mathbf{W}_t}\rangle| < 1.04\eta d\gamma^2 + 1.03\eta\sqrt{d}\varepsilon$

#### H.3.2 $\Delta g_t \approx \eta \text{Tr}(\nabla \mathsf{L}(\mathbf{W}_t))$

We want to approximate $\overline{\mathbf{I}+\mathbf{W}_t}$ with $\mathbf{I}$. Below is the error bound.

$$
\begin{aligned}
&|\langle \nabla \mathsf{L}(\mathbf{W}_t), \overline{\mathbf{I}+\mathbf{W}_t} - \mathbf{I}\rangle| = |\langle \nabla \mathsf{L}(\mathbf{W}_t) + \mathbf{Q}_t - \mathbf{Q}_t, \overline{\mathbf{I}+\mathbf{W}_t} - \mathbf{I}\rangle| \\
&\overset{①}{=} d \cdot 61\gamma^2 \cdot 2.05\gamma + \sum_{i=1}^{d} 2.05\gamma \left\| (\mathbf{Q}_t - \frac{\pi}{2}(\mathbf{W}^* - \mathbf{W}_t)uu^\top)_i \right\|_2 + \left\langle \frac{\pi}{2}(\mathbf{W}^* - \mathbf{W}_t)uu^\top, \overline{\mathbf{I}+\mathbf{W}_t} - \mathbf{I} \right\rangle \\
&\overset{②}{\le} 1.251 d\gamma^2 + 2.05 d\gamma \left( \pi\gamma + 2\gamma + 4\gamma + \frac{1+\gamma}{1-\gamma}|g_t| \right) + \text{Tr}\left( \left[ \frac{\pi}{2}(\mathbf{W}^* - \mathbf{W}_t)u \right] \left[ u^\top \overline{\mathbf{I}+\mathbf{W}_t} - \mathbf{I} \right]^\top \right) \\
&\overset{③}{\le} 20 d\gamma^2 + 2.1 d\gamma |g_t| + \left\| \frac{\pi}{2}(\mathbf{W}^* - \mathbf{W}_t)u \right\|_2 \left\| (\overline{\mathbf{I}+\mathbf{W}_t} - \mathbf{I})u \right\|_2 \overset{④}{\le} 20 d\gamma^2 + 2.1 d\gamma |g_t| + \frac{2.05\pi}{2}\|s\|_2 \gamma \sqrt{d}
\end{aligned}
$$

where ① uses Cauchy Schwartz and Lemma F.15, ② uses the definition of $\mathbf{Q}$ and Lemma F.1, ③ holds as for any vector $u, v$, $\text{Tr}(uv^\top) \le \|u\|_2 \|v\|_2$, ④ uses Lemma F.15.

Hence,

$$
\begin{aligned}
&|\triangle g_t - \eta \langle \nabla \mathsf{L}(\mathbf{W}_t), \mathbf{I} \rangle| \\
&\le 1.04 \eta d\gamma^2 + 1.03 \eta \sqrt{d}\varepsilon + |\eta \langle \nabla \mathsf{L}(\mathbf{W}_t), \overline{\mathbf{I}+\mathbf{W}_t} - \mathbf{I}\rangle| \\
&< 1.04 \eta d\gamma^2 + 1.03 \eta \sqrt{d}\varepsilon + 20 \eta d\gamma^2 + 2.1 \eta d\gamma |g_t| + \frac{2.05\pi}{2}\eta \|s\|_2 \gamma \sqrt{d} \\
&< 21.1 \eta d\gamma^2 + 1.03 \eta \sqrt{d}\varepsilon + 2.1 \eta d\gamma |g_t| + \frac{2.05\pi}{2}\eta \|s\|_2 \gamma \sqrt{d}
\end{aligned}
$$

So with approximation loss of $21.1\eta d\gamma^2 + 1.03\eta\sqrt{d}\varepsilon + 2.1\eta d\gamma|g_t| + \frac{2.05\pi}{2}\eta\|s\|_2\gamma\sqrt{d}$, it suffices to consider $\eta \text{Tr}(\nabla \mathsf{L}(\mathbf{W}_t))$.

#### H.3.3 $\Delta g_t \approx -\eta(d + \frac{\pi}{2} - 1) g_t$

According to Lemma C.1, with approximation loss of $61\gamma^2$, we can use $-\mathbf{Q}_t$ to approximate $\nabla \mathsf{L}(\mathbf{W}_t)$.

$$
\begin{aligned}
&\text{Tr}(\mathbf{Q}_t) = \frac{\pi}{2}\text{Tr}\left( (\mathbf{W}^* - \mathbf{W}_t)\left(\mathbf{I} + uu^\top\right) \right) + \text{Tr}(\mathbf{W}^* - \mathbf{W}_t)^\top - 2\text{Tr}(\text{Diag}(\mathbf{W}^* - \mathbf{W}_t)) + g\text{Tr}(\overline{\mathbf{I}+\mathbf{W}_t}) \\
&= \left( \frac{\pi}{2} - 1 \right) \text{Tr}(\mathbf{W}^* - \mathbf{W}_t) + \frac{\pi}{2}\text{Tr}\left( (\mathbf{W}^* - \mathbf{W}_t)\left(uu^\top\right) \right) + g\text{Tr}(\overline{\mathbf{I}+\mathbf{W}_t}) \\
&= \left( \frac{\pi}{2} - 1 \right) (\text{Tr}(\mathbf{W}^* - \mathbf{W}_t) - g_t) + \left( \frac{\pi}{2} - 1 \right) g_t + \frac{\pi}{2}\text{Tr}\left( (\mathbf{W}^* - \mathbf{W}_t)\left(uu^\top\right) \right) + g_t \text{Tr}(\overline{\mathbf{I}+\mathbf{W}_t})
\end{aligned}
$$

Therefore,

$$
\begin{aligned}
&\left| \text{Tr}(\mathbf{Q}_t) - g_t \text{Tr}(\mathbf{I}) - \left( \frac{\pi}{2} - 1 \right) g_t \right| = \left| \text{Tr}(\mathbf{Q}_t) - \left( d + \frac{\pi}{2} - 1 \right) g_t \right| \\
&\le \left| \left( \frac{\pi}{2} - 1 \right) (\text{Tr}(\mathbf{W}^* - \mathbf{W}_t) - g_t) + \frac{\pi}{2}\text{Tr}\left( (\mathbf{W}^* - \mathbf{W}_t)\left(uu^\top\right) \right) + g_t(\text{Tr}(\overline{\mathbf{I}+\mathbf{W}_t} - \mathbf{I})) \right| \\
&\overset{①}{\le} 6.07 \left( \frac{\pi}{2} - 1 \right) d\gamma^2 + \frac{\pi}{2}\|s_t\|_2 \sqrt{d} + 2.05|g_t| d\gamma
\end{aligned}
$$

where ① uses Lemma F.14 and Lemma F.15. Thus,

$$
\begin{aligned}
&\left| \triangle g_t - \left[ -\eta \left( d + \frac{\pi}{2} - 1 \right) g_t \right] \right| \\
&\le \eta \left[ 21.1 d\gamma^2 + 1.03\sqrt{d}\varepsilon + 2.1 d\gamma |g_t| + \frac{2.05\pi}{2}\|s\|_2 \gamma \sqrt{d} + 61 d\gamma^2 + 2.05|g_t| d\gamma + 6.07 \left( \frac{\pi}{2} - 1 \right) d\gamma^2 + \frac{\pi}{2}\|s_t\|_2 \sqrt{d} \right] \\
&\le \eta \left[ 86 d\gamma^2 + 1.03\sqrt{d}\varepsilon + 4.15 d\gamma |g_t| + 4.8 \|s_t\|_2 \gamma \sqrt{d} \right]
\end{aligned}
$$

Now we have

$$|g_{t+1}| = |g_t + \triangle g_t| \leq \left(1 - \eta\left(d + \frac{\pi}{2} - 1 - 4.15d\gamma\right)\right)|g_t| + 86\eta d\gamma^2 + 1.03\eta\sqrt{d}\varepsilon + 4.8\eta\|s_t\|_2\gamma\sqrt{d}$$
$$\leq (1 - 0.95\eta d)|g_t| + 86\eta d\gamma^2 + 1.03\eta\sqrt{d}\varepsilon + 4.8\eta\|s_t\|_2\gamma\sqrt{d}$$

## H.4 Proof for Lemma C.4

By definition of $s_t$,

$$\triangle s_t \triangleq s_{t+1} - s_t = (\mathbf{W}_t - \mathbf{W}_{t+1})u = \eta(\nabla \mathsf{L}(\mathbf{W}_t) + \mathbf{E}_t)u = -\eta\mathbf{Q}_t u + \eta(\mathbf{Q}_t + \nabla \mathsf{L}(\mathbf{W}_t) + \mathbf{E}_t)u$$

By definition of $\mathbf{Q}_t$,

$$\mathbf{Q}_t u = \left(\frac{\pi}{2}(\mathbf{W}^* - \mathbf{W}_t)\left(\mathbf{I} + uu^\top\right) + (\mathbf{W}^* - \mathbf{W}_t)^\top - 2\mathrm{Diag}(\mathbf{W}^* - \mathbf{W}_t) + g_t\overline{\mathbf{I} + \mathbf{W}_t}\right)u$$
$$= \frac{(d+1)\pi}{2}s_t + \left((\mathbf{W}^* - \mathbf{W}_t)^\top - 2\mathrm{Diag}(\mathbf{W}^* - \mathbf{W}_t) + g_t\overline{\mathbf{I} + \mathbf{W}_t}\right)u$$

Thus, we know

$$\left\|\mathbf{Q}_t u - \frac{(d+1)\pi}{2}s_t\right\|_2 = \left\|\left((\mathbf{W}^* - \mathbf{W}_t)^\top - 2\mathrm{Diag}(\mathbf{W}^* - \mathbf{W}_t) + g_t\overline{\mathbf{I} + \mathbf{W}_t}\right)u\right\|_2$$
$$\leq \sqrt{d}\left(\|(\mathbf{W}^* - \mathbf{W}_t)^\top\|_2 + 2\|\mathrm{Diag}(\mathbf{W}^* - \mathbf{W}_t)\|_2 + \|g_t\overline{\mathbf{I} + \mathbf{W}_t}\|_2\right)$$
$$\overset{①}{\leq} \sqrt{d}\left(2\gamma + 4\gamma + |g_t|\frac{1+\gamma}{1-\gamma}\right) < (6\gamma + 1.03|g_t|)\sqrt{d}$$

where ① uses Lemma F.1 and Lemma F.10.

By Lemma C.1, $\|\triangle s_t - [-\eta\frac{(d+1)\pi}{2}s_t]\|_2 < \eta(6\gamma + 1.03|g_t|)\sqrt{d} + \eta\|(\mathbf{Q}_t + \nabla\mathsf{L}(\mathbf{W}_t) + \mathbf{E}_t)u\|_2 \leq \eta(6.61\gamma + 1.03|g_t| + \varepsilon)\sqrt{d}$.

## H.5 Proof for Lemma C.5

Combining Lemma C.3 and Lemma C.4, we get

$$|g_{t+1}| + \|s_{t+1}\|_2$$
$$\leq (1 - 0.95\eta d)(|g_t| + \|s_t\|_2) + \eta(6.6\gamma + 1.03|g_t| + \varepsilon)\sqrt{d} + 86\eta d\gamma^2 + 1.03\eta\sqrt{d}\varepsilon + (4.8\eta\gamma\sqrt{d} - 0.62\eta d)\|s_t\|_2$$
$$\overset{①}{\leq} (1 - 0.95\eta d)(|g_t| + \|s_t\|_2) + 6.6\eta\gamma\sqrt{d} + 86\eta d\gamma^2 + \eta 1.03|g_t|\sqrt{d} + 2.03\eta\sqrt{d}\varepsilon$$
$$\overset{②}{\leq} (1 - 0.84\eta d)(|g_t| + \|s_t\|_2) + 6.6\eta\gamma\sqrt{d} + 87\eta d\gamma^2$$

where ① uses $\gamma \leq \frac{1}{100}$, $d \geq 100$, ② uses $\varepsilon \leq \gamma^2$ and $d \geq 100$. So if the following inequality holds, $|g_t| + \|s_t\|_2$ will always decrease by factor at least $1 - 0.5\eta d$.

$$0.34\eta d(|g_t| + \|s_t\|_2) \geq 6.6\eta\gamma\sqrt{d} + 87\eta d\gamma^2$$

Which gives

$$|g_t| + \|s_t\|_2 \geq \frac{6.6\eta\gamma\sqrt{d} + 87\eta d\gamma^2}{0.34\eta d} = \frac{6.6\gamma}{0.34\sqrt{d}} + \frac{87\gamma^2}{0.34}$$

where the last expression is smaller than $4.5\gamma$. Hence, $|g_t| + \|s_t\|_2$ will keep decreasing by $1 - 0.5\eta d$ as long as it is larger than $4.5\gamma$. So we have $\|s_t\|_2 \leq 4.5\gamma$. Now plug it back to the updating rule of $|g_t|$:

$$|g_{t+1}| \leq (1 - 0.95\eta d)|g_t| + 86\eta d\gamma^2 + 1.03\eta\sqrt{d}\varepsilon + 4.8\eta\|s_t\|_2\gamma\sqrt{d}$$
$$\leq (1 - 0.95\eta d)|g_t| + 86\eta d\gamma^2 + 1.03\eta\sqrt{d}\varepsilon + 21.6\eta\gamma^2\sqrt{d}$$

In order to get factor $1-0.5\eta d$, we have

$$0.45\eta d|g_t| \geq 86\eta d\gamma^2 + 1.03\eta\sqrt{d}\varepsilon + 21.6\eta\gamma^2\sqrt{d}$$

Solve this inequality, we get

$$\frac{86\eta d\gamma^2 + 1.03\eta\sqrt{d}\varepsilon + 21.6\eta\gamma^2\sqrt{d}}{0.45\eta d} = \frac{86\gamma^2}{0.45} + \frac{1.03\varepsilon + 21.6\gamma^2}{0.45\sqrt{d}} \leq 197\gamma^2$$

The last inequality uses $d \geq 100, \varepsilon \leq \gamma^2$. So even after $|g_t| + \|s_t\|_2$ is below $4.5\gamma$, $|g_t|$ will keep decreasing by factor $1-0.5\eta d$ until it is smaller than $197\gamma^2$.

Finally we bound the number of steps to arrive $197\gamma^2$. Let $\gamma = \frac{1}{400}, \gamma_0 = \frac{1}{8000}$. Again, the constants here are pretty loose. Since $|g_t| \leq (1-0.5\eta d)^t|g_0| \leq (1-0.5\eta d)^t 2d\gamma_0$, in order to let $g_t \leq 197\gamma^2$, it suffices to have $t \geq \frac{\log\frac{197\gamma^2}{2d\gamma_0}}{\log(1-\frac{\eta d}{2})}$. Since $\eta d$ is small, by Taylor expansion we know $\log(1-\frac{\eta d}{2}) \approx -\frac{\eta d}{2}$. Thus, it suffices to let $t \geq \frac{2\log(0.203d)}{\eta d}$. Notice that $\frac{\log(0.203d)}{d}$ is decreasing for $d \geq 100$, we know it suffices to let $t \geq \frac{1}{16\eta}$.

## H.6 Proof for Lemma C.6

Let $\mathbf{H} = \mathbf{W} - \mathbf{W}^*$, by the updating rule of $\mathbf{W}_t$ and the definition of $\mathbf{Q}_t$, we know

$$\mathbf{H}_{t+1} = \mathbf{H}_t - \eta\mathbf{H}_t\left(\frac{\pi}{2}uu^\top + \frac{\pi}{2}\right) - \eta\mathbf{H}_t^\top + 2\eta\mathrm{Diag}(\mathbf{H}_t) + \eta g_t\overline{\mathbf{I}+\mathbf{W}} - \eta(\mathbf{G}_t + \mathbf{Q}_t)$$

That gives,

$$\begin{aligned}
&\|\mathbf{H}_{t+1} + \mathbf{H}_{t+1}^\top\|_2 \\
\leq &\left\|(\mathbf{H}_t + \mathbf{H}_t^\top)\left(\mathbf{I} - \eta\left(\frac{\pi}{2}uu^\top + \frac{\pi}{2} + 1\right)\right)\right\|_2 + 2\eta\left\|\mathrm{Diag}(\mathbf{H}_t + \mathbf{H}_t^\top)\right\|_2 + 2\eta|g_t|\|\overline{\mathbf{I}+\mathbf{W}}\|_2 + 2\eta\left\|\mathbf{E}_t + \nabla \mathsf{L}(\mathbf{W}_t) + \mathbf{Q}_t\right\|_2 \\
\overset{①}{\leq} &\left(\mathbf{I} - \eta\left(\frac{\pi}{2} + 1\right)\right)\|\mathbf{H}_t + \mathbf{H}_t^\top\|_2 + 2\eta\|\mathbf{H}_t + \mathbf{H}_t^\top\|_2 + \frac{2(1+\gamma)\eta|g_t|}{1-\gamma} + 2\eta\varepsilon + 122\eta\gamma^2 \\
\overset{②}{\leq} &\left(\mathbf{I} - \eta\left(\frac{\pi}{2} - 1\right)\right)\|\mathbf{H}_t + \mathbf{H}_t^\top\|_2 + 2.05\eta|g_t| + 124\eta\gamma^2 \qquad (15)
\end{aligned}$$

where ① uses Lemma F.18, Lemma F.10, $\|\mathbf{E}_t\|_2 \leq \varepsilon$ and Lemma C.1. ② uses $\varepsilon \leq \gamma^2$ and $\gamma \leq \frac{1}{100}$.

Similarly, we get

$$\begin{aligned}
&\|\mathbf{H}_{t+1} - \mathbf{H}_{t+1}^\top\|_2 \\
\overset{①}{\leq} &\left\|(\mathbf{H}_t - \mathbf{H}_t^\top)\left(\mathbf{I} - \eta\left(\frac{\pi}{2}uu^\top + \frac{\pi}{2} - 1\right)\right)\right\|_2 + \eta|g_t|\|\overline{\mathbf{I}+\mathbf{W}} - \mathbf{I} + \mathbf{I} - \overline{\mathbf{I}+\mathbf{W}}^\top\|_2 + 2\eta\left\|\mathbf{E}_t + \nabla \mathsf{L}(\mathbf{W}_t) + \mathbf{Q}_t\right\|_2 \\
\overset{②}{\leq} &\left(\mathbf{I} - \eta\left(\frac{\pi}{2} - 1\right)\right)\|\mathbf{H}_t - \mathbf{H}_t^\top\|_2 + 4.10\eta\gamma|g_t| + 124\eta\gamma^2 \qquad (16)
\end{aligned}$$

where ① holds as the diagonal terms cancel out, ② uses Lemma F.18, Lemma F.15.

Adding (15) and (16), we get

$$\begin{aligned}
&\|\mathbf{H}_{t+1} + \mathbf{H}_{t+1}^\top\|_2 + \|\mathbf{H}_{t+1} - \mathbf{H}_{t+1}^\top\|_2 \\
\leq &\left(\mathbf{I} - \eta\left(\frac{\pi}{2} - 1\right)\right)\left(\|\mathbf{H}_t + \mathbf{H}_t^\top\|_2 + \|\mathbf{H}_t - \mathbf{H}_t^\top\|_2\right) + 2.1\eta|g_t| + 248\eta\gamma^2 \qquad (17)
\end{aligned}$$

For any $T > 0$, by applying (17) recursively, we have

$$\|\mathbf{H}_T + \mathbf{H}_T^\top\|_2 + \|\mathbf{H}_T - \mathbf{H}_T^\top\|_2 \leq \|\mathbf{H}_0 + \mathbf{H}_0^\top\|_2 + \|\mathbf{H}_0 - \mathbf{H}_0^\top\|_2 + 2.1\eta\sum_{t=0}^{T-1}|g_t| + 248\eta T\gamma^2$$

By Lemma F.4 we know $|g_0| \le 2d\gamma_0$, so $2.1\eta \sum_{t=0}^{T-1} |g_t| \le \frac{2.1\eta|g_0|(1-(1-0.5\eta d)^T)}{(0.5\eta d)} \le \frac{4.2|g_0|}{d} \le 8.4\gamma_0$.

By the proof of Lemma C.5, we know $T \le \frac{1}{16\eta}$, so $248\eta T\gamma^2 \le 15.5\gamma^2$.

By triangle inequality, we know $\|\mathbf{H}_0\|_2 \le \|\mathbf{W}_0\|_2 + \|\mathbf{W}^*\|_2 \le 2\gamma_0$, so $\|\mathbf{H}_0 + \mathbf{H}_0^\top\|_2 + \|\mathbf{H}_0 - \mathbf{H}_0^\top\|_2 \le 4\|\mathbf{H}_0\|_2 \le 8\gamma_0$.

By triangle inequality again we get

$$\|\mathbf{H}_T\|_2 \le \|\mathbf{H}_T + \mathbf{H}_T^\top\|_2 + \|\mathbf{H}_T - \mathbf{H}_T^\top\|_2 \le \|\mathbf{H}_0 + \mathbf{H}_0^\top\|_2 + \|\mathbf{H}_0 - \mathbf{H}_0^\top\|_2 + 19\gamma^2 + 8.4\gamma_0 \le 16.4\gamma_0 + 15.5\gamma^2$$

Recall we set $\gamma = \frac{1}{400}, \gamma_0 = \frac{1}{8000}$ in the proof of Lemma C.5, we know $\|\mathbf{W}_T\|_2 \le \|\mathbf{W}^*\|_2 + \|\mathbf{H}_T\|_2 \le 17.4\gamma_0 + 15.5\gamma^2 \le \frac{1}{440} \le \gamma$.

### H.7 Proof for Lemma C.7

First, by the proof of Lemma C.5, we know $|g_t|$ will keep small if $\|\mathbf{W}_t\|_2 \le \gamma \le \frac{1}{100}$.

Adding (15) and (16), we get

$$\begin{aligned}
&\|\mathbf{H}_{t+1} + \mathbf{H}_{t+1}^\top\|_2 + \|\mathbf{H}_{t+1} - \mathbf{H}_{t+1}^\top\|_2 \\
\le &\left(\mathbf{I} - \eta\left(\frac{\pi}{2} - 1\right)\right)\left(\|\mathbf{H}_{t+1} + \mathbf{H}_{t+1}^\top\|_2 + \|\mathbf{H}_{t+1} - \mathbf{H}_{t+1}^\top\|_2\right) + 2.1\eta|g_t| + 248\eta\gamma^2 \\
\overset{①}{\le} &\left(\mathbf{I} - \eta\left(\frac{\pi}{2} - 1\right)\right)\left(\|\mathbf{H}_{t+1} + \mathbf{H}_{t+1}^\top\|_2 + \|\mathbf{H}_{t+1} - \mathbf{H}_{t+1}^\top\|_2\right) + 661\eta\gamma^2 \qquad (18)
\end{aligned}$$

where ① holds as $|g_t| \le 197\gamma^2$. So either $\|\mathbf{H}_{t+1} + \mathbf{H}_{t+1}^\top\|_2 + \|\mathbf{H}_{t+1} - \mathbf{H}_{t+1}^\top\|_2$ keeps decreasing, or it increases, i.e.,

$$\eta\left(\frac{\pi}{2} - 1\right)\left(\|\mathbf{H}_{t+1} + \mathbf{H}_{t+1}^\top\|_2 + \|\mathbf{H}_{t+1} - \mathbf{H}_{t+1}^\top\|_2\right) \le 197\eta\gamma^2$$

That gives,

$$\|\mathbf{H}_{t+1} + \mathbf{H}_{t+1}^\top\|_2 + \|\mathbf{H}_{t+1} - \mathbf{H}_{t+1}^\top\|_2 \le \frac{197\gamma^2}{\frac{\pi}{2} - 1} \le 346\gamma^2$$

Therefore, combined with the proof of Lemma C.6, we know $\|\mathbf{H}_{t+1} + \mathbf{H}_{t+1}^\top\|_2 + \|\mathbf{H}_{t+1} - \mathbf{H}_{t+1}^\top\|_2$ will keep decreasing until it is at most $346\gamma^2$. Now,

$$\|\mathbf{W}_t\|_2 \le \|\mathbf{H}_t\|_2 + \|\mathbf{W}^*\|_2 \le \|\mathbf{H}_{t+1} + \mathbf{H}_{t+1}^\top\|_2 + \|\mathbf{H}_{t+1} - \mathbf{H}_{t+1}^\top\|_2 + \gamma_0 \overset{①}{\le} (346 + 20)\gamma^2 \le \gamma$$

where ① holds as $\gamma_0 = \frac{1}{8000}$. So $\|\mathbf{W}_t\|_2$ is always bounded by $\gamma$.

## I Proofs for Section D

For notational simplicity, denote

$$\begin{aligned}
x_j &\triangleq \left(\overline{e_j + w_j} \cdot \overline{e_j + w_j}^\top\right)(w_j^* - w_j), \\
\mathbf{X} &\triangleq (x_1, \cdots, x_d) \qquad (19) \\
y_j &\triangleq \left(\mathbf{I} - \overline{e_j + w_j} \cdot \overline{e_j + w_j}^\top\right)(w_j^* - w_j), \\
\mathbf{Y} &\triangleq (y_1, \cdots, y_d) \qquad (20) \\
z_j &\triangleq \left(\mathbf{I} - \frac{1}{2}\overline{e_j + w_j} \cdot \overline{e_j + w_j}^\top\right)(w_j^* - w_j), \\
\mathbf{Z} &\triangleq (z_1, \cdots, z_d)
\end{aligned}$$

We have the following relationship between $x_j, y_j, z_j$.

**Lemma I.1.**

$$\|z_j\|_2^2 = \frac{1}{4}\|x_j\|_2^2 + \|y_j\|_2^2, \;\; \|x_j\|_2^2 + \|y_j\|_2^2 = \|w_j^* - w_j\|_2^2 \tag{21}$$

*Proof for Lemma I.1.* By definition,

$$\begin{aligned}
\|z_j\|_2^2 =& \|w_j^* - w_j\|_2^2 \left(\mathbf{I} - \frac{1}{2}\overline{e_j + w_j} \cdot \overline{e_j + w_j}^\top\right)^\top \left(\mathbf{I} - \frac{1}{2}\overline{e_j + w_j} \cdot \overline{e_j + w_j}^\top\right) \\
=& \|w_j^* - w_j\|_2^2 \left(\mathbf{I} - \overline{e_j + w_j} \cdot \overline{e_j + w_j}^\top + \frac{1}{4}\overline{e_j + w_j} \cdot \overline{e_j + w_j}^\top \overline{e_j + w_j} \cdot \overline{e_j + w_j}^\top\right) \\
=& \|w_j^* - w_j\|_2^2 \left(\mathbf{I} - \frac{3}{4}\overline{e_j + w_j} \cdot \overline{e_j + w_j}^\top\right),
\end{aligned}$$

and similarly

$$\begin{aligned}
\|y_j\|_2^2 =& \|w_j^* - w_j\|_2^2 \left(\mathbf{I} - \overline{e_j + w_j} \cdot \overline{e_j + w_j}^\top\right)^\top \left(\mathbf{I} - \overline{e_j + w_j} \cdot \overline{e_j + w_j}^\top\right) = \|w_j^* - w_j\|_2^2 \left(\mathbf{I} - \overline{e_j + w_j} \cdot \overline{e_j + w_j}^\top\right), \\
\|x_j\|_2^2 =& \|w_j^* - w_j\|_2^2 \left(\overline{e_j + w_j} \cdot \overline{e_j + w_j}^\top\right)^\top \left(\overline{e_j + w_j} \cdot \overline{e_j + w_j}^\top\right) = \|w_j^* - w_j\|_2^2 \left(\overline{e_j + w_j} \cdot \overline{e_j + w_j}^\top\right)
\end{aligned}$$

The lemma follows. □

## I.1 Proof for Lemma D.1

In this proof, we heavily use the following trick between the summation of four vector products, and the trace of four matrix products. We give one example below, and other cases are similar.

**Lemma I.2.** $\sum_{i,j} z_j^\top (e_i + w_i^*)(\overline{e_i + w_i^*} - \overline{e_i + w_i})^\top \overline{e_j + w_j} = \mathrm{Tr}\left(\left[\mathbf{Z}^\top(\mathbf{I} + \mathbf{W}^*)\right]\left[(\overline{\mathbf{I} + \mathbf{W}^*} - \overline{\mathbf{I} + \mathbf{W}})^\top \overline{\mathbf{I} + \mathbf{W}}\right]\right)$.

*Proof.* By definition, $\mathrm{Tr}(\mathbf{AB}) = \sum_{j=1}^d (\mathbf{AB})_{j,j} = \sum_{i,j} \mathbf{A}_{j,i}\mathbf{B}_{i,j}$. Thus,

$$\mathrm{Tr}\left(\left[\mathbf{Z}^\top(\mathbf{I} + \mathbf{W}^*)\right]\left[(\overline{\mathbf{I} + \mathbf{W}^*} - \overline{\mathbf{I} + \mathbf{W}})^\top \overline{\mathbf{I} + \mathbf{W}}\right]\right) = \sum_{i,j} \left[\mathbf{Z}^\top(\mathbf{I} + \mathbf{W}^*)\right]_{j,i} \left[(\overline{\mathbf{I} + \mathbf{W}^*} - \overline{\mathbf{I} + \mathbf{W}})^\top \overline{\mathbf{I} + \mathbf{W}}\right]_{i,j}$$

By definition, $\left[\mathbf{Z}^\top(\mathbf{I} + \mathbf{W}^*)\right]_{j,i} = z_j^\top(e_i + w_i^*)$, and $\left[(\overline{\mathbf{I} + \mathbf{W}^*} - \overline{\mathbf{I} + \mathbf{W}})^\top \overline{\mathbf{I} + \mathbf{W}}\right]_{i,j} = (\overline{e_i + w_i^*} - \overline{e_i + w_i})^\top \overline{e_j + w_j}$, so the lemma follows. □

Now we proceed to prove Lemma D.1. We first bound $\sum_{j=1}^d z_j^\top \mathbf{A}_j \overline{e_j + w_j}$ below by splitting $\mathbf{A}_j$ into three parts, and then improve the lower bound in Lemma I.4.

**Lemma I.3.** *If $\|\mathbf{W}\|_2, \|\mathbf{W}^*\|_2 \le \gamma \le \frac{1}{100}$, we have*

$$\sum_{j=1}^d z_j^\top \mathbf{A}_j \overline{e_j + w_j} \ge -8\gamma\|\mathbf{W}^* - \mathbf{W}\|_F^2 - \sqrt{\|\mathbf{W}^* - \mathbf{W}\|_f^2 - \frac{3}{4}\|\mathbf{X}\|_F^2}\sqrt{\|\mathbf{W}^* - \mathbf{W}\|_F^2 - \|\mathbf{X}\|_F^2}$$

.

*Proof.* We rewrite $\mathbf{A}_j$ as

$$\mathbf{A}_j = \mathbf{B}_j + \frac{1}{2}\mathbf{C}_j + \mathbf{D}_j \tag{22}$$

where

$$\mathbf{B}_j = \sum_{i \ne j}(e_i + w_i^*)(\overline{e_i + w_i^*} - \overline{e_i + w_i})^\top, \quad \mathbf{C}_j = \sum_{i \ne j}\langle w_i^* - w_i, \overline{e_i + w_i}\rangle \overline{e_i + w_i} \cdot \overline{e_i + w_i}^\top, \quad \mathbf{D}_j = \left(\sum_{i \ne j} z_i \overline{e_i + w_i}^\top\right)$$

For notational simplicity, we also write $\mathbf{B}, \mathbf{C}, \mathbf{D}$ as the corresponding terms with sum $\sum_{i=1}^{d}$ instead of $\sum_{i\neq j}$, so they do not depend on index $j$. We estimate $\mathbf{B}, \mathbf{C}, \mathbf{D}$ first, then estimate $\mathbf{B}_j, \mathbf{C}_j, \mathbf{D}_j$ respectively by taking the differences.

**1.** From $\mathbf{B}$ to $\mathbf{B}_j$:

$$\sum_{j=1}^{d} z_j^\top \mathbf{B}\overline{e_j+w_j} = \sum_{i,j} z_j^\top (e_i+w_i^*)(\overline{e_i+w_i^*} - \overline{e_i+w_i})^\top \overline{e_j+w_j}$$

$$\overset{①}{=}\mathrm{Tr}\left(\left[\mathbf{Z}^\top(\mathbf{I}+\mathbf{W})\right]\left[(\overline{\mathbf{I}+\mathbf{W}^*} - \overline{\mathbf{I}+\mathbf{W}})^\top \overline{\mathbf{I}+\mathbf{W}}\right]\right) \overset{②}{\geq} -\left\|(\mathbf{I}+\mathbf{W})^\top \mathbf{Z}\right\|_F \left\|\overline{\mathbf{I}+\mathbf{W}}^\top(\overline{\mathbf{I}+\mathbf{W}^*} - \overline{\mathbf{I}+\mathbf{W}})\right\|_F$$

$$\overset{③}{\geq} -\|\mathbf{I}+\mathbf{W}\|_2 \|\overline{\mathbf{I}+\mathbf{W}}\|_2 \left\|\mathbf{Z}\right\|_F \left\|\overline{\mathbf{I}+\mathbf{W}^*} - \overline{\mathbf{I}+\mathbf{W}}\right\|_F \overset{④}{\geq} -\frac{(1+\gamma)^2}{1-\gamma}\left\|\mathbf{Z}\right\|_F \left\|\overline{\mathbf{I}+\mathbf{W}^*} - \overline{\mathbf{I}+\mathbf{W}}\right\|_F \tag{23}$$

where ① uses Lemma I.2, ② uses $\mathrm{Tr}(\mathbf{AB}) \geq -\|\mathbf{A}\|_F\|\mathbf{B}\|_F$, ③ uses $\|\mathbf{AB}\|_F \leq \|\mathbf{A}\|_2\|\mathbf{B}\|_F$, and ④ uses Lemma F.1. By Lemma E.1 term 1, we have

$$\left\|\overline{\mathbf{I}+\mathbf{W}^*} - \overline{\mathbf{I}+\mathbf{W}}\right\|_F \leq \sqrt{\frac{\sum_{i=1}^{d}\|y_i\|_2^2}{1-2\gamma}} = \frac{\|\mathbf{Y}\|_F}{\sqrt{1-2\gamma}} \tag{24}$$

On the other hand,

$$\sum_{j=1}^{d} z_j^\top(\mathbf{B}_j - \mathbf{B})\overline{e_j+w_j} = \sum_{j=1}^{d} z_j^\top (e_j+w_j^*)(\overline{e_j+w_j^*} - \overline{e_j+w_j})^\top \overline{e_j+w_j}$$

$$=\sum_{j=1}^{d}(w_j^*-w_j)^\top(\mathbf{I} - \frac{1}{2}\overline{e_j+w_j}\cdot\overline{e_j+w_j}^\top)(e_j+w_j^*)(\overline{e_j+w_j^*} - \overline{e_j+w_j})^\top\overline{e_j+w_j}$$

For any vector $x$, $\overline{e_j+w_j}\cdot\overline{e_j+w_j}^\top x$ is the projection of $x$ onto the direction $\overline{e_j+w_j}$, so $\frac{1}{2} \leq \|\mathbf{I} - \frac{1}{2}\overline{e_j+w_j}\cdot\overline{e_j+w_j}^\top\|_2 \leq 1$, and

$$|(w_j^*-w_j)^\top(e_j+w_j^*)(\overline{e_j+w_j^*} - \overline{e_j+w_j})^\top\overline{e_j+w_j}| \overset{①}{\leq} |(w_j^*-w_j)^\top(e_j+w_j^*)|\frac{\|w_j^*-w_j\|_2^2}{2(1-2\gamma)}$$

$$\overset{②}{\leq}\frac{\|w_j^*-w_j\|_2^3(1+\gamma)}{2(1-2\gamma)} \leq \frac{\|w_j^*-w_j\|_2^2(1+\gamma)\gamma}{1-2\gamma} \tag{25}$$

where ① uses Lemma E.1 term 2, and ② uses Cauchy-Schwartz.

Combining (23),(24),(25), we get

$$\sum_{j=1}^{d} z_j^\top \mathbf{B}_j\overline{e_j+w_j} \geq -\frac{(1+\gamma)^2}{(1-\gamma)\sqrt{1-2\gamma}}\left\|\mathbf{Z}\right\|_F \left\|\mathbf{Y}\right\|_F - \frac{(1+\gamma)\gamma}{1-2\gamma}\|\mathbf{W}^*-\mathbf{W}\|_F^2$$

**2.** From $\mathbf{C}$ to $\mathbf{C}_j$:

$$\sum_{j=1}^{d} z_j^\top \mathbf{C}\overline{e_j+w_j} = \sum_{i,j} z_j^\top \langle w_i^*-w_i, \overline{e_i+w_i}\rangle \overline{e_i+w_i}\cdot\overline{e_i+w_i}^\top\overline{e_j+w_j}$$

$$\overset{①}{=}\mathrm{Tr}(\left[\mathbf{Z}^\top\mathbf{X}\right]\left[\overline{\mathbf{I}+\mathbf{W}}^\top\overline{\mathbf{I}+\mathbf{W}}\right]) = \mathrm{Tr}(\mathbf{Z}^\top\mathbf{X}) + \mathrm{Tr}(\mathbf{Z}^\top\mathbf{X}(\overline{\mathbf{I}+\mathbf{W}}^\top\overline{\mathbf{I}+\mathbf{W}} - \mathbf{I}))$$

$$\overset{②}{\geq}\mathrm{Tr}(\mathbf{Z}^\top\mathbf{X}) - \|\mathbf{Z}\|_F\|\mathbf{X}\|_F\|\overline{\mathbf{I}+\mathbf{W}}^\top\overline{\mathbf{I}+\mathbf{W}} - \mathbf{I}\|_2 \overset{③}{\geq} \mathrm{Tr}(\mathbf{Z}^\top\mathbf{X}) - \frac{4\gamma}{(1-\gamma)^2}\|\mathbf{Z}\|_F\|\mathbf{X}\|_F$$

where ① uses Lemma I.2 and $x_j = \langle w_j^* - w_j, \overline{e_j + w_j}\rangle \overline{e_j + w_j}$, ② uses $\mathrm{Tr}(\mathbf{AB}) \geq -\|\mathbf{A}\|_F\|\mathbf{B}\|_F$, and $\|\mathbf{AB}\|_F \leq \|\mathbf{A}\|_2\|\mathbf{B}\|_F$, and ③ uses Lemma F.1. On the other hand,

$$\sum_{j=1}^{d} z_j^\top (\mathbf{C} - \mathbf{C}_j)\overline{e_j + w_j} = \sum_{j=1}^{d} z_j^\top \langle w_j^* - w_j, \overline{e_j + w_j}\rangle \overline{e_j + w_j} \cdot \overline{e_j + w_j}^\top \overline{e_j + w_j}$$

$$= \sum_{j=1}^{d} z_j^\top \langle w_j^* - w_j, \overline{e_j + w_j}\rangle \overline{e_j + w_j} = \mathrm{Tr}(\mathbf{Z}^\top \mathbf{X})$$

That implies, $\frac{1}{2}\sum_{j=1}^{d} z_j^\top \mathbf{C}_j \overline{e_j + w_j} \geq -\frac{2\gamma}{(1-\gamma)^2}\|\mathbf{Z}\|_F\|\mathbf{X}\|_F$.

**3.** From $\mathbf{D}$ to $\mathbf{D}_j$:

$$\sum_{j=1}^{d} z_j^\top \mathbf{D}\overline{e_j + w_j} = \sum_{i,j} z_j^\top z_i \overline{e_i + w_i}^\top \overline{e_j + w_j} = \mathrm{Tr}\left(\left[\mathbf{Z}^\top \mathbf{Z}\right]\left[\overline{\mathbf{I} + \mathbf{W}}^\top \overline{\mathbf{I} + \mathbf{W}}\right]\right) \geq \frac{(1-\gamma)^2}{(1+\gamma)^2}\|\mathbf{Z}\|_F^2$$

where the last inequality holds by Lemma F.1. On the other hand,

$$z_j^\top (\mathbf{D} - \mathbf{D}_j)\overline{e_j + w_j} = \|z_j\|_2^2$$

That gives,

$$\sum_{j} z_j^\top \mathbf{D}_j \overline{e_j + w_j} \geq -\frac{4\gamma}{(1+\gamma)^2}\|\mathbf{Z}\|_F^2$$

Now, combining $\mathbf{B}_j, \mathbf{C}_j, \mathbf{D}_j$ together, using (22), we have

$$\sum_{j=1}^{d} z_j^\top \mathbf{A}_j \overline{e_j + w_j} \geq -\frac{(1+\gamma)^2}{(1-\gamma)\sqrt{1-2\gamma}}\|\mathbf{Z}\|_F\|\mathbf{Y}\|_F - \frac{(1+\gamma)\gamma}{1-2\gamma}\|\mathbf{W}^* - \mathbf{W}\|_F^2$$
$$- \frac{2\gamma}{(1-\gamma)^2}\|\mathbf{Z}\|_F\|\mathbf{X}\|_F - \frac{4\gamma}{(1+\gamma)^2}\|\mathbf{Z}\|_F^2$$

By definition, we know $\|\mathbf{X}\|_F \leq \|\mathbf{W}^* - \mathbf{W}\|_F, \|\mathbf{Y}\|_F \leq \|\mathbf{W}^* - \mathbf{W}\|_F, \|\mathbf{Z}\|_F \leq \|\mathbf{W}^* - \mathbf{W}\|_F$, and $\gamma \leq \frac{1}{100}$, so

$$-\frac{(1+\gamma)\gamma}{1-2\gamma}\|\mathbf{W}^* - \mathbf{W}\|_F^2 - \frac{2\gamma}{(1-\gamma)^2}\|\mathbf{Z}\|_F\|\mathbf{X}\|_F - \frac{4\gamma}{(1+\gamma)^2}\|\mathbf{Z}\|_F^2 \geq -7\gamma\|\mathbf{W}^* - \mathbf{W}\|_F^2 \tag{26}$$

Moreover,

$$-\left(\frac{(1+\gamma)^2}{(1-\gamma)\sqrt{1-2\gamma}} - 1\right)\|\mathbf{Z}\|_F\|\mathbf{Y}\|_F \geq -0.05\gamma\|\mathbf{W}^* - \mathbf{W}\|_F^2 \tag{27}$$

Thus, those are small order terms. The only term left is $\|\mathbf{Z}\|_F\|\mathbf{Y}\|_F$. By (21), we know

$$\|\mathbf{Z}\|_F\|\mathbf{Y}\|_F \leq \sqrt{\|\mathbf{W}^* - \mathbf{W}\|_F^2 - \frac{3}{4}\|\mathbf{X}\|_F^2}\sqrt{\|\mathbf{W}^* - \mathbf{W}\|_F^2 - \|\mathbf{X}\|_F^2} \tag{28}$$

Combining (26), (27), (28), we get:

$$\sum_{j=1}^{d} z_j^\top \mathbf{A}_j \overline{e_j + w_j} \geq -8\gamma\|\mathbf{W}^* - \mathbf{W}\|_F^2 - \sqrt{\|\mathbf{W}^* - \mathbf{W}\|_f^2 - \frac{3}{4}\|\mathbf{X}\|_F^2}\sqrt{\|\mathbf{W}^* - \mathbf{W}\|_F^2 - \|\mathbf{X}\|_F^2} \qquad \square$$

Now it remains to bound $\sqrt{\|\mathbf{W}^* - \mathbf{W}\|_f^2 - \frac{3}{4}\|\mathbf{X}\|_F^2}\sqrt{\|\mathbf{W}^* - \mathbf{W}\|_F^2 - \|\mathbf{X}\|_F^2}$.

**Lemma I.4.**

$$-\sqrt{\|\mathbf{W}^* - \mathbf{W}\|_F^2 - \frac{3}{4}\|\mathbf{X}\|_F^2}\sqrt{\|\mathbf{W}^* - \mathbf{W}\|_F^2 - \|\mathbf{X}\|_F^2} \geq -1.3\|\mathbf{W}^* - \mathbf{W}\|_F^2 + \|\mathbf{W}^* - \mathbf{W}\|_F\|\mathbf{X}\|_F$$

*Proof.* Consider the function $f(x) = \sqrt{y^2 - \frac{3}{4}x^2}\sqrt{y^2 - x^2} + xy$, where $x \in [0, y]$. It suffices to show that $f(x) \leq 1.3y^2$.

Indeed, we know

$$f'(x) = \frac{x(6x^2 - 7y^2)}{2\sqrt{4y^2 - 3x^2}\sqrt{y^2 - x^2}} + y$$

When $x = 0$, $f'(x) = y > 0$, and when $x \to y$, $f'(x) < 0$. We want to find the place where $f'(x) = 0$, which gives the maximum value. Assume $x = \lambda y$, this is equivalent to solve

$$\lambda y(6(\lambda y)^2 - 7y^2) = -2y\sqrt{4y^2 - 3(\lambda y)^2}\sqrt{y^2 - (\lambda y)^2}$$

Cancel all $y$, and we get the solution $x \approx 0.566y$, where $f(x) \approx 1.2845y^2 < 1.3y^2$. □

*Proof of Lemma D.1.* Combining Lemma I.3 and Lemma I.4, we have proved Lemma D.1. □

## I.2 Proof for Lemma D.2

Again, we first consider the full sum, $g = \sum_{i=1}^d (\|e_i + w_i^*\|_2 - \|e_i + w_i\|_2)$.

By Lemma F.3, we have

$$|g - g_j| = |\|e_j + w_j^*\|_2 - \|e_j + w_j\|_2| \leq \|w_j^* - w_j\|_2$$

Thus by Cauchy Schwartz,

$$|(g - g_j)\langle w_j^* - w_j, \overline{e_j + w_j}\rangle| \leq \|w_j^* - w_j\|_2\|x_j\|_2$$

Summing over $j$, we get

$$\sum_{j=1}^d |(g - g_j)\langle w_j^* - w_j, \overline{e_j + w_j}\rangle| \leq \sum_{j=1}^d \|w_j^* - w_j\|_2\|x_j\|_2 \leq \|\mathbf{W}^* - \mathbf{W}\|_F\|\mathbf{X}\|_F \tag{29}$$

where the last inequality is by Cauchy Schwartz.

Now

$$\begin{aligned} g\sum_{j=1}^d \langle w_j^* - w_j, \overline{e_j + w_j}\rangle &= g\sum_{j=1}^d \langle e_j + w_j^* - e_j + w_j, \overline{e_j + w_j}\rangle \\ &= g\sum_{j=1}^d (\|e_j + w_j^*\|_2 - \|e_j + w_j\|_2 + \langle e_j + w_j^*, \overline{e_j + w_j} - \overline{e_j + w_j^*}\rangle) = g^2 + gb \geq gb \end{aligned} \tag{30}$$

where $b$ is defined to be $\sum_{j=1}^d \langle e_j + w_j^*, \overline{e_j + w_j} - \overline{e_j + w_j^*}\rangle$. By Lemma E.1 term 2 we know

$$-\frac{(1+\gamma)\|\mathbf{W}^* - \mathbf{W}\|_F^2}{2(1 - 2\gamma)} \leq b \leq 0$$

Combining (29), (30), the lemma follows.

$$\begin{aligned} \sum_{j=1}^d \langle g_j\overline{e_j + w_j}, w_j^* - w_j\rangle &= \sum_{j=1}^d \langle (g_j - g)\overline{e_j + w_j}, w_j^* - w_j\rangle + \sum_{j=1}^d \langle g\overline{e_j + w_j}, w_j^* - w_j\rangle \\ &\geq -\|\mathbf{W}^* - \mathbf{W}\|_F\|\mathbf{X}\|_F + g^2 + gb \geq -\|\mathbf{W}^* - \mathbf{W}\|_F\|\mathbf{X}\|_F - \frac{(1+\gamma)g\|\mathbf{W}^* - \mathbf{W}\|_F^2}{2(1-2\gamma)} \end{aligned}$$

### I.3 Proof for Lemma D.3

$$\sum_{j=1}^{d}\langle \mathbf{P}_{3,j}, w_j^* - w_j\rangle = \sum_{j=1}^{d}\langle \frac{\pi}{2}(w_j^* - w_j) - \theta_{j^*,j}(e_j + w_j^*) + \|e_j + w_j^*\| \sin\theta_{j^*,j}\overline{e_j + w_j}, w_j^* - w_j\rangle$$

$$\overset{①}{=}\sum_{j=1}^{d}\langle \frac{\pi}{2}(w_j^* - w_j) - \theta_{j^*,j}\|e_j + w_j^*\|_2(\overline{e_j + w_j^*} - \overline{e_j + w_j}) + \frac{\alpha_{j^*,j}|\theta_{j^*,j}|^3\|e_j + w_j^*\|\overline{e_j + w_j}}{3}, w_j^* - w_j\rangle$$

$$\overset{②}{\geq}\frac{\pi}{2}\|\mathbf{W}^* - \mathbf{W}\|_F^2 - \sum_{j=1}^{d} 1.001(1+\gamma)\|w_j^* - w_j\|_2^2\|\overline{e_j + w_j^*} - \overline{e_j + w_j}\|_2 - \sum_{j=1}^{d} 0.335(1+\gamma)\|w_j^* - w_j\|_2^4$$

$$\overset{③}{\geq}\frac{\pi}{2}\|\mathbf{W}^* - \mathbf{W}\|_F^2 - \sum_{j=1}^{d}\frac{1.001(1+\gamma)}{\sqrt{1-2\gamma}}\|w_j^* - w_j\|_2^3 - \sum_{j=1}^{d} 0.335(1+\gamma)\|w_j^* - w_j\|_2^4$$

$$\overset{④}{\geq}\left(\frac{\pi}{2} - 0.021\right)\|\mathbf{W}^* - \mathbf{W}\|_F^2$$

where ① uses Taylor's Theorem for $\sin\theta_{j^*,j}$, so we know $|\alpha_{j^*,j}| \leq 1$. ② uses Lemma E.1 term 3 and Cauchy Schwartz, ③ uses Lemma E.1 term 1, ④ holds since $\gamma \leq \frac{1}{100}$, and the two small order terms can be bounded by $0.021\|\mathbf{W}^* - \mathbf{W}\|_F^2$.

# J Proofs for Section 2

## J.1 Proof for Lemma 2.5

By the updating rule, we have

$$\mathbb{E}\|\mathbf{W}_{t+1} - \mathbf{W}^*\|_F^2 = \mathbb{E}\|\mathbf{W}_t - \mathbf{W}^* - \eta\mathbf{G}_t\|_F^2 = \mathbb{E}\|\mathbf{W}_t - \mathbf{W}^*\|_F^2 - 2\langle\mathbf{W}_t - \mathbf{W}^*, \eta\nabla f(\mathbf{W})\rangle + \eta^2\|\mathbf{G}_t\|_F^2$$
$$\leq \mathbb{E}\|\mathbf{W}_t - \mathbf{W}^*\|_F^2 - 2\langle\mathbf{W}_t - \mathbf{W}^*, \eta\nabla f(\mathbf{W})\rangle + \eta_t^2 G^2 \leq (1-2\eta\delta)\mathbb{E}\|\mathbf{W}_t - \mathbf{W}^*\|_F^2 + \eta^2 G^2$$

Now if $\eta\delta\mathbb{E}\|\mathbf{W}_t - \mathbf{W}^*\|_F^2 \geq \eta^2 G^2$, we know the $\mathbb{E}\|\mathbf{W}_t - \mathbf{W}^*\|_F^2$ will decrease by a factor of $(1-\eta\delta)$ for every step. Otherwise, although it could increase, we know

$$\mathbb{E}\|\mathbf{W}_t - \mathbf{W}^*\|_F^2 \leq \frac{\eta G^2}{\delta}$$

By setting $\eta = \frac{(1+\alpha)\log T}{\delta T}$, we know after $T$ steps, either $\mathbb{E}\|\mathbf{W}_T - \mathbf{W}^*\|_F^2$ is already smaller than $\frac{\eta G^2}{\delta} = \frac{(1+\alpha)\log T G^2}{\delta^2 T}$, or it is decreasing by factor of $(1-\eta\delta)$ for every step, which means

$$\mathbb{E}\|\mathbf{W}_T - \mathbf{W}^*\|_F^2 \leq \mathbb{E}\|\mathbf{W}_0 - \mathbf{W}^*\|_F^2(1-\eta\delta)^T \leq D^2 e^{-\eta\delta T} = D^2 e^{-(1+\alpha)\log T} = \frac{D^2 T^{-\alpha}}{T} \leq \frac{(1+\alpha)\log T G^2}{\delta^2 T}.$$

The last inequality holds since

$$T^\alpha \log T \geq \frac{D^2\delta^2}{(1+\alpha)G^2}$$

Thus, $\mathbb{E}\|\mathbf{W}_T - \mathbf{W}^*\|_F^2$ will be smaller than $\frac{(1+\alpha)\log T G^2}{\delta^2 T}$.